\documentclass{article}
\usepackage[margin=1in]{geometry}
\usepackage{graphicx} 
\usepackage{xr}
\usepackage{xr-hyper}
\usepackage[superscript]{cite}
\usepackage{authblk}
\usepackage{hyperref}
\usepackage{amsmath,amsfonts}
\usepackage[ruled,linesnumbered]{algorithm2e}
\usepackage{multirow}
\usepackage{tabularx}
\usepackage{url}
\usepackage{float}
\usepackage{svg}
\usepackage{booktabs}
\usepackage{fullpage}
\usepackage{spverbatim}
\usepackage{tcolorbox}

\author[1]{Reilly Cannon}
\author[1]{Nicolette M. Laird}
\author[2]{Caesar Vazquez}
\author[1]{Andy Lin}
\author[1, 2,*]{Amy Wagler}
\author[3,*]{Tony Chiang}

\affil[1]{Pacific Northwest National Laboratory, Richland, Washington 99354, United States}
\affil[2]{Department of Public Health Sciences, University of Texas at El Paso, El Paso, Texas 79968, United States}
\affil[3]{Department of Mathematics, University of Washington, Seattle, Washington 98195, United States}
\affil[*]{Corresponding authors}

\SetKwComment{Comment}{/* }{ */}

\title{Assessing Generative Models for Structured Data}
\date{\today}

\begin{document}

\maketitle

\begin{abstract}
Synthetic tabular data generation has emerged as a promising method to address limited data availability and privacy concerns. With the sharp increase in the performance of large language models in recent years, researchers have been interested in applying these models to the generation of tabular data. However, little is known about the quality of the generated tabular data from large language models. The predominant method for assessing the quality of synthetic tabular data is the train-synthetic-test-real approach, where the artificial examples are compared to the original by how well machine learning models, trained separately on the real and synthetic sets, perform in some downstream tasks. This method does not directly measure how closely the distribution of generated data approximates that of the original. This paper introduces rigorous methods for directly assessing synthetic tabular data against real data by looking at inter-column dependencies within the data. We find that large language models (GPT-2), both when queried via few-shot prompting and when fine-tuned, and GAN (CTGAN) models do not produce data with dependencies that mirror the original real data. Results from this study can inform future practice in synthetic data generation to improve data quality. 
\end{abstract}

\section{Introduction}
Limited data availability for training machine learning models has resulted in new methods for generating synthetic examples to feed to these models. Tabular data, containing a mix of real-valued (continuous), ordinal, and categorical features, is the predominant data format for the highly specialized sub-fields of health, medicine, and the biological and social sciences, often suffers from limited availability or has additional restrictions due to privacy issues when the data concerns individuals. Large language models (LLMs) have become a popular choice for generating synthetic data\cite{borisov2023language, seedatcurated}, largely replacing other synthetic data generation methods, such as tabular Generative Adversarial Networks (GANs) and Variational Auto-Encoders (VAEs) \cite{ctgan}. However, little work has been done on evaluating the quality of LLM-generated synthetic data and synthetic tabular data in general. Indeed, the most popular methods to evaluate synthetic tabular data quality, such as train-synthetic-test-real, indirectly measure data quality and utility. New methods and statistical techniques in general need to be developed to compare synthetic data to real data directly. 


This work proposes quantitative methods for evaluating the quality of synthetic tabular data, which we apply to assess the quality of synthetic data generated by LLMs. Our methods explore the relationships between columns in the tabular data, starting with marginal distributions, moving to pairwise dependencies, and then higher-order relationships in the features as measured through third and fourth-order joint cumulants. We use these quantities to measure the quality of tabular data generated from few-shot and fine-tuning methods. We also compare the LLM generated data to tabular data generated from a popular GAN-based method. Our results show that all of the tested generative methods fail at accurately reproducing the dependencies in the original data, even those of the training partition. On the surface (via marginal distributions) these generative models appear to replicate some of the real training data columns, but at higher-order relationships, both LLM and GAN methods fail to reproduce real data. Specifically, our LLM sampled via few-shot failed at producing second-order dependencies, while our GAN and fine-tuned LLM have mixed performance at the second, third, and fourth-order relationships.

\section{Related work}
\subsection{Synthetic Data Generation}
In the absence of sufficient or high-quality data, synthetic samples can be generated to create additional examples that can be used in downstream analyses. In a perfect world, these synthetic examples directly mimic the distribution of the original data. Several methods have been developed to model and sample these distributions. We focus on two types of synthetic data generation methods --- Generative Adversarial Networks (GANs) and Large Language Models (LLMs).

\subsubsection{Generative Adversarial Networks (GANs)}
GANs are a popular approach for generating synthetic data. Typically, a GAN consists of two competing neural networks - a generator that produces new synthetic examples, and a discriminator which evaluates them against real examples \cite{gan}. Although originally created to generate images \cite{gan}, they have proven effective, with modifications, when applied to tabular data sets. For example, MedGAN \cite{choi2017generating} utilizes a GAN with an autoencoder to generate continuous or discrete data. In addition to the neural networks of the generator and discriminator, TableGAN \cite{park2018data} adds a third neural network called a classifier to prevent non-sensical values in the generated data. CTGAN \cite{ctgan} incorporates a GAN architecture with mode-specific normalization through variational Gaussian mixture models (VGM). This approach uses a conditional generator and training-by-sampling to handle imbalanced discrete columns while incorporating fully connected networks in both the generator and discriminator to capture all possible correlations between columns. CTAB-GAN \cite{zhao2021ctab} modifies the loss function and conditional vector from CTGAN, and its successor \cite{zhao2022ctab} allows for the handling of mixed discrete-continuous data. However, while GANs have shown considerable success with continuous data (like images), they struggle with discrete data, such as categorical variables. This is primarily due to the nature of how GANs operate, which is better suited to continuous variates and the quality of this generated data can vary significantly based on the architecture and training process \cite{camino2018generating}. This work uses CTGAN, a popular open-source model with commercial backing, as the class representative.

\subsubsection{Large Language Models (LLMs)}

The rapid increase in the performance of LLMs in writing tasks has prompted interest in their use for synthetic data generation. Recent work has investigated the applicability of LLMs in generating synthetic tabular data. The authors of \cite{borisov2023language} propose a method for fine-tuning language models to output tabular data. Other work has shown that LLMs can be prompted to output tabular data without further training \cite{seedatcurated}. Others propose a token padding and compression procedure to reduce the computational cost of training LLMs on tabular data \cite{zhao2023tabula}. 

In this work we sample synthetic data generated from GPT-2 \cite{gpt2}, using both few-shot and fine-tuning methods.

\subsection{Measuring synthetic data quality}

Measuring the quality of synthetic data is a challenging task, especially when generating high-dimensional mixed-type data, such as tabular data. The structured nature of tabular data makes it especially difficult to analyze and compare synthetic data distributions to real ones. Several papers have sought to develop metrics to analyze synthetic tabular data quality. For example, some \cite{chen2019validity} have investigated the validity of tabular electronic health records (EHR) data generated from GAN and other synthetic data generators and proposed measures for detecting validity and proper use of the data. They considered both predictive reliability and descriptive fidelity of the EHR data using a variety of metrics. Others have proposed a generative framework for tabular data and evaluate their framework using statistical similarity measures on individual features and utility measures based on training downstream classifiers \cite{yoon2023ehr}. One work studied the utility of synthetic tabular data in downstream tasks \cite{manousakas2023usefulness}. Others use embedding models to create vector representations of the text and image data and analyze the vectors \cite{vargas2024understanding}.

A standard method to analyze synthetic data quality across multiple data types (structured and unstructured) is the train-synthetic-test-real (TSTR) approach \cite{esteban2017real}. In this approach, a variety of classifiers are trained on both real and synthetic data and then evaluated on a holdout set of real data. The rationale is that if the synthetic data is of high quality, the classifiers trained on this synthetic data should perform as well as the ones trained on real data. Below are the steps taken in the TSTR approach:
\begin{enumerate}
    \item Split the real (original) data into train and test partitions.
    \item Train the synthetic data generator on the train partition.
    \item Generate a synthetic data set of the same cardinality as the train partition.
    \item Select a variety of statistical and machine learning models for a desired downstream task.
    \item Train pairs of these task-specific models on the train and synthetic sets separately.
    \item Test the models using the test set and metrics relevant to the downstream task.
\end{enumerate}

However, this approach requires the user to know the downstream tasks they are interested in before evaluating synthetic data. Additionally, TSTR can only indirectly assess the predictive performance of synthetic data, potentially missing spurious correlations or out-of-distribution samples created in synthetic data. Thus, it is not informative for measuring how accurately the synthetic data models the distribution of the real data. In contrast, our evaluation metrics are task-independent, and we compare the synthetic and real (original) distributions directly.

\section{Methods}

\subsection{Datasets}
We generated synthetic data that resembled the following three publicly available datasets commonly used for classification tasks: Adult\cite{adult_2}, Breast Cancer Wisconsin (Diagnostic)\cite{Street1993NuclearFE, breast_cancer_wisconsin_(diagnostic)_17}, and Credit Card Fraud Detection\cite{credit-ds} (Table \ref{data_table}). These datasets were chosen as representative examples of the type of tabular data that can be found. For example, the Breast Cancer and Credit datasets represent datasets with a small and large number of samples, respectively.  Each dateset also contains a different mix of column types. For example, the Adult dateset only contains columns with discrete values (i.e. ordinal and categorical) while the the other two datasets both have columns that contain discrete or continuous values. Additionally, the continuous values have different ranges, with the Breast Cancerdataset values falling within a small range while the continuous columns within the Credit dataset have a large range of values. Since these datasets are often used in classification tasks, they also represent different class imbalance levels. For example, the Breast Cancer and Adults datasets are fairly balanced, while the Credit dataset's target feature is highly imbalanced.

\begin{table}
    \centering
    \small 
    \setlength{\tabcolsep}{4pt} 
    \begin{tabular}{ccccc}
        \textbf{Dataset} & \textbf{\# Rows} & \textbf{\# Continuous Cols} & \textbf{\# Ordinal Cols}& \textbf{\# Categorical Cols} \\
        \hline 
        Adult & 45,222 & 0 & 6 & 9 \\
        Breast Cancer & 569 & 30 & 0 & 1 \\
        Credit & 284,807 & 29 & 1 & 1 \\
        \hline 
    \end{tabular}
    \caption{{\bf Datasets.} Table describing the number and type of columns in the three datasets used in this work.}
    \label{data_table}
\end{table}

The Adult dataset was retrieved directly from the UCI ML data repository and contains contains data used to predict whether income exceeds \$50k per year based off census data.
We excluded the variables ``fnlwgt'' and ``education'' because they are either already encapsulated in another feature or represent an estimated or unknown value. The ``fnlwgt'' feature represents the ``final weight'' of the record and can be interpreted as the number of people represented by the row. The information in the ``education'' feature can be inferred from the ``education num'' variable and is thus redundant.
On the other hand, the Breast Cancer dataset was retrieved from the UCI ML data repository via \texttt{scikit-learn}\cite{scikit-learn}, and this dataset is used to predict the occurrence of breast cancer. Finally, the Credit dataset was obtained from Kaggle\footnote{\url{https://www.kaggle.com/datasets/mlg-ulb/creditcardfraud/}}, and this dataset contains data simulating credit bureau data used to predict credit fraud.


\subsection{Synthetic Data Generation}
We used three models to generate synthetic tabular data: a few-shot prompted GPT-2, a fine-tuned GPT-2 \cite{gpt2}, and a GAN.
For the few-shot prompted model, we modified a previously described prompt\cite{seedatcurated}.
In this prompt, GPT-2 was given a set of real example data points and asked to generate a target number of synthetic examples (Supplemental Fig. \ref{fig:zero_shot_prompt} and \ref{fig:zero_shot_prompt_example}).
Following the generation of the response, a sanity check was performed to remove malformed examples. Entries were considered malformed if they contained an incorrect number of column entries and/or if the response returned header names. We re-prompted the model to generate the desired number of samples giving new examples from the original train dataset each time. We set a 5000-hour time cutoff for the few-shot task to complete. GPT-2 was sometimes unable to generate the specified number of valid samples in the specified time limit. In these cases, we perform further analyses on the valid samples only, even if there are very few of them.

Following the generation of the samples, we further checked the validity of synthetically generated samples by filtering out entries that did not match features found in the original data. An entry was considered invalid if any column value could not be coerced into the a format and type that matched that of the original real dataset. For example, if a value could not be coerced into a numerical value for a numerical column, then that entry was deemed invalid and removed. For categorical data, an addition filter conducted to ensure that the synthetic entries contained values that were present in the original dataset. 

To create the fine-tuned GPT-2 model, we obtained the original model weights from Hugging Face \footnote{\url{https://huggingface.co/openai-community/gpt2}} and fine-tuned it using the \texttt{GReaT} framework \cite{borisov2023language} with a max token length of 500, a batch size of 9, and a temperature of 0.7. We created a different fine-tuned model for each dataset, and prompting of the fine-tuned model was performed by the \texttt{GReaT} package. 
Training hyperparameters for this process can be found in Supplemental Table \ref{table:GPT_2_ft_ds_hp} and \ref{table:GPT-2_ft_hyperparams}.

Finally, for a comparison with a non-language model approach, we used the CTGAN\cite{ctgan} implementation from Synthetic Data Vault \cite{SDV}. 
CTGAN is a method used to fine-tune GANs to generate realistic synthetic tabular data.
Training hyperparameters for this process can be found in Supplemental Table \ref{table:gan_hyperparams}.

The process described above was performed $15$ times per model per dataset. For the models that required training, the dataset was split into a training and test set with a 70/30 split with different random seeds for each of the three datasets. These splits were constant across all methods and random seeds. No model was trained on any point from the test partition, and all models were trained on the same training set. For the few-shot prompted GPT-2 model, all prompt examples were taken from the train partition. 

\subsection{Assessing quality of tabular synthetic data}

To measure the quality of synthetically-generated tabular data we compared how accurately synthetic data simulates the distribution of real data at various levels. Specifically, we considered whether the cumulants of the synthetic data matched real data.
Our method is predicated on the fact that two distributions are identical when the cumulants are also identical. Note that cumualants can only be calculated on ordinal and continious variables. 

First, we compared first-order cumulants of real and synthetic by comparing the marginal distributions for each of the features within a dataset. Next, the second-order cumulants were compared by considering the association between pairs of features within a dataset. A $\chi^2$ test determined the association between pairs of features. Finally, higher-order dependencies between three (third-order) and four (fourth-order) sets of features were calculated. Additional information regarding the generation and comparison of second-order and higher cumulants is below. 

\subsubsection{Dependency between pairs of features}

To understand the dependencies between arbitrary pairs of features within a dataset, we chose to calculate the effect size. Effect size is a measure of the strength of dependency between features. We chose effect size because they are interpretable and have standardized ways to analyze dependencies between large sets of features. Specifically, we reported the effect size based on a $\chi^{2}$ association test as the bi-variate similarity measure. We calculated effect size using
\begin{equation}
    \sqrt{\chi^{2} / (n*df)}, 
\end{equation} where $df$ is the degrees of freedom of the $\chi^{2}$ test of interest and $n$ is the total number of observations. Typically $df$ is defined as $(r-1)\times (c-1)$, where $r$ is the number of rows and $c$ the number of columns, but it is subject to the number of levels in the observed contingency tables, which varied for some synthetic datasets \cite{effectsize}.
We chose this measure over other correlation measures, such as Pearson and Spearman, due to the prevalence of ordinal features within tabular data. 

Following the generation of the dependencies values for each pair of features within a dataset, the partial bi-variate association values was estimated using graphical lasso (glasso) estimation with Extended Bayesian Information Criterion (EBIC)\cite{friedman2008sparse,friedman2011glasso}. In brief, glasso computes a sparse Gaussian graphical model by estimating the precision matrix of a latent multivariate Gaussian distribution that provides conditional independence relationships between variables based on observed data. The tuning parameter for the graphical lasso is chosen using EBIC, which balances model fit with complexity.

After the generation of the partial bi-variate association values, the resulting values were visualized using a graph where nodes are features and edges are effect sizes between features. Edges in the networks are color-coded to reflect the directionality of the relationships (red for negative correlations, green for positive correlations). The dependency networks are built in \texttt{igraph} \cite{Csardi2006,Csardi2024}.


\subsubsection{Community detection in dependency networks}
Following the generation of the dependency networks the community detection was performed using the Louvain algorithm to identify highly dependent groups of features \cite{louvain}. The Louvain algorithm maximizes the objective function in the network by recursively merging nodes (features) into a community. Note a Constant Potts Model (CPM) objective function was used due to the mix of categorical and numerical features within the datasets \cite{tran2023potts}. CPM assesses how well communities are allocated while capturing the trade-off between intra-community cohesion and inter-community separation. A brief description of the steps in the Louvain algorithm can be found in the Supplement (Supplemental \ref{more_louvain}).

\subsubsection{Measuring higher-order dependence}
Joint cumulants provide information about how well the higher-level dependencies are preserved (beyond second-order) when synthesizing data. We computed these cumulants up to the fourth order on the numerical (continuous and integer) columns, and we summarized the similarities between the real and synthetic data dependencies using visualizations and by reporting the sensitivity and specificity of the joint cumulants to indicate similar higher-order dependencies. 

Using standard notation \cite{speed1983cumulants} and assuming \(X\) is a random vector with components denoted $(X_{1}, X_{2}, \dots, X_{p})$, joint cumulants are computed via the cumulant generating function $\log\{E(\exp(\langle t,x \rangle) \}$ by taking the derivative and evaluating it at \(t=0\). We use the notation \(\kappa(X_{i})\) to denote a first order cumulant with \(i=1,\ldots,p\). Note that this notation is easily extended to include joint cumulants such as \(\kappa(X_{i},X_{j})\) or \(\kappa(X_{i},X_{j}, X_{k})\) where \(i, j, k = 1,\ldots,p\).  For a more accessible definition, we recognize that joint cumulants are tensor-valued high-order statistics and, borrowing from Speed's notation, we alternatively define joint cumulants in the following manner: For a set of \(p\) random variables \(X_{1},\ldots,X_{p}\), the {\em p}-dimensional joint cumulant operates on the product of these $p$ variables and is defined by 
\begin{equation} \label{eq:cum}
 \kappa \{X_{1}, X_{2}, \ldots, X_{p}\}=\sum_{\sigma} (-1)^{b(\sigma)-1} (b(\sigma)-1)! \prod_{a=1}^{b(\sigma)} E\{\prod_{i \in \sigma_{a}} X_{i} \}
\end{equation}
where the sum is over all partitions \(\sigma\) of \({1,\ldots,p}\) into \(b(\sigma)\) blocks (denoted \(\sigma_{1}, \sigma_{2}, \ldots, \sigma_{p}\)). Since the joint cumulants are functions of the product moments of the partitions of the random variables \(\{X_{i}\}\), they may be calculated from these raw moments using point estimates of the product moments. This view of joint cumulants also implies that the cumulant exists and is finite when all the corresponding moment and all lower-order moments also exist and are finite. The joint cumulant estimates are provided for all features with a numerical scale and ordering, thus features are label encoded to enable computation. See Di Nardo and Guarino for computational details of the calculations utilized for estimating the joint central cumulants \cite{DiNardo2022}.

\section{Results}

\subsection{LLMs and GANs can accurately approximate marginal distributions}

To understand whether LLMs and other machine learning models faithfully generate synthetic data we first considered whether these models can create synthetic data where the marginal distributions for each feature match real data. A good quality synthetic dataset would be one where the marginal distributions match across real and synthetic data.

To test whether these models could produce good quality synthetic data we performed a two-sample Kolmogorov-Smirnov (KS) test over the continuous columns of the datasets, where the reference sample was the training set. In addition, we visually compared the marginal distributions from the real data against the marginal distributions created from the synthetic data. 

\begin{figure}
    \centering
    \begin{tabular}{l}
        a \\
        \includegraphics[width=0.5\linewidth]{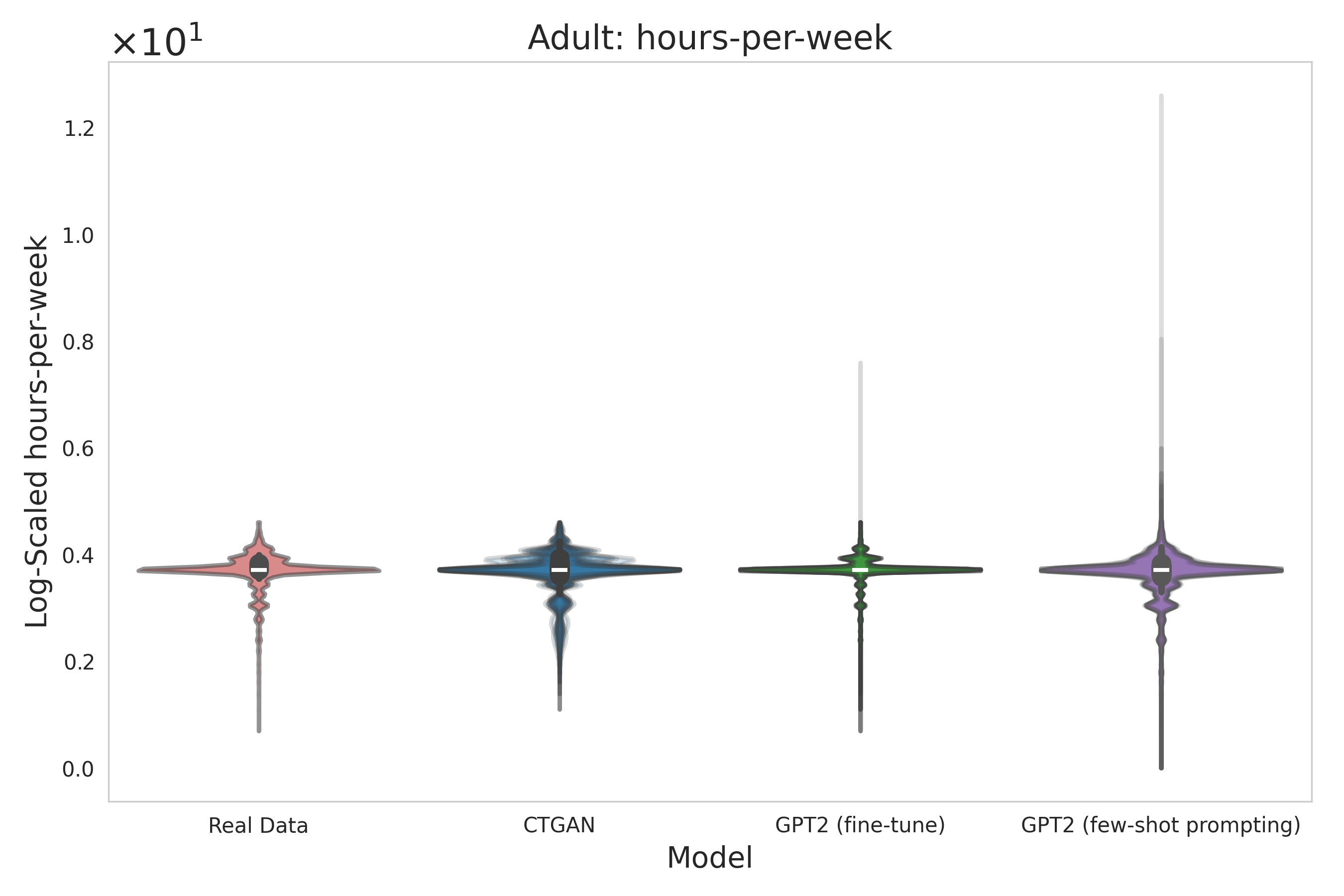}\\
        b \\
        \includegraphics[width=0.5\linewidth]{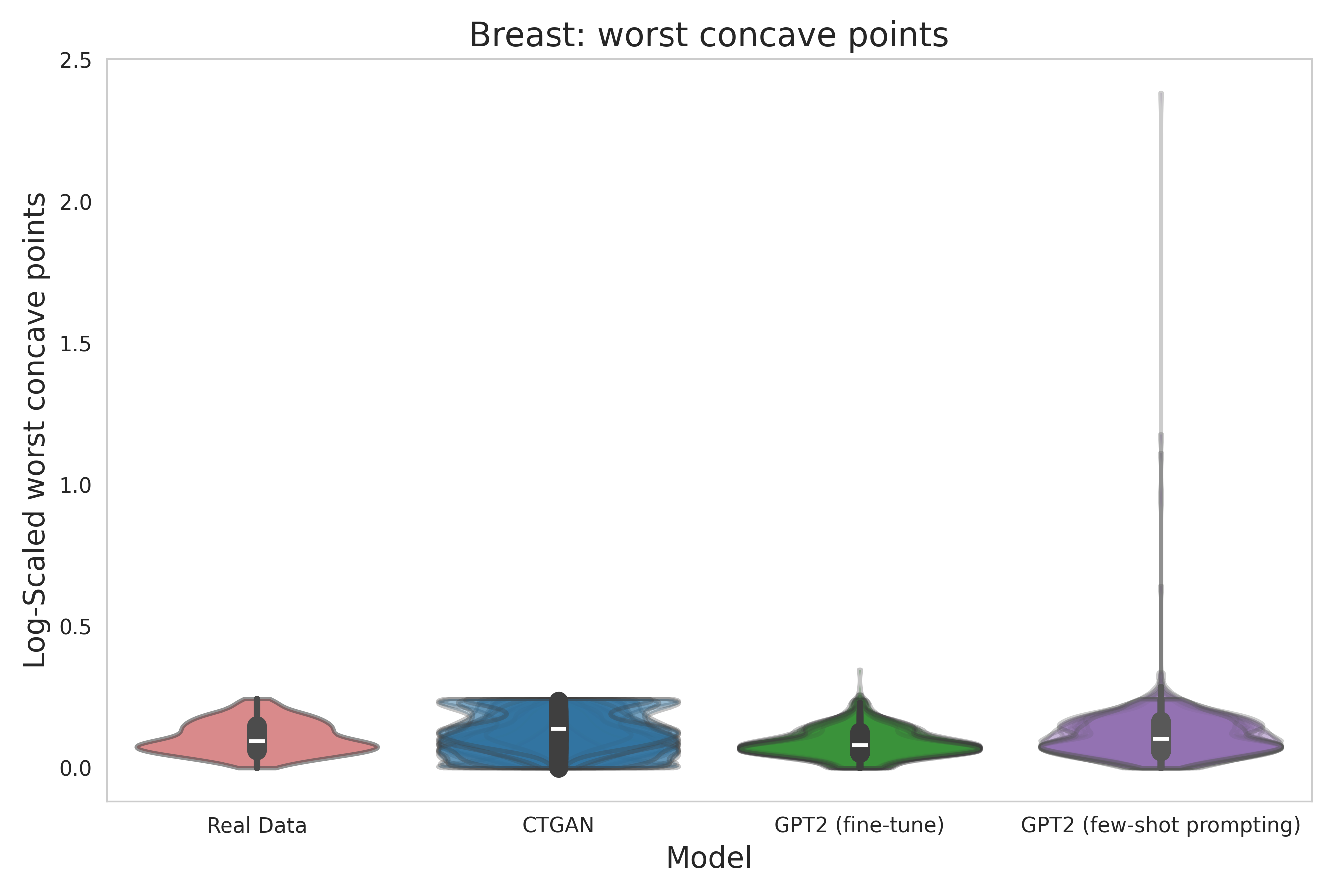}\\
        c \\
        \includegraphics[width=0.5\linewidth]{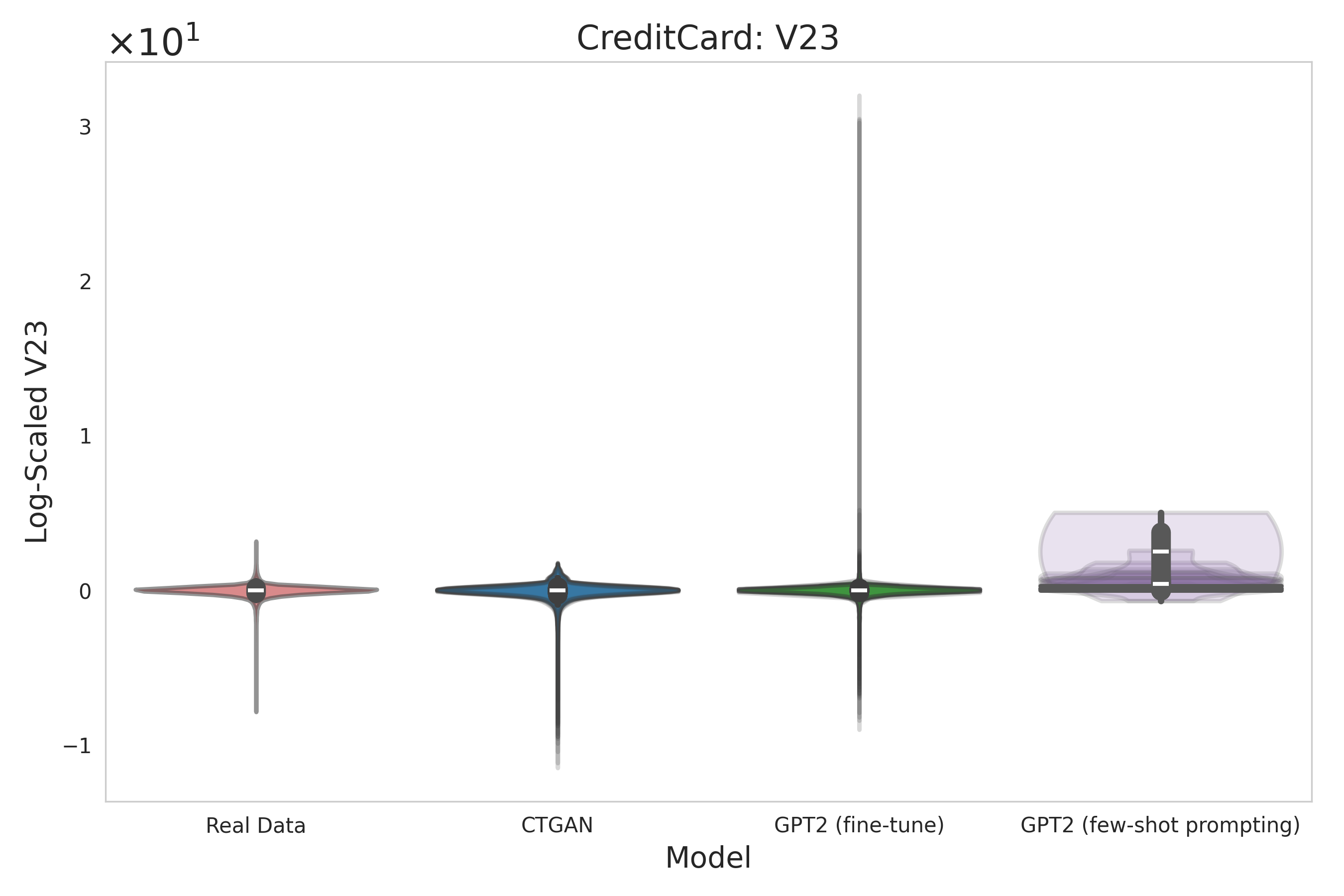}\\
    \end{tabular}
    \caption{{\bf Marginal distributions}. Violin plots showing the marginal distribution of a column from real and synthetic data. a) Marginal distribution of the ``hours-per-week'' column from the Adult dataset. b) Marginal distribution of the ``worst concave points'' column from the Breast Cancer dataset. c) Marginal distribution of the ``V23'' column from the Credit dataset. There is a separate plot for each of the 15 trials conducted for CTGAN, fine-tuned GPT-2, and few-shot prompted GPT-2. The real data distributions comes from the training set. In general, the synthetic data produced by fine-tuned GPT-2 most closely matched the real data.
    Violin plots for the remaining continuous columns can be found at Supplemental Fig. \ref{fig:adult_violins}--\ref{fig:credit5_violins}.}
    \label{fig:marginals}
\end{figure}

We found that both versions of GPT-2 (fine-tuned and few-shot prompted) and CTGAN do a reasonable, but imperfect job of producing synthetic data that result in marginal distributions of continuous features that match real data (Figure \ref{fig:marginals} and Supplemental Fig. \ref{fig:adult_violins}--\ref{fig:credit5_violins}). In addition to continuous columns, we also considered categorical columns and found that the values in the real data had the similar proportions as the synthetic data.

Overall, we found that both GPT-2 models generally outperformed CTGAN (Table \ref{ks_tests}). When comparing the two GPT-2 models together, we found that the fine-tuned GPT-2, on average, performed better than the few-shot prompted GPT-2 model. However, the few-shot prompted GPT-2 model performed exceptionally well on the Breast Cancer dataset. We speculate that this high performance was due to generated examples were near-replicas of those in the prompts, with a few values changed.

\begin{table}[H]
    \centering
    \begin{tabular}{lcccc}
    & CTGAN & GPT-2 (fine-tune) & GPT-2 (few-shot) & Test Partition\\
    \hline
    Breast & 0.50 & 0.18 & 0.06 & 0.11 \\
    Credit & 0.24 & 0.11 & 0.45 & 0.01 \\
    \hline
    \end{tabular}
    \caption{{\bf Two sample Kolmogorov-Smirnov test statistic.} Comparisons are made using the original training data as the reference sample. This statistic was calculated by (1) for each of the 15 trials, taking the largest KS-statistic over the continuous column, then (2) reporting the lowest number across the trials from (1). Thus the reported number corresponds to the best-performing training run as measured by the worst-case KS-statistic over the continuous columns. Note that we did not perform a KS test on the Adult dataset as it did not contain any continuous columns.}
    \label{ks_tests}
\end{table}

In addition to the KS-test, a visual inspection of the marginals also showed that the fine-tuned GPT-2 model and CGTAN usually had the most and least accurate synthetic data, respectively (Figure \ref{fig:marginals} and Supplemental Fig. \ref{fig:adult_violins}--\ref{fig:credit5_violins}). The fine-tuned GPT-2 model usually accurately approximated the marginal distributions, with some exceptions that most commonly occurred in the credit dataset. In addition, this model had low variability across runs. Furthermore, the range of the distributions generally closely approximated those of the real data for the Adult and Breast Cancer dataset, despite not using explicit methods to ensure this. On the other hand, this model generated values far beyond the range of the real data for the credit dataset (Figure \ref{fig:marginals}C and Supplemental Fig. \ref{fig:credit1_violins}--\ref{fig:credit5_violins}). Finally, we saw that fine-tuned GPT-2 frequently underestimated the density of the tails of the marginals.

When considering synthetic data generated by few-shot prompted GPT-2 we found that the model could accurately, on average, reproduce the bulk of the marginal distributions across the 15 trials. However, it frequently produced values that were far outside the range of the real data and had large variance across tails. In comparison, CTGAN had moderate variability across the 15 trials, likely because the models were randomly initialized with different seeds for each run. In addition, this model frequently overestimated the density at the tails of the distributions and often had a hard cut off at either end of the distribution. This is explained by our enabling of the parameter that restricts CTGAN to generate data within the range of the training data.

\subsection{Fine-tuned LLMs and GANs show mixed performance at modeling pairwise dependencies}

Our analyses so far suggest that LLMs and GANs can generate synthetic data with marginal distributions that match marginal distributions derived from real data. Next, we investigated whether these models are capable of generating synthetic data that match second order cumulants from real data. To test this, we calculated the bi-variate feature dependencies for all pairs of features in a dataset by first calculating the effect size of each pair of features using a $\chi^2$ statistic. Then, we estimated the partial bi-variate associations using graphical lasso. Finally, we visualized the difference in feature pair dependency values between real and synthetic data using a heatmap where rows and columns are features and cells are the difference in association effect size between real and synthetic data for a specific pair of features. 

\begin{figure}
   \centering
   \includegraphics[width=.9\linewidth]{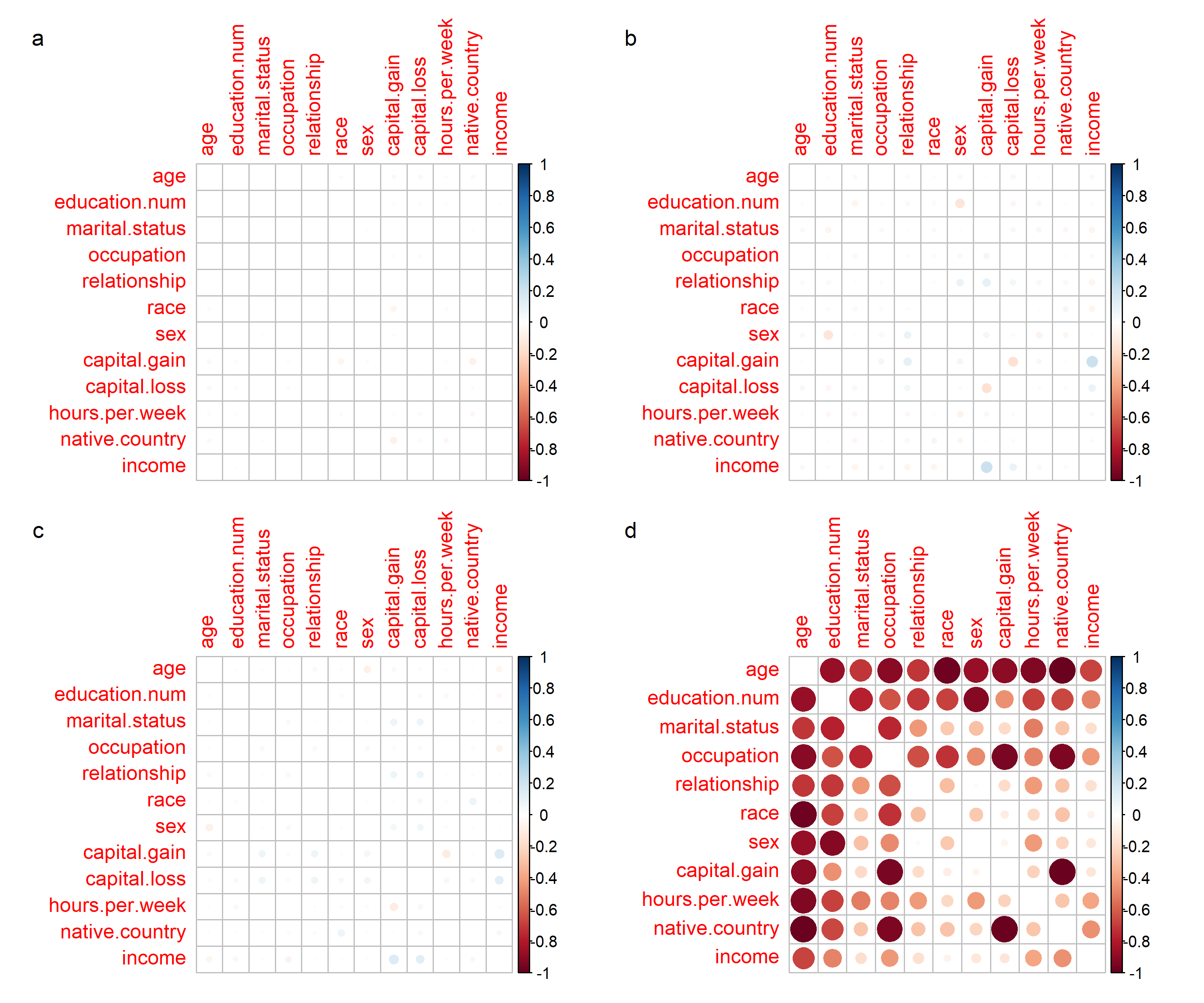}
   \caption{{\bf Difference in association effect size between real and synthetic data.} A heatmap showing the difference in association between real and synthetic data created by (top left) resampling of the train set, (top right) CTGAN, (bottom left) fine-tuned GPT-2 fine-tuned, and (bottom right) few-shot prompted GPT-2 for pairs of features in the Adult dataset.
   The heatmaps are shown are the best quality data generated by each method over the $15$ trails according to the lowest absolute determinant of the dependency matrix. Heatmaps for the Breast Cancer and Credit dataset can be found in Supplementary Fig. \ref{fig:breast_heatmap} and \ref{fig:credit_heatmap}, respectively. 
   }
   \label{fig:assoc_plots}
\end{figure}

Overall, we found that the three models obtained mixed performance at modeling pairwise dependencies (Figure \ref{fig:assoc_plots}, Supplementary Fig. \ref{fig:breast_heatmap}, and Supplementary Fig. \ref{fig:credit_heatmap}).
Specifically, we found models that are fine-tuned or trained on a specific real dataset generally generated better quality synthetic data. For example, the fine-tuned GPT-2 model consistently generated data with a similar data structure as the corresponding real dataset for all three datasets (Figure \ref{fig:assoc_plots}B and Supplementary Fig. \ref{fig:breast_heatmap}--\ref{fig:credit_heatmap}B). 
In addition, the CTGAN model created synthetic data that closely matched the Adult dataset (Figure \ref{fig:assoc_plots}C). However, it failed to closely match the Breast Cancer and Credit dataset (Supplementary Fig. \ref{fig:breast_heatmap}--\ref{fig:credit_heatmap}B). Finally, we found that the data obtained from the few-shot prompted GPT-2 had little fidelity to the original data structure (Figure \ref{fig:assoc_plots}C and Supplementary Fig. \ref{fig:breast_heatmap}--\ref{fig:credit_heatmap}C).

Next, we aimed directly compared the fine-tuned GPT-2 model against the CTGAN model to understand which model generated the best quality synthetic data. We measured synthetic data quality by calculating the absolute value of the determinant of the difference between the dependency matrix created from synthetic data and the same matrix produced from real data. This process was repeated for each of the $15$ trials for each of the datasets. 

\begin{table}
    \centering
    \begin{tabular}{lllll}
        dataset & model & min & median & max \\
        \hline
        \multirow{2}{*}{Adult} & GPT-2 (fine-tune) & $9.5\times10^{-18}$ & $2.7\times10^{-16}$ & $7.59\times10^{-15}$ \\
        & CTGAN & $1.4\times10^{-16}$ & $1.4\times10^{-16}$ & $1.0\times10^{-13}$ \\
        \multirow{2}{*}{Breast} & GPT-2 (fine-tune) & $3.3\times10^{-24}$ & $1.4\times10^{-20}$ & $1.7\times10^{-18}$ \\
        & CTGAN & $1.8\times10^{-4}$ & $8.7\times10^{-4}$ & $2.1\times10^{-2}$ \\
        \multirow{2}{*}{Credit}  & GPT-2 (fine-tune) & $1.7\times10^{-38}$ & $1.9\times10^{-23}$ & $2.1\times10^{-18}$ \\
        & CTGAN & $9.1\times10^{-16}$ & $2.5\times10^{-15}$ & $1.2\times10^{-14}$ \\
        \hline
    \end{tabular}
    \caption{Minimum, median, and maximum over the $15$ trials of the magnitude of the determinant of the difference between the pairwise dependencies in the synthetic dataset and those in the real dataset.}
    \label{table:volumes}
\end{table}

We found that CTGAN outperformed the fine-tuned GPT-2 model on the Breast Cancer and Credit dataset (Table \ref{table:volumes}). For example, on the Breast Cancer dataset, CTGAN outperformed the fine-tuned GPT-2 model with minimum, median and maximum determinants that were at least $10^{16}$ times smaller than those of GPT-2. As another example, the best performing fine-tuned model performs significantly outperformed CTGAN for the Credit dataset (min column in Table \ref{table:volumes}). On the hand, the fine-tuned GPT-2 either obtained slightly better or equal performance as CTGAN on the Adult dataset. We speculate that CTGAN performed worse than fine-tuned GPT-2 on the Breast Cancer and Credit datasets because of the lack of good training data. Specifically, the Breast Cancer dataset had a small number of samples and the Credit dataset was relatively sparse.


Looking across all three datasets we found that each model was most successful in synthetically generating the Adult dataset and least successful generating the Breast Cancer and Credit dataset. We speculate that this phenomenon, as before, is due to the small number or sparsity of samples used for training. 

\begin{figure}
    \centering
    \includegraphics[width=0.9\linewidth]{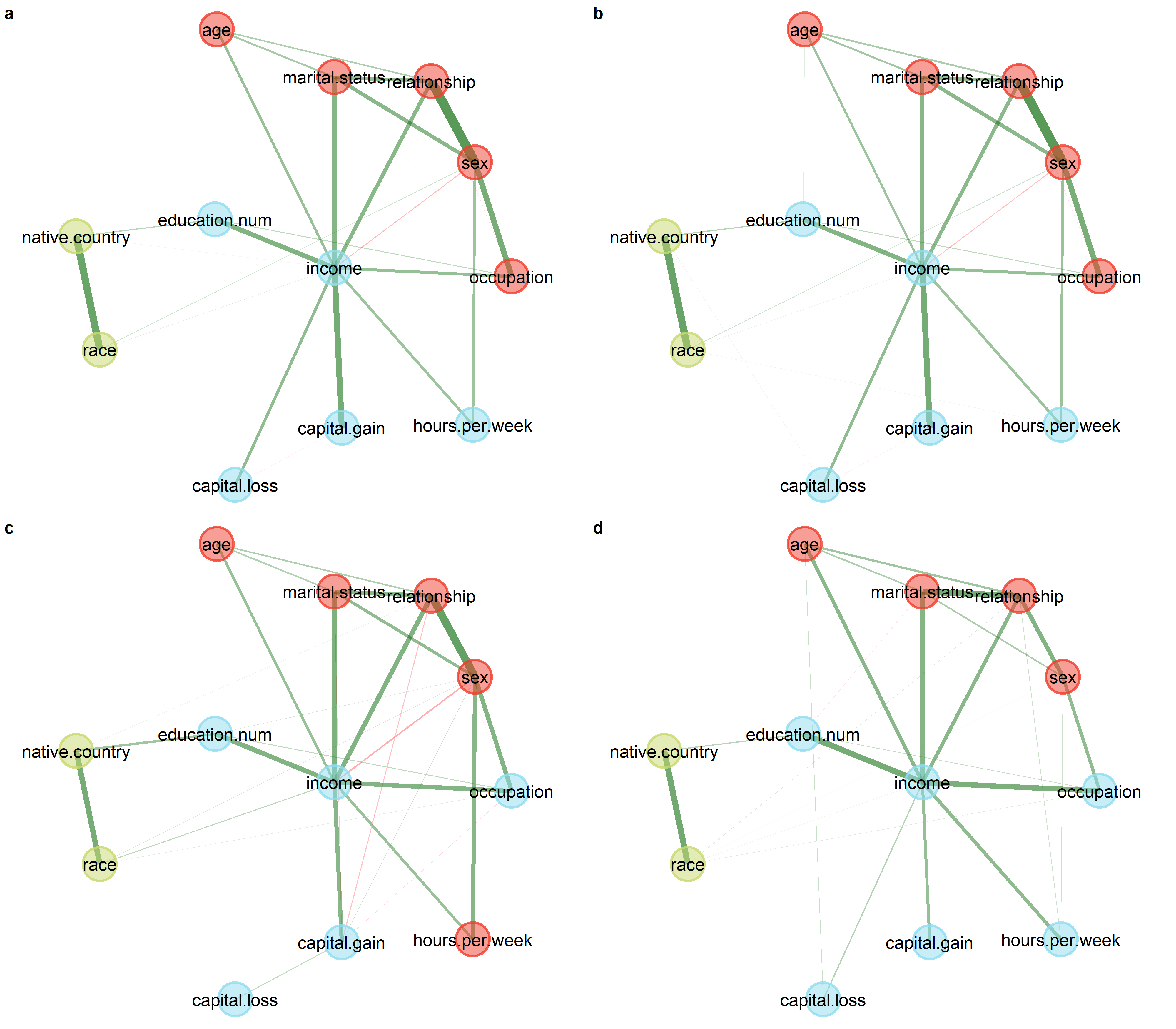}
    \caption{Representative network plots of the dependency between real and synthetic data for the Adults dataset: (a) Adult train set, (b) resampling of the train set, (c) CTGAN, and (d) GPT-2 (fine-tune). The node colors represent the clusters found using the Louvain community detection algorithm. The edge colors reflect the directionality of the relationships (red for negative correlations, green for positive correlations). Displayed models were chosen according to the lowest absolute determinant of the dependency matrix.}
    \label{fig:network_plots}
\end{figure}

To further analyze the performance of fine-tuned GPT-2 and CTGAN, we generated a graph from the matrix of pairwise feature association values and performed Louvain community detection on the graph to determine whether groups of highly correlated features matched across real and synthetic data. In this graph nodes are features (i.e., columns in dataset) and edges denote the association between two features where edge width represent strength of an association. Nodes that are part of the same community are colored the same. Note that we conducted this process separately for each of the $15$ trails for each dataset. 

Overall we found that CTGAN and fine-tuned GPT-2 failed to generate synthetic data that yielded the same clusters as real data despite their ability to generate high-fidelity pairwise dependencies (Figure \ref{fig:network_plots} and Supplementary Fig. \ref{fig:breast_graph}, \ref{fig:credit_graph}).
For example, in the Adult dataset both models created data resulted in the correct number of clusters (i.e., three) but incorrectly caused the node labeled  ``occupation'' to be clustered with nodes such as ``income'' and ``capital gain'' (blue nodes in Figure \ref{fig:network_plots}) instead of nodes such as ``sex'' and ``relationship'' (red nodes in Figure \ref{fig:network_plots}). For the credit dataset both models failed to create data that yielded the same number of clusters as real data (Supplementary Fig. \ref{fig:credit_graph}). Specifically, CTGAN generated synthetic data that yielded three clusters and the fine-tuned GPT-2 yielded four clusters when the real data only had two clusters. 
Finally, for the Breast Cancer dataset, CTGAN generated data that incorrectly indicated that most features were not part of any cluster (Supplemental Fig. \ref{fig:breast_graph}C). On the other hand, the fine-tuned GPT-2 created data that had the correct number of clusters but failed to create clusters with the correct feature set (Supplemental Fig. \ref{fig:breast_graph}D).


\subsection{Fine-tuned LLM and GAN methods fail to reproduce higher-order dependencies}

After investigating the ability of LLMs and GANs to generate synthetic data with lower-order feature dependencies that matched real data, we next investigated the ability of these models to generate data that also result in higher-order dependencies that matched real data. To this end, we compared third and fourth order joint cumulants from synthetic data generated by CTGAN and the fine-tuned GPT-2 model against real data. Third and fourth order cumulants can be thought of as the associations between sets of three and four features, respectively. We hypothesize that generating data with these more complex associations would be more difficult for machine learning models to achieve. Note that we did not include the few-shot prompted GPT-2 model because of its previous poor performance. In addition, we note that this analysis was only conducted on ordinal and continuous columns from the datasets as cumulants are unable to be calculated on categorical data

\begin{figure}
    \centering
    \begin{tabular}{lll}
        & \multicolumn{1}{c}{\textbf{third order}} & \multicolumn{1}{c}{\textbf{fourth order}} \\
        
        \rotatebox{90}{\hspace*{0.8in}\textbf{Adult}} & 
        \includegraphics[width=2.5in]{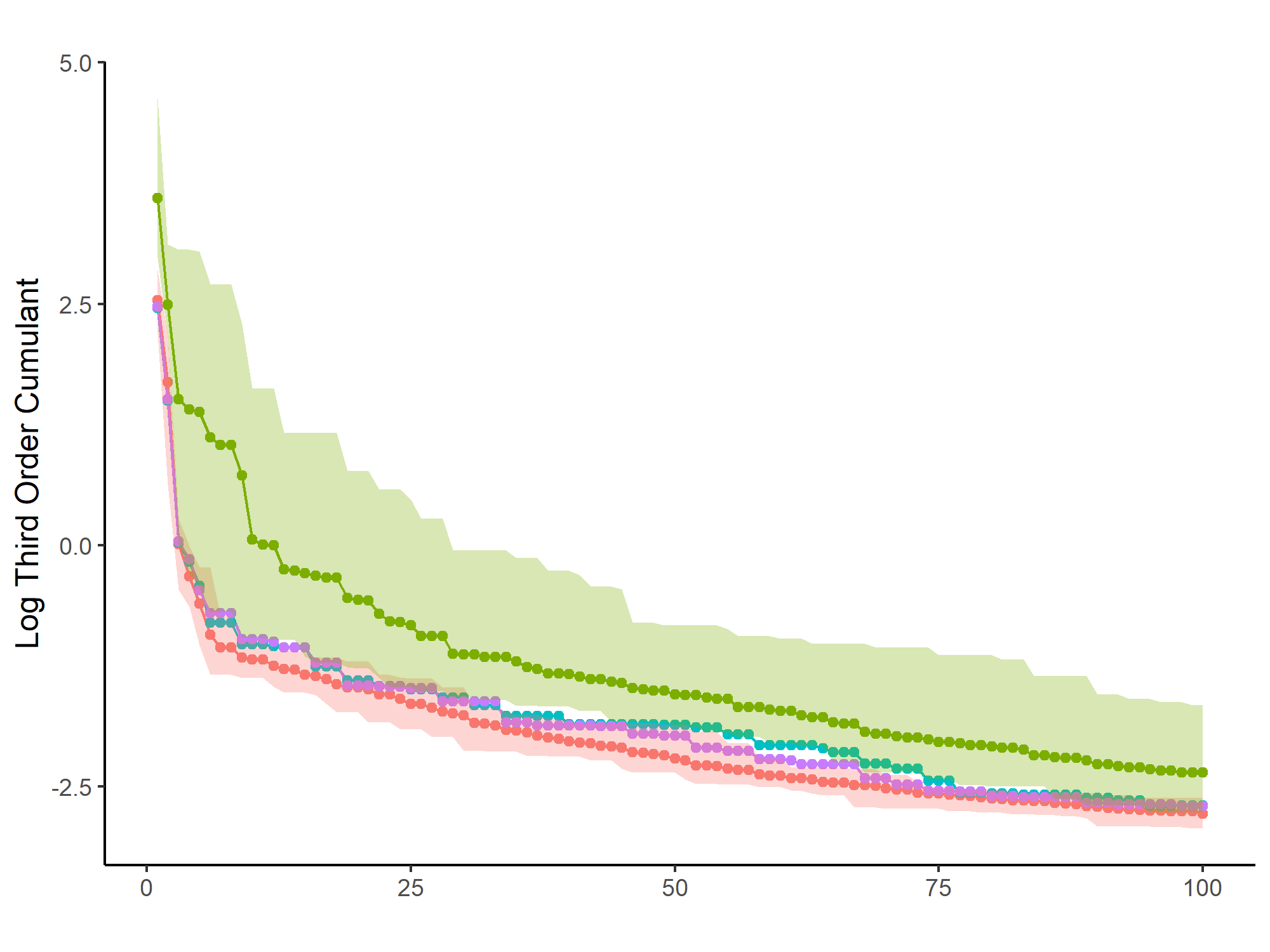} &
        \includegraphics[width=2.5in]{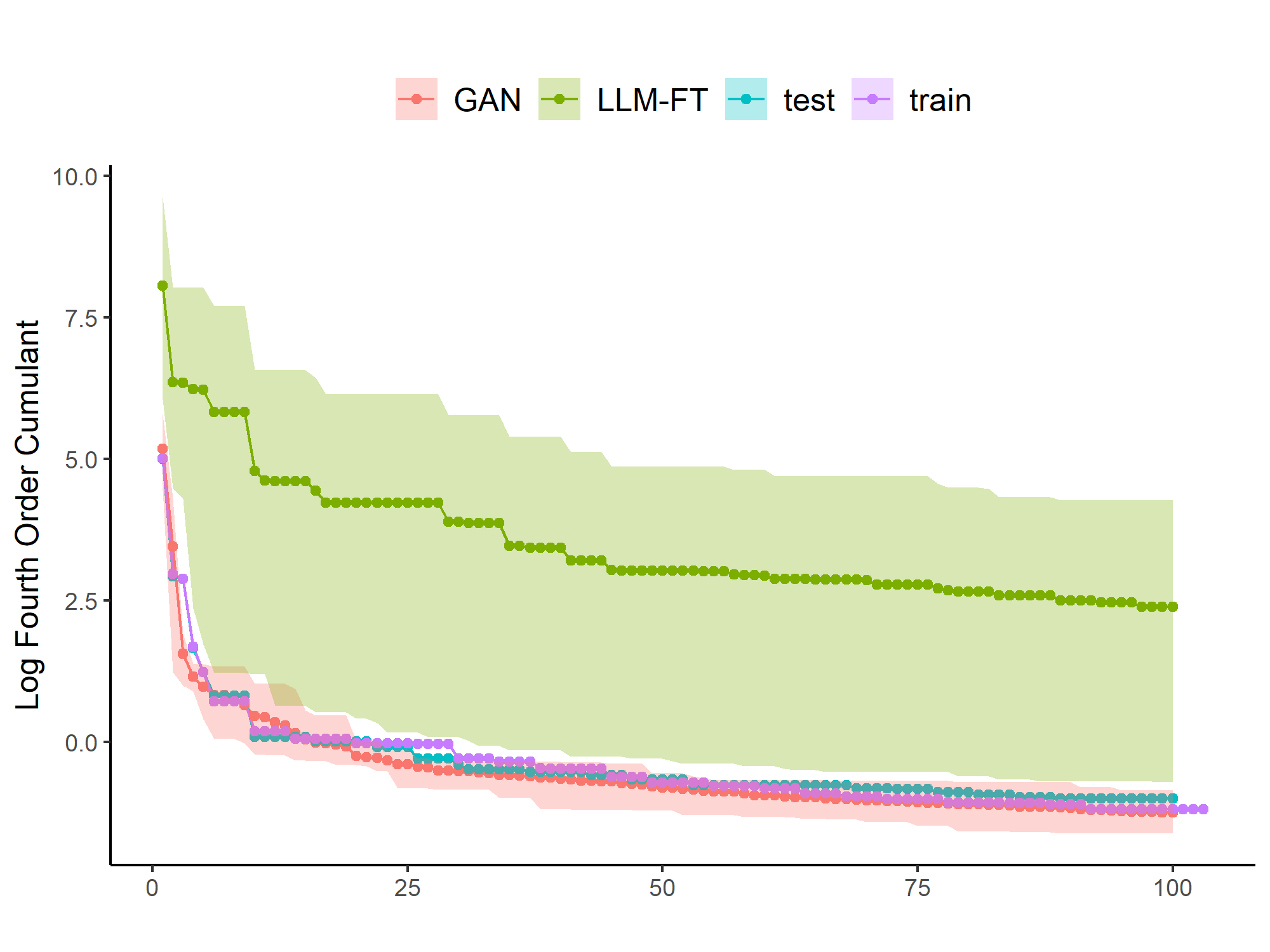}\\
        
        \rotatebox{90}{\hspace*{0.8in}\textbf{Breast}} &
        \includegraphics[width=2.5in]{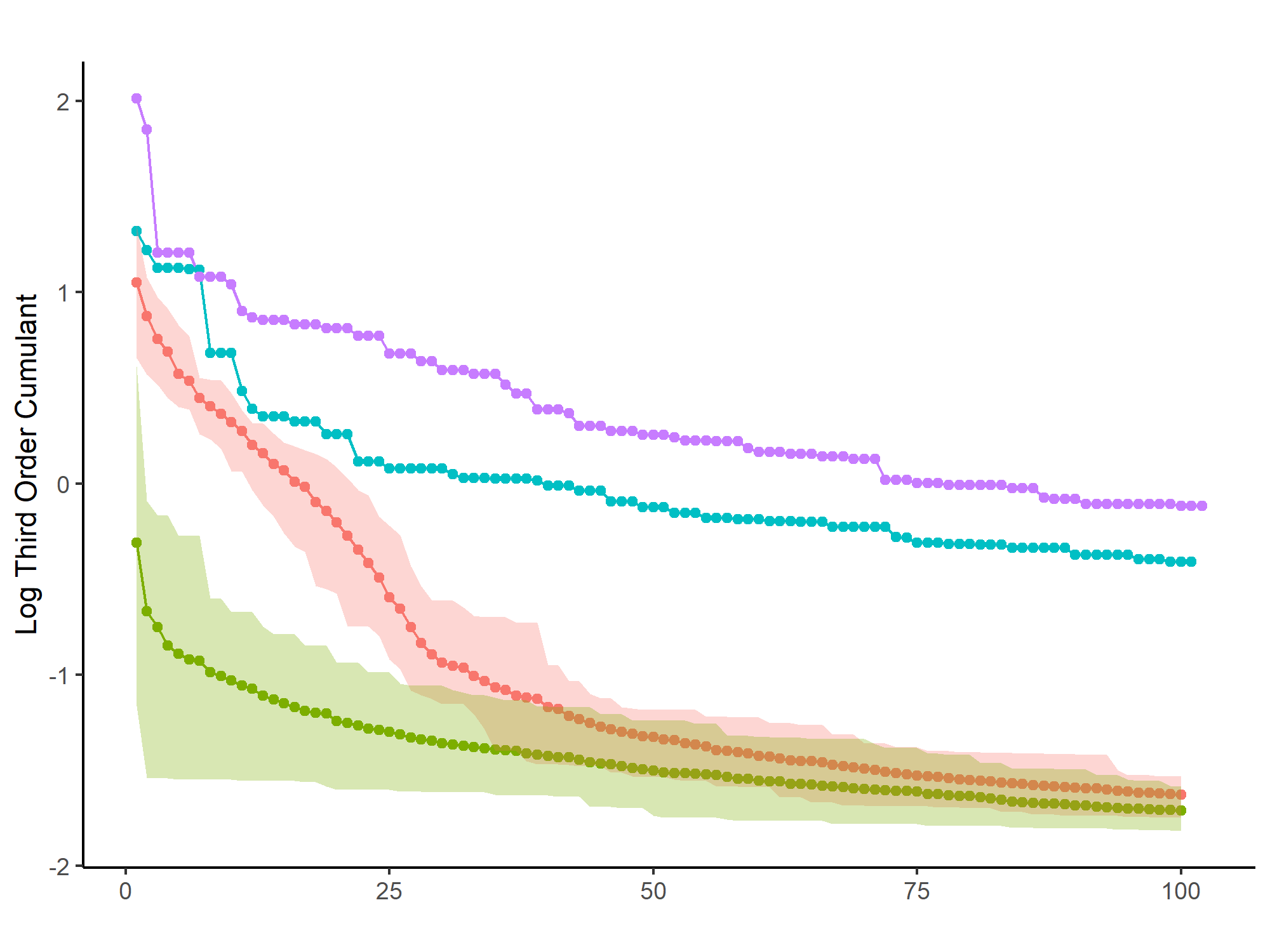} &
        \includegraphics[width=2.5in]{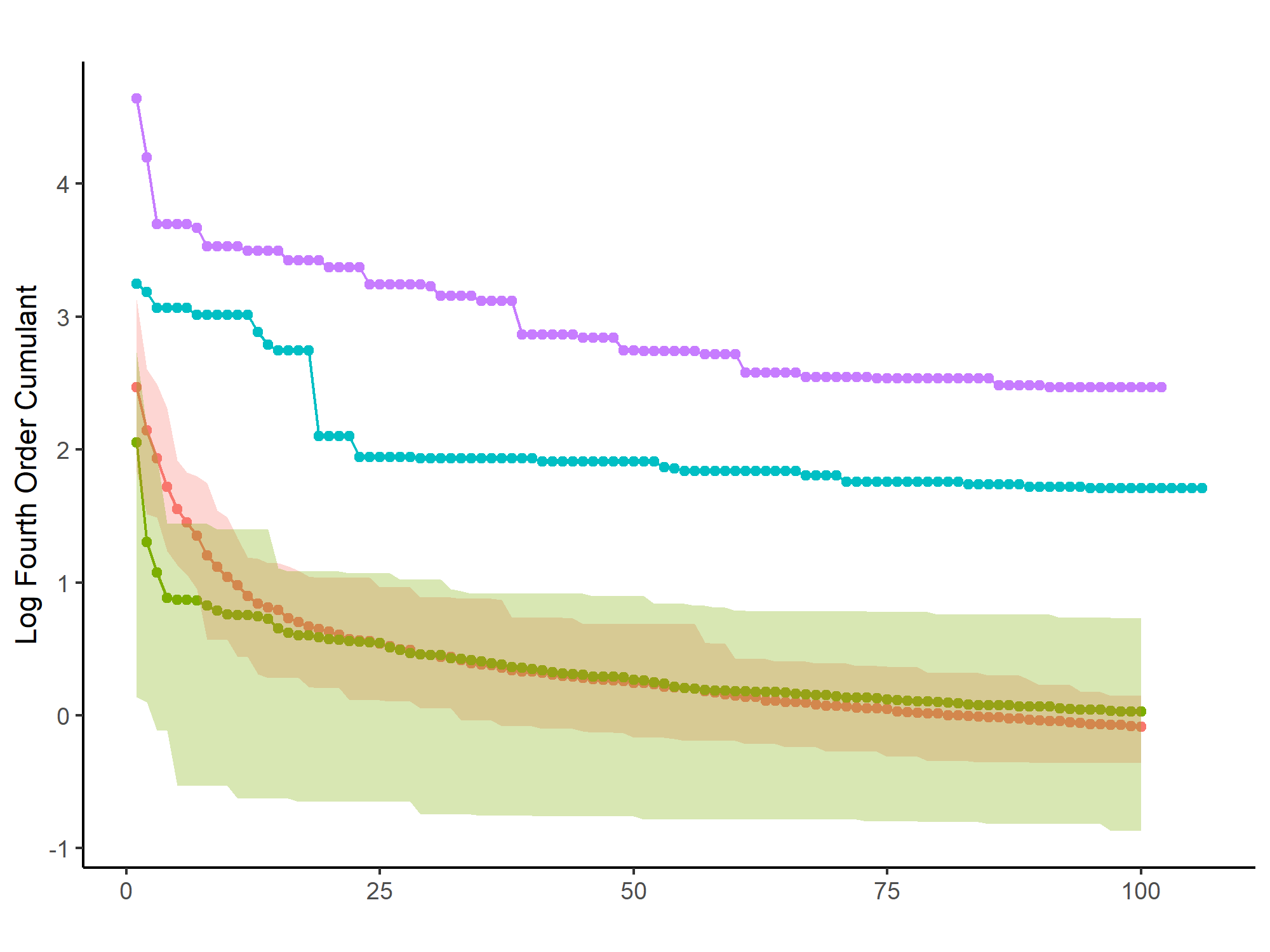} \\
        
        \rotatebox{90}{\hspace*{0.8in}\textbf{Credit}} &
        \includegraphics[width=2.5in]{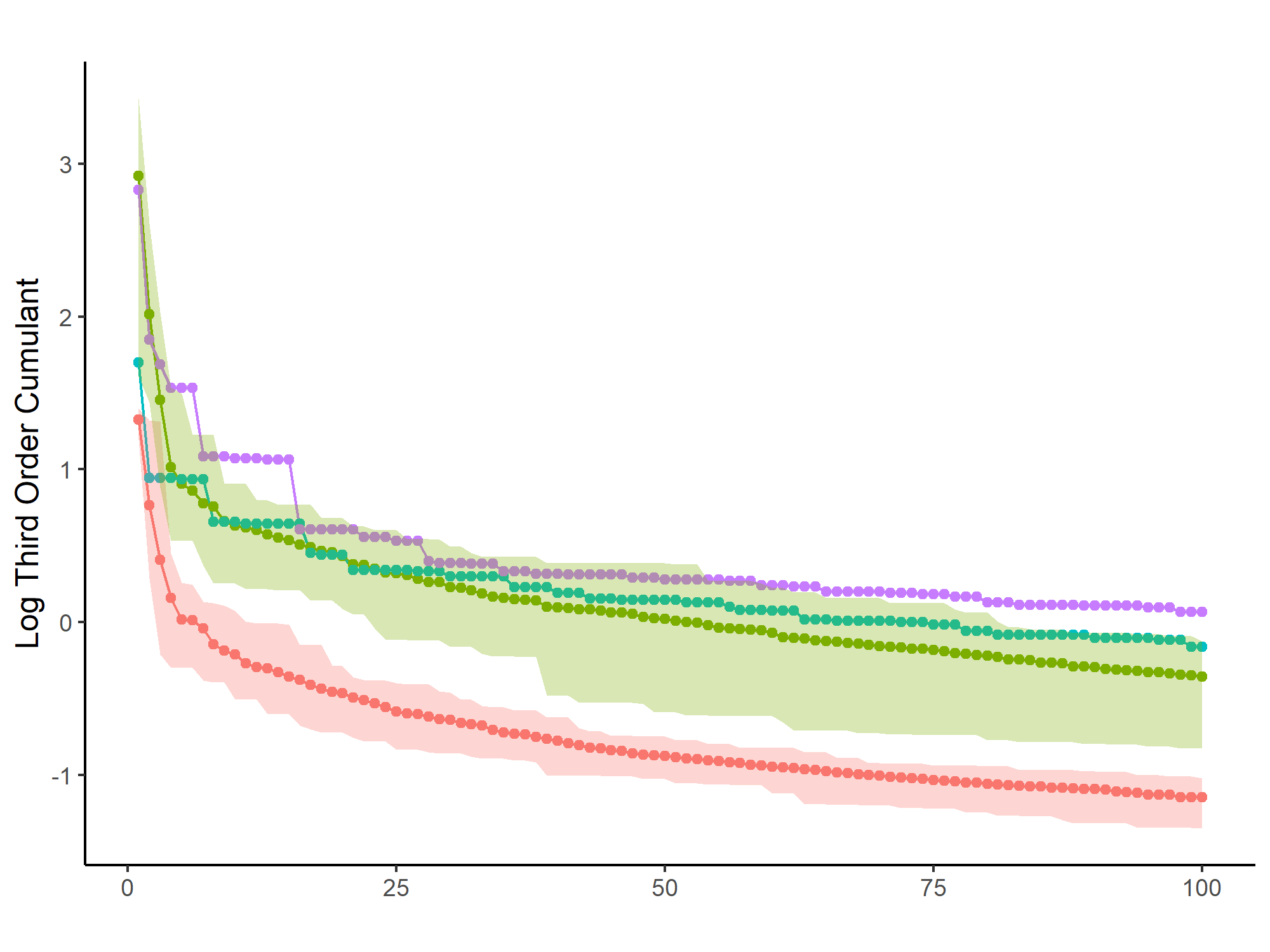} &
        \includegraphics[width=2.5in]{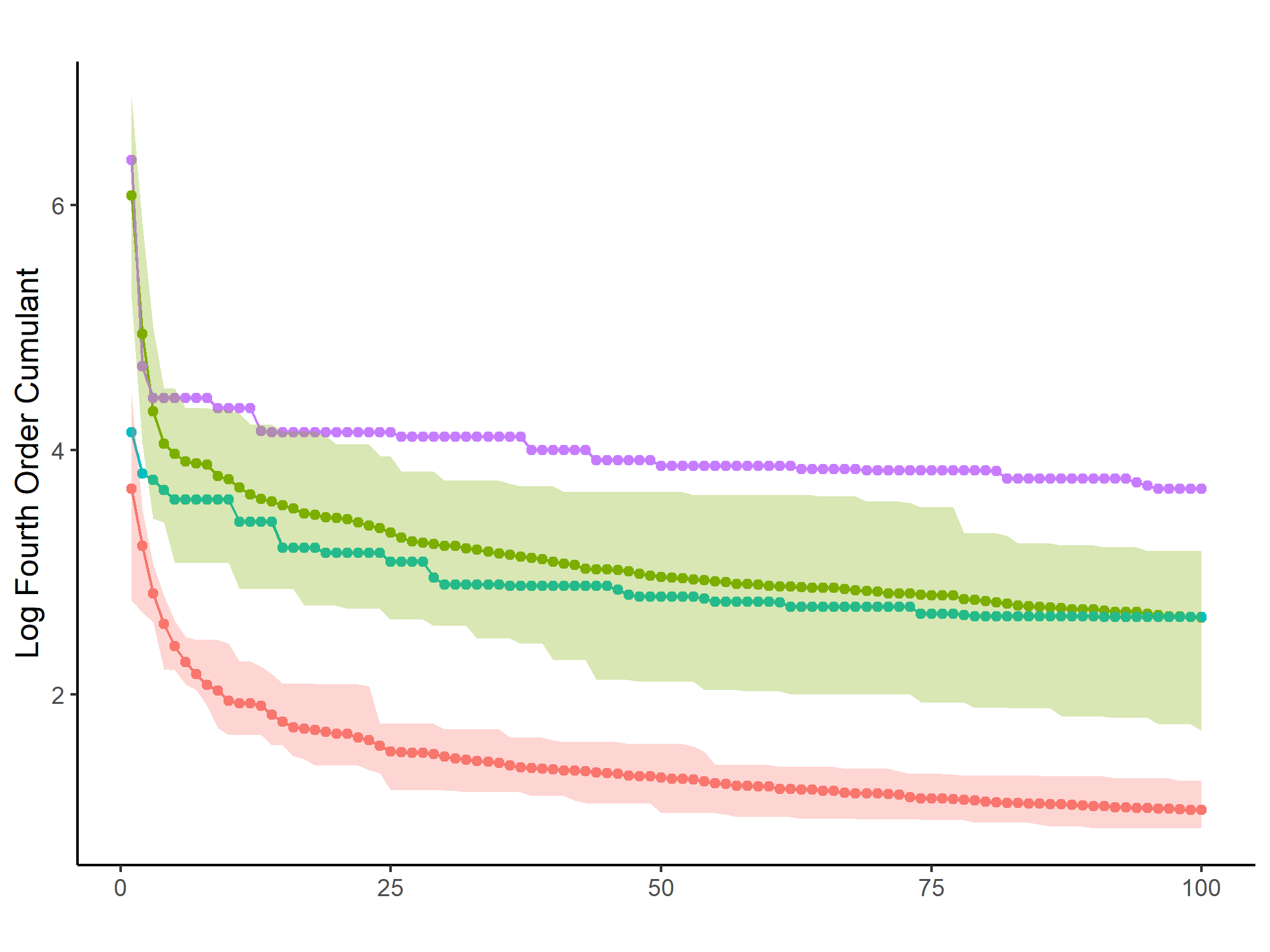}\\
    \end{tabular}
    \caption{{\bf Comparison of higher-order cumulants.} Plots of the largest $100$ third- and fourth-order cumulants from different combinations of models and datasets. Each row of panels represents a different dataset, and each column represents a different higher-order joint cumulant.
    For the synthetic data, the dots represent the average value over the $15$ runs of generating synthetic data for each model, and the shaded region denotes the min-max spread. In general, synthetically produced data fail to reproduce higher-order dependencies that result from real data.}
    \label{fig:cumulant_screes}
\end{figure}

We indeed found that both CTGAN and fine-tuned GPT-2 model were unable to generate synthetic data that replicated the dependency structure of real data.
A comparison of the $100$ largest third- and fourth-order cumulants between the real data and synthetically generated data shows that, for all three datasets, neither model consistently approximated the joint cumulants in the real data (Figure \ref{fig:cumulant_screes}). This held true across both the testing (blue dots) and training (purple dots) split of the real data. The only exceptions were that the fine-tuned GPT-2 could replicate the Credit data and CTGAN could replicate the Adult dataset. 

A comparison between the cumulants of the training and test split show that they are comparable for the Adult dataset and diverge for the other two dataset (green and red dots in Figure \ref{fig:cumulant_screes}). This is unexpected as two splits of the same dataset should yield the same set of cumulants. For the Breast Cancer dataset, we suspect that this high variance is partially due to the small number of samples. These small number of training samples may also partially explain the inability of CTGAN and fine-tuned GPT-2 to model these higher-order cumulants. On the other hand we speculate that the sparsity of the columns in the Credit dataset contributed to the diverging contrast between the cumulants of the train and test set.


In addition to comparing the largest cumulant values, we also investigated the rate that the same higher-order cumulant derived from synthetic and real data produced zeros values at the same time. In addition, we also investigated the rate of the opposite scenario where the same cumulant produces non-zero values across real and synthetic data. Both of these phenomenon occurring at a high rate would be indicators of a good quality synthetic dataset. For this analysis we defined two metrics. The first metric, the true positive rate (TPR), is the proportion of true positives in the synthetic data over the sum of true positives and true negatives. The true negative rate is the proportion of true negatives over the number of true negatives and false positives. A true positive occurs when the same joint cumulant in both the real and synthetic dataset is deemed to exceed zero via using the empirical rule of 2 standard deviations about the mean for all joint cumulants estimated. A true negative occurs when the same joint cumulant in both the real and synthetic dataset is deemed to equal zero. 

The number of true positives and true negatives is determined by identifying positive outcomes in the real data, evaluating the corresponding cumulants in the synthetic data, and recording whether they are deemed significant (true positive) or not (true negative). Similarly, the number of true negatives and false positives are computed by compiling the negative results in the real data and determining which of the corresponding cumulants are positive or negative in the synthetic data. 

Table \ref{table:cumulants} shows the true negative rate (TNR) and true positive rate (TPR) for whether the synthetic data generated by LLMs produces nonzero higher-order cumulants when the real data does. TPR was calculated by saving an index of the joint cumulants in the training data that were deemed to exceed zero via the use of the empirical rule. Then for the CTGAN and fine-tuned models, the same indices were used to assess whether those that were positive in the training data remained positive in the CTGAN and fine-tuned synthetic data. The analogous procedure was used to calculate the TNR.  We see that the TPR is moderately high for the adult data and low for Breast Cancer and Credit data. Overall, the TPR indicates that higher-order dependencies are not well replicated in the synthetic data, from either method. The TNR is also quite high for the adult data set, indicating that the negative results are replicated in synthetic CTGAN and fine-tuned data. The Breast Cancer and Credit synthetic data also maintained relatively high TNR values. In sum, it appears that the synthetic data methods as a whole fail to replicate the significant higher-order dependencies present in the training data. 

\begin{table}
    \centering
    \begin{tabular}{ccccccccccc}
        & & & \multicolumn{4}{c}{True Negative Rate} & \multicolumn{4}{c}{True Positive Rate}  \\ 
        Cumulant & Dataset & Method & N & Min. & Med. & Max. & N & Min. & Med. & Max.\\
        \hline
        \multirow{6}{*}{3} & \multirow{2}{*}{Adult} &  CTGAN &  1,719 & 1.00 & 1.00 & 1.00 & 9 & 0.56 & 0.78 & 1.00 \\
        & & GPT-2-FT & 1,719 & 1.00 & 1.00 & 1.00 & 9 & 0.00 & 0.56 & 0.67 \\
        & \multirow{2}{*}{Breast} & CTGAN  & 21,114 &  0.78 & 0.89 & 1.00 & 838 & 0.00 & 0.00 & 0.11 \\
        & & GPT-2-FT & 21,114 &  0.67 & 0.89 & 1.00 & 838 & 0.00 & 0.11 & 0.33 \\
        & \multirow{2}{*}{Credit} & CTGAN & 25,950& 0.56 & 1.00 & 1.00 & 1,050 & 0.11 & 0.33 & 0.56 \\
        &  & GPT-2-FT & 25,950& 0.89 & 1.00 & 1.00 & 1,050 & 0.00 & 0.22 & 0.33 \\
        \hline 
        \multirow{6}{*}{4} & \multirow{2}{*}{Adult} & CTGAN & 20,715 & 1.00 & 1.00 & 1.00 & 21 & 0.78 & 0.89 & 1.00 \\
        & & GPT-2-FT & 20,715 & 1.00 & 1.00 & 1.00 & 21 & 0.44 & 0.75 & 1.00 \\
        & \multirow{2}{*}{Breast} & CTGAN & 599,497 & 0.89 & 1.00 & 1.00 & 15,159 & 0.00 & 0.00 & 0.22 \\
        & & GPT-2-FT & 599,497 & 0.33 & 0.67 & 1.00 & 15,159 & 0.11 & 0.33 & 0.56 \\
        & \multirow{2}{*}{Credit} & CTGAN & 787,215 & 0.44 & 0.78 & 0.89 & 22,785 & 0.11 & 0.33 & 0.67 \\
        & & GPT-2-FT &787,215 & 0.78 & 1.00 & 1.00 & 22,785 & 0.00 & 0.11 & 0.44 \\
        \hline
    \end{tabular}
    \caption{{\bf Rate that higher-order cumulants are both zero or non-zero across real and synthetic data.} The true positive rate represents whether a cumulant in the synthetic data is non-zero whenever it is so in the train. The true negative rate represents whether a cumulant in the synthetic data is zero when it is zero in the train set. The minimum, median, and maximum are taken over the $15$ independent runs of fine-tuning GPT-2 on the adult data.}
    \label{table:cumulants}
\end{table}

Table \ref{table:cumulants} shows the true negative rate (TNR) and true positive rate (TPR) for whether the synthetic data generated by LLMs produces nonzero higher-order cumulants when the real data does. TPR was calculated by saving an index of the joint cumulants in the training data that were deemed to exceed 0 via the use of the empirical rule, eg 2 standard deviations about the mean for all joint cumulants estimated. Then for the CTGAN and fine-tuned models, the same indices were used to assess whether those that were positive in the training data remained positive in the CTGAN and fine-tuned synthetic data. The analogous procedure was used to calculate the TNR.  We see that the TPR is moderately high for the adult data and low for Breast Cancer and Credit data. Overall, the TPR indicates that higher-order dependencies are not well replicated in the synthetic data, from either method. The TNR is also quite high for the adult data set, indicating that the negative results are replicated in synthetic CTGAN and fine-tuned data. The Breast Cancer and Credit synthetic data also maintained relatively high TNR values. In sum, it appears that the synthetic data methods as a whole fail to replicate the significant higher-order dependencies present in the training data. 

\section{Discussion}

In this work, we describe a new methodology that can be used to determine whether synthetically-generated tabular data is similar to real data that it was derived from. Our approach improves upon the current approach (i.e, train-synthetic-test-real approach) because it directly compares values derived from real and synthetic datasets instead of comparing whether models trained on real and synthetic data can obtain the same performance. Since our methodology is based off assessing statistical characteristics of datasets, our approach is mathematically rigorous and is agnostic to how the synthetic dataset was generated or what downstream tasks are conducted on the dataset.

Another benefit of our approach is that it can be used to determine good quality synthetic tabular data has been generated and can inform properties in need of improvement and refinement. Specifically, it can identify specific pairwise and higher-order associations that differ between the datasets and describe univariate statistics that differ as well. As a result, this approach can be used to improve the generation of synthetic data even for cases involving the simulation of highly complex and dependent variates. For example, during an initial assessment of the Adult dataset we found nominal variables with fairly high cardinality (for example, occupation). This would traditionally be handled using label or one-hot encoders prior to synthetic data generation. More work should be done to assess what encoders best enable synthetic data models to represent the complex dependency structures of these high-cardinality nominal features that are so common in tabular data.

Overall, our evaluations suggest that LLMs and GANs are unable to consistently generate high-quality synthetic tabular datasets. For example, we found that a few-shot prompted LLM model failed at reproducing simpler pairwise dependencies and a fine-tuned LLM model had mixed results modeling both these pairwise and more complex, higher-order dependencies.
On the other hand, our analyses unsurprisingly suggested that models can be improved with fine-tuning. However, improvement of these models is predicted on a large set of training data that can be used for fine-tuning, as suggested by the performance boost the fine-tuned GPT-2 model obtained on the Adult dataset compared to the Breast Cancer and Credit dataset.
This has implications for fields, such as the biological sciences, where data is often limited, and therefore there is a need to generate synthetic data.

Another challenge associated with using models to generate synthetic data is that they can create malformed and nonsensical examples. For example, a model could generate an example with the incorrect number of columns or generate an example that has an  alphabetical string for numerical column. In some cases, we found that off-the-shelf LLMs were unable to generate valid examples a majority of the time (Supplementary Table \ref{table:cleaned_counts}). For example, the few-shot prompted GPT-2 model was unable to generate valid examples, across 15 trials, ~63\% of the time for the Credit dataset. On the other hand, this model was unable to generate valid examples ~34.8\% and ~1.3\% of the time for the Adult and Breast Cancer dataset, respectively. This inability to  consistently produce valid samples across datasets provides further evidence for not using off-the-shelf LLMs for synthetic data generation.

It is increasingly common to use machine learning-generated synthetic data to augment training data to reduce bias and to increase the size of the training set \cite{naaz2021generative,svabensky2024evaluating, jaipuria2020deflating,kang2024synthetic,kwon2021increasing, jordon2022synthetic}.
In addition, models are also increasingly used to create differentially private synthetic data in fields such as health care \cite{jordon2022synthetic,tran2024differentially,tornqvist2024text, giuffre2023harnessing,qian2024synthetic}. Our results suggest utilizing synthetic data generated by LLMs and GANs in subsequent analyses increases bias and may lead to invalid results since the generated synthetic data is not high-quality. 
Therefore, we suggest that researchers avoid utilizing synthetic datasets for model training or other analyses.

\section{Acknowledgments}

This research was supported by the Laboratory Directed Research and Development Program at Pacific Northwest National Laboratory, a multiprogram national laboratory operated by Battelle for the U.S. Department of Energy.
We would also like to thank PNNL Research Computing for access to computational resources. Pacific Northwest National Laboratory is a multiprogram national laboratory operated by Battelle Memorial Institute for the United States Department of Energy under contract DE-AC06-76RLO.

\bibliography{main}
\end{document}


\maketitle

\section{Methods}

\subsection{Experimental Evaluation}
To compare and contrast the structure and statistical properties of the real and synthetic (GAN and LLM) data, we took the following steps:

\begin{enumerate}

\item Load and Preprocess Data: Real and synthetic datasets were loaded from CSV files and co-linear columns were removed. Low-cardinality categorical features were hot encoded to convert them into a format suitable for modeling.

\item Feature Association Analysis: Association calculations were made between variables within and across the real and synthetic data. Since the data is a mix of nominal, ordinal, and numerical values and the numerical values are not real-valued measures, but integers, we chose to summarize the similarity between variables by extracting a Cramer's V effect size measure for all pairwise associations among the $13$ features in the data as described in the above section.  

\item Reduce the dimensionality of real and synthetic data and calculate similarity measures: Using PCA, FAMD, and UMAP algorithms, we depicted scores from the dimension reduction algorithms to compare real and synthetic data. To compare the lower-dimensional representations of the real and synthetic data, we computed confidence regions of the projections and computed overlap. 

\item Fit Maximum Likelihood Estimated Logistic Regression Models: Logistic regression models were fitted to real and synthetic data. Parameters and confidence intervals were extracted to compare feature impact between the real and synthetic data. Percentage overlap of confidence intervals provides a numerical measure of similarity. 

\item Estimate dependency networks for real and synthetic data and make comparisons: Network models summarize the dependency structure and visually depict differences across the real and synthetic data. Summaries of graph model metrics such as degree, eigencentrality, density, and diameter were also recorded. Community detection was performed for the real and synthetic data and comparisons of variable communities were made. 

\item Compute higher order dependencies for real and synthetic datasets: Using joint cumulants and standardized mean square errors of orders up to 4, we compared the dependency structure in a more nuanced yet descriptive manner. Scree plots depict differences in the higher-order dependencies for the top 100 joint cumulants, both third and fourth-order.   
\end{enumerate}

\subsection{Louvain Community Detection}
\label{more_louvain}

The Louvain algorithm has the following steps:
\begin{enumerate}
    \item  Initialize the community by setting each node as a separate community
    \item Connect to a single node (node 1) and calculate the change of modularity if node 2 is assigned to each neighbor community. Node 1 is assigned to the community that maximizes modularity.
    \item Iterate over remaining nodes using step 2 until moving nodes does not increase modularity.
    \item  Using the communities of step 3, merge these communities in the fashion of step 2 until all nodes are merged into one community. The arrangement of these merged communities that maximize modularity is reported as the final community structure.
\end{enumerate}

\begin{figure}
    \centering
    \begin{textbox}
    Produce \{number of samples to produce\} data samples which mirrors the given examples 
    
    example data: \\
    \{data\}
    \\
    The output should be a comma-separated value (CSV) table with the following columns:  
    \\
    \{dataset columns separated by commas\}
    \end{textbox}
    \caption{{\bf Few-shot prompt}. Prompt used for few-shot prompting of LLMs.}
    \label{fig:zero_shot_prompt}
\end{figure}

\begin{figure}
    \centering
    \begin{textbox}
    Produce 31655 data samples which mirrors the given examples \\
    example data: \\
    age,workclass,fnlwgt,education,education-num,marital-status,occupation,relationship,race,sex,capital-gain,capital-loss,hours-per-week,native-country,income \\
    32, Private,347623, HS-grad,9, Married-civ-spouse, Machine-op-inspct, Husband, White, Male,0,0,40, United-States, $\leq50$K \\
    39, Private,247733, HS-grad,9, Divorced, Priv-house-serv, Unmarried, Black, Female,0,0,16, United-States, $\leq50$K \\
    27, Private,110931, HS-grad,9, Married-civ-spouse, Adm-clerical, Husband, White, Male,0,0,40, United-States, $\leq50$K \\
    62, State-gov,202056, Bachelors,13, Divorced, Prof-specialty, Not-in-family, White, Male,14084,0,40, United-States, $\geq50$K \\
    24, Private,85088, HS-grad,9, Never-married, Sales, Own-child, White, Female,0,1762,32, United-States, $\leq50$K \\
    38, Private,52187, Bachelors,13, Married-civ-spouse, Sales, Husband, White, Male,15024,0,50, United-States, $\geq50$K \\
    53, Local-gov,164300, Bachelors,13, Divorced, Adm-clerical, Unmarried, White, Female,0,0,38, Dominican-Republic, $\leq50$K \\
    38, Private,254114, Some-college,10, Married-spouse-absent, Prof-specialty, Own-child, Black, Female,0,0,40, United-States, <=50K \\
    31, Private,228873, HS-grad,9, Married-civ-spouse, Craft-repair, Husband, White, Male,0,0,40, United-States, $\leq50$K \\
    25, Local-gov,270379, HS-grad,9, Never-married, Adm-clerical, Not-in-family, Black, Female,0,0,40, United-States, $\leq50$K

    The output should be a comma-separated value (CSV) table with the following columns:
    age, workclass, fnlwgt, education, education-num, marital-status, occupation, relationship, race, sex, capital-gain, capital-loss, hours-per-week, native-country, income
    \end{textbox}
    \caption{{\bf Example of a few-shot prompt.} Few-shot prompt with example data from the Adults dataset.}
    \label{fig:zero_shot_prompt_example}
\end{figure}

\begin{table}
\centering
    \begin{tabular}{ccccc} 
        \hline
        Parameter & Adults & Breast & Credit\\  
        \hline
        Examples per prompt & $10$ & $3$ & $2$ \\
        Samples generated &  $31655$ & $398$ & $199364$ \\
        \hline
    \end{tabular}
    \caption{{\bf Number of samples generated by GPT-2.} A table of the total number of samples generated by GPT-2 by dataset. We asked GPT-2 to generate the same number of samples as the training dataset used in LLM fine-tuning experiments so that the sampled sets were comparable. Number of exampes from real dataset per prompt. And total number of samples generated}
    \label{table:zeroshot_examples_samples}
\end{table}

\begin{table}
    \centering
    \begin{tabular}{cccc} 
        \hline
        Parameter & Adults & Breast & Credit \\ 
        \hline
        Epochs & $300$ & $300$ & $50$ \\
        Batch size & $180$ & $50$ & $30$ \\
        \hline
    \end{tabular}
    \caption{Fine-tuning hyperparameters. These correspond to the arguments to the Huggingface Trainer class. The values listed here are varied with the dataset the model was trained on.}
    \label{table:GPT_2_ft_ds_hp}
\end{table}

\begin{table}
    \centering
    \begin{tabular}{cc} 
        \hline
        Parameter & Value \\
        \hline
        learning rate & $5\times10^{-5}$ \\
        learning rate scheduler & linear \\
        bf16 & True \\
        bf16 full eval & True \\
        \hline
    \end{tabular}
    \caption{{\bf Fine-tuning hyperparameters.} These correspond to the arguments to the Huggingface Trainer class. The values listed here are were constant over across the datasets.}
    \label{table:GPT-2_ft_hyperparams}
\end{table}

\begin{table}
\centering
    \begin{tabular}{cc} 
    \hline
    Parameter & Value\\ [0.5ex] 
    \hline
    discriminator dimension & $(256, 256)$ \\
    discriminator steps & $1$ \\
    embedding dimension & $128$ \\
    enforce min and max values & True \\
    enforce rounding & True \\
    epochs & $300$ \\
    generator dimension & $(256, 256)$ \\
    gen/disc learning rate & $2\times10{-4}$ \\
    gen/disc weight decay & $10^{-6}$ \\
    gen/disc adam betas & $(0.5, 0.99)$ \\
    log frequency sampling & True \\
    pack size & $10$ \\
    \hline
    \end{tabular}
    \caption{CTGAN Training Hyperparameters}
    \label{table:gan_hyperparams}
\end{table}

\begin{figure}
    \centering
    \includegraphics[width=0.49\linewidth]{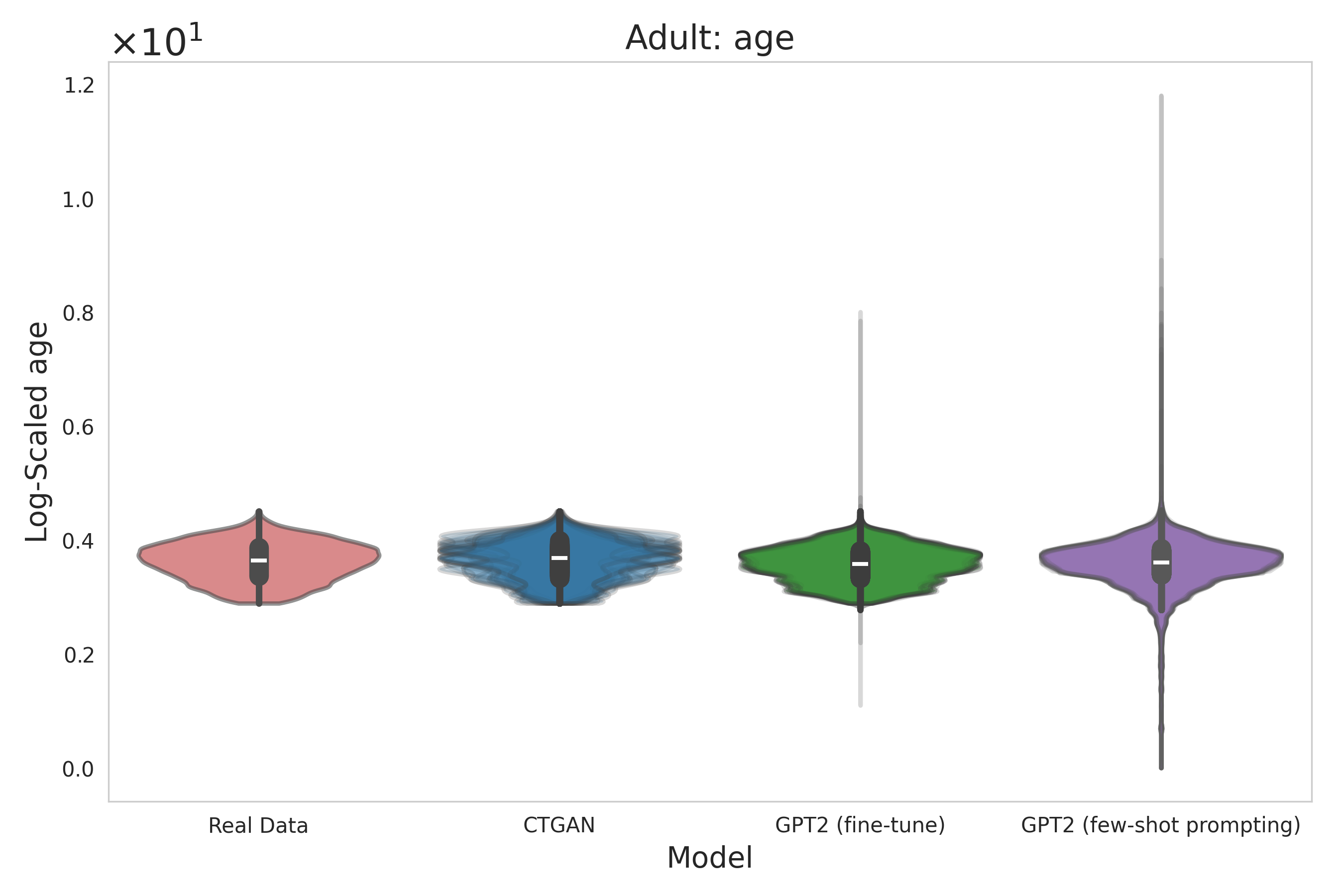}
    \includegraphics[width=0.49\linewidth]{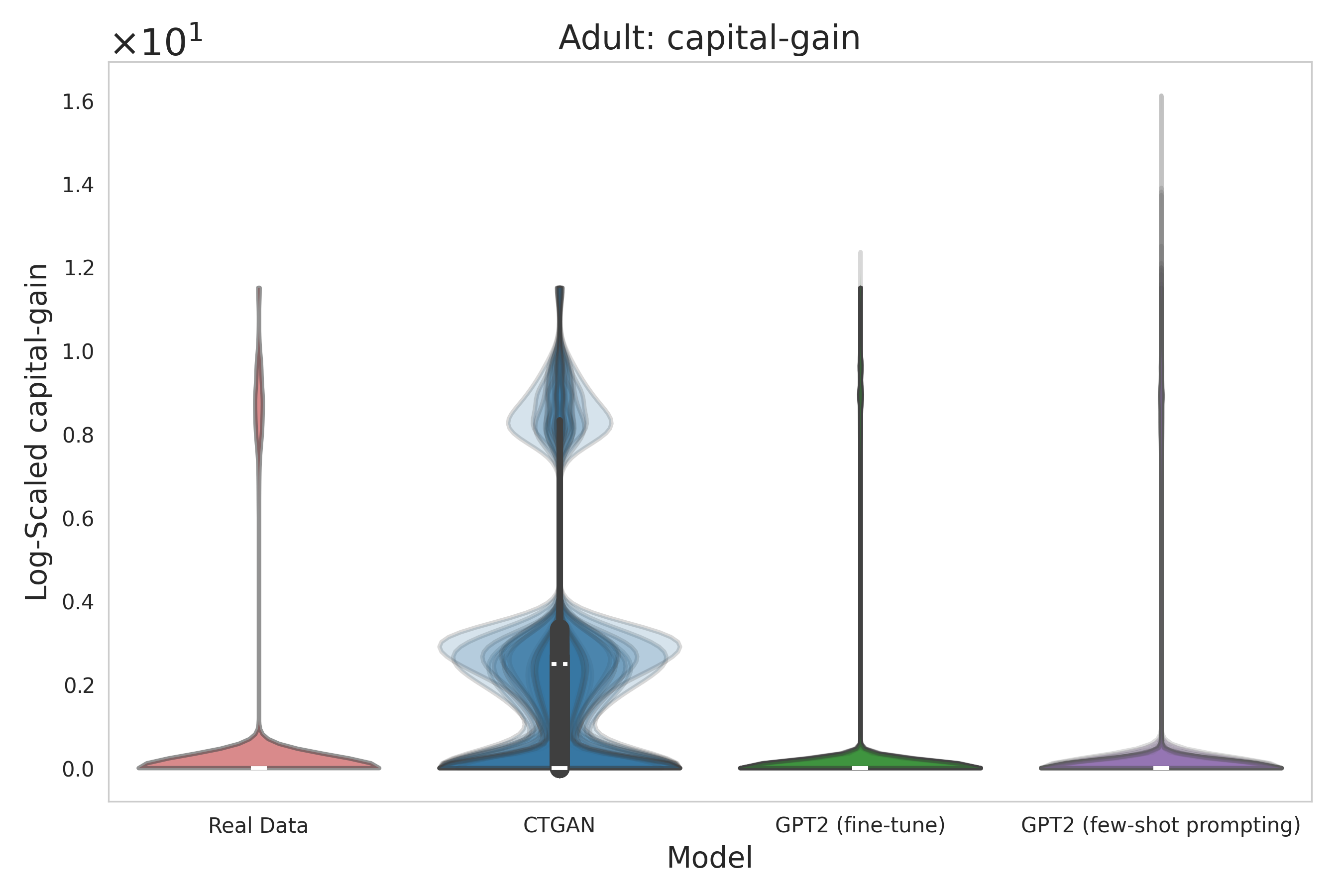}\\
    \includegraphics[width=0.49\linewidth]{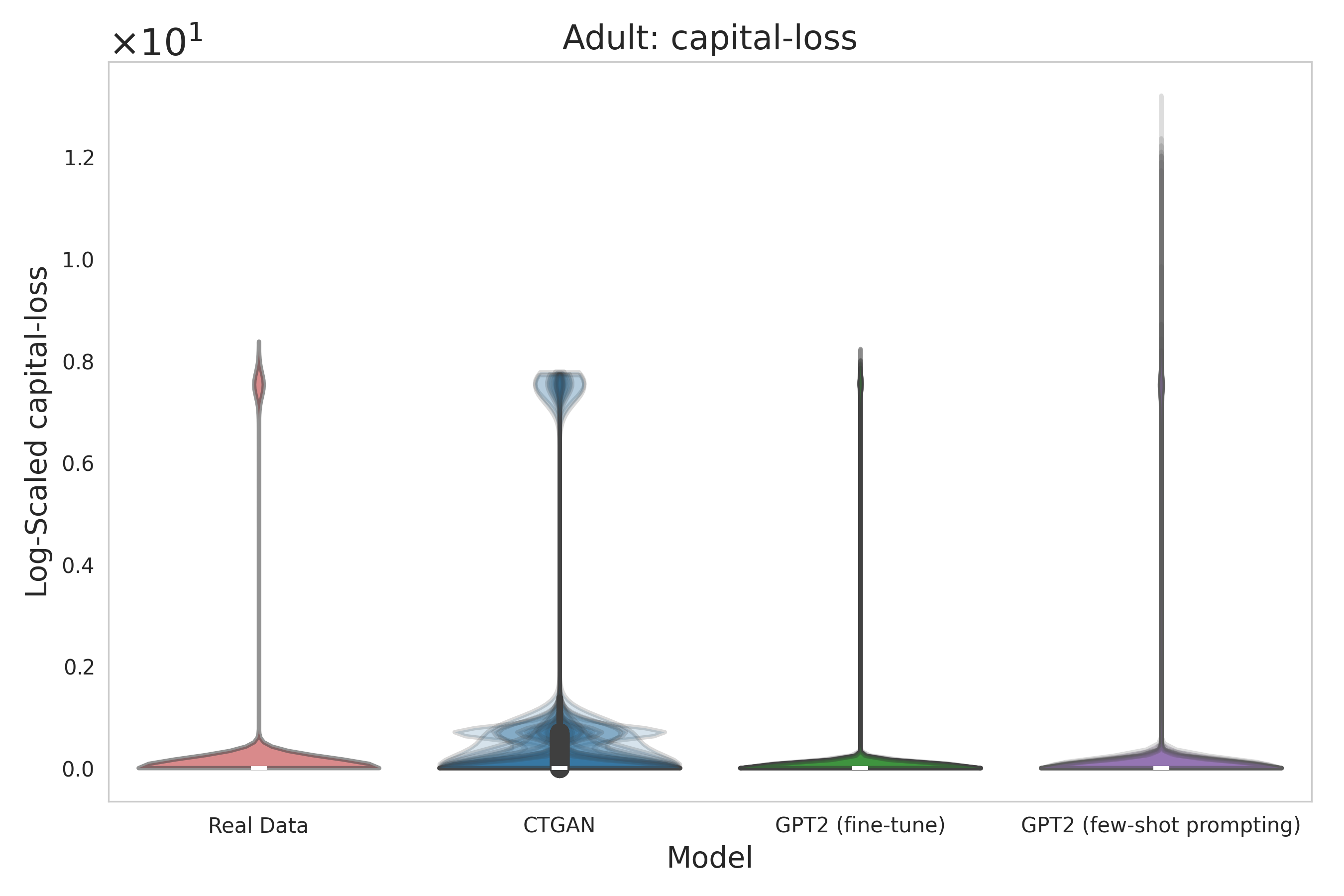}
    \includegraphics[width=0.49\linewidth]{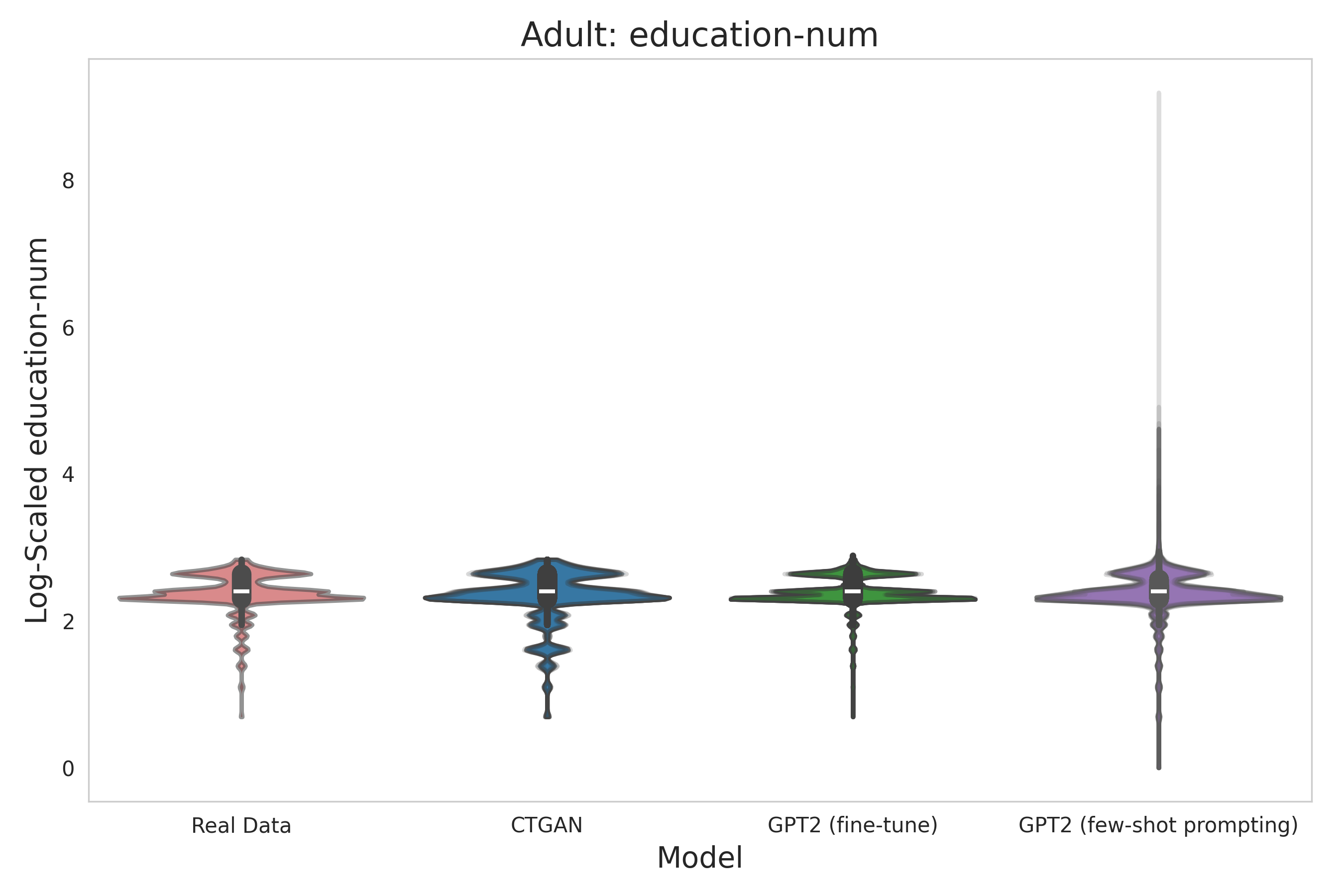}\\
    \includegraphics[width=0.49\linewidth]{figures/violins/adult/adult_hours-per-week.png}\\
    \caption{Violin plots show the distributions of continuous columns in the Adult dataset.}
    \label{fig:adult_violins}
\end{figure}

\begin{figure}
    \centering
    \includegraphics[width=0.49\linewidth]{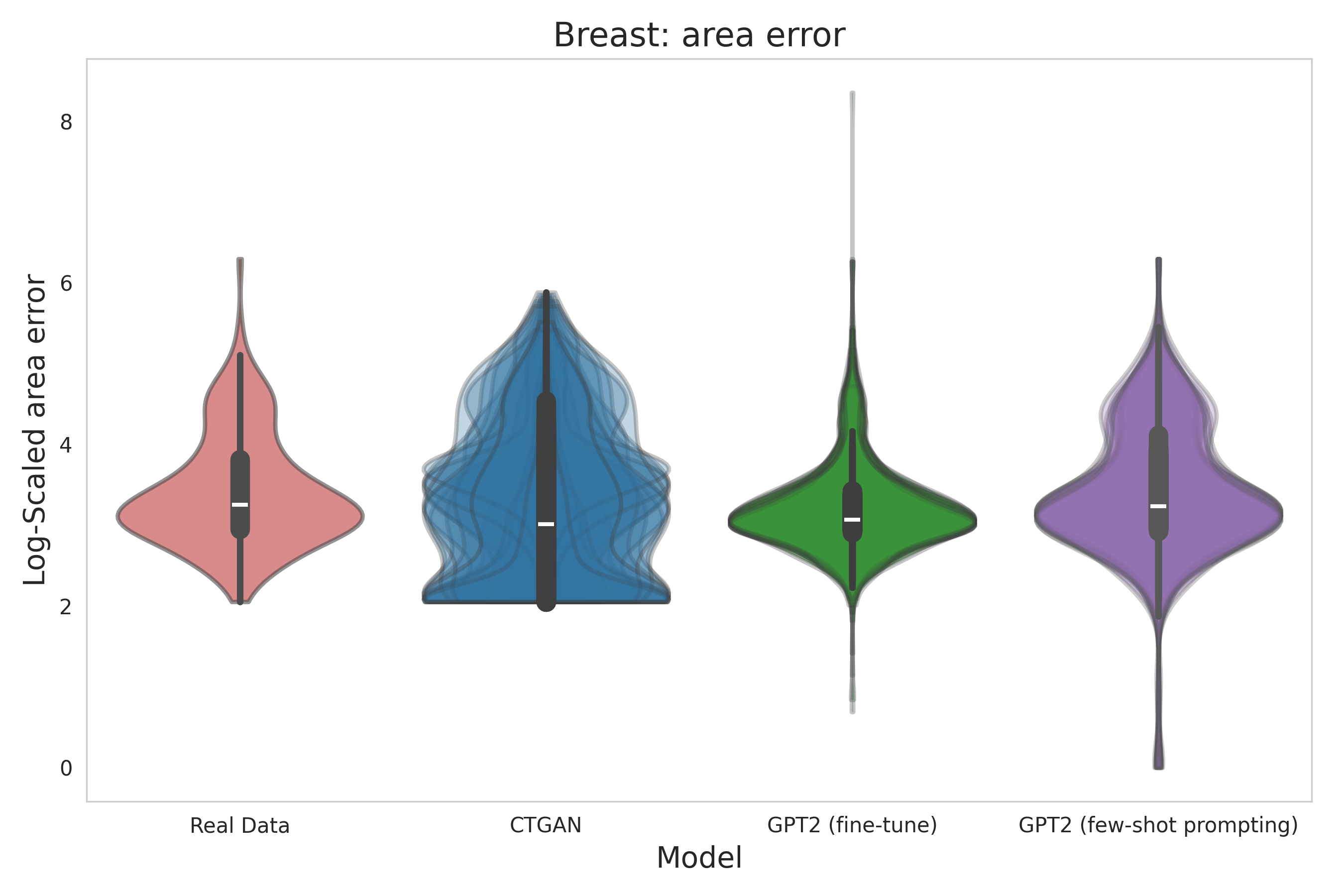}
    \includegraphics[width=0.49\linewidth]{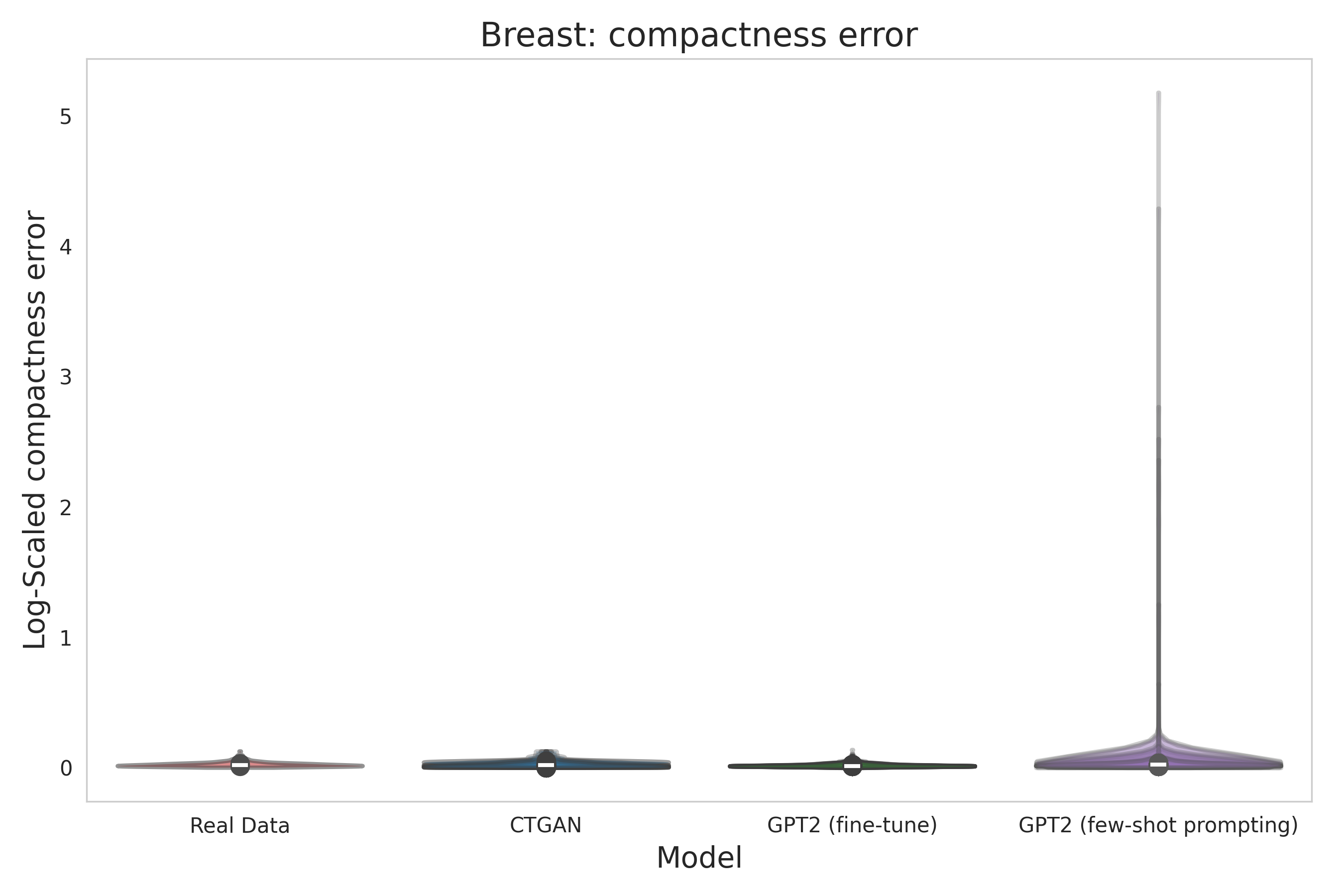}\\
    \includegraphics[width=0.49\linewidth]{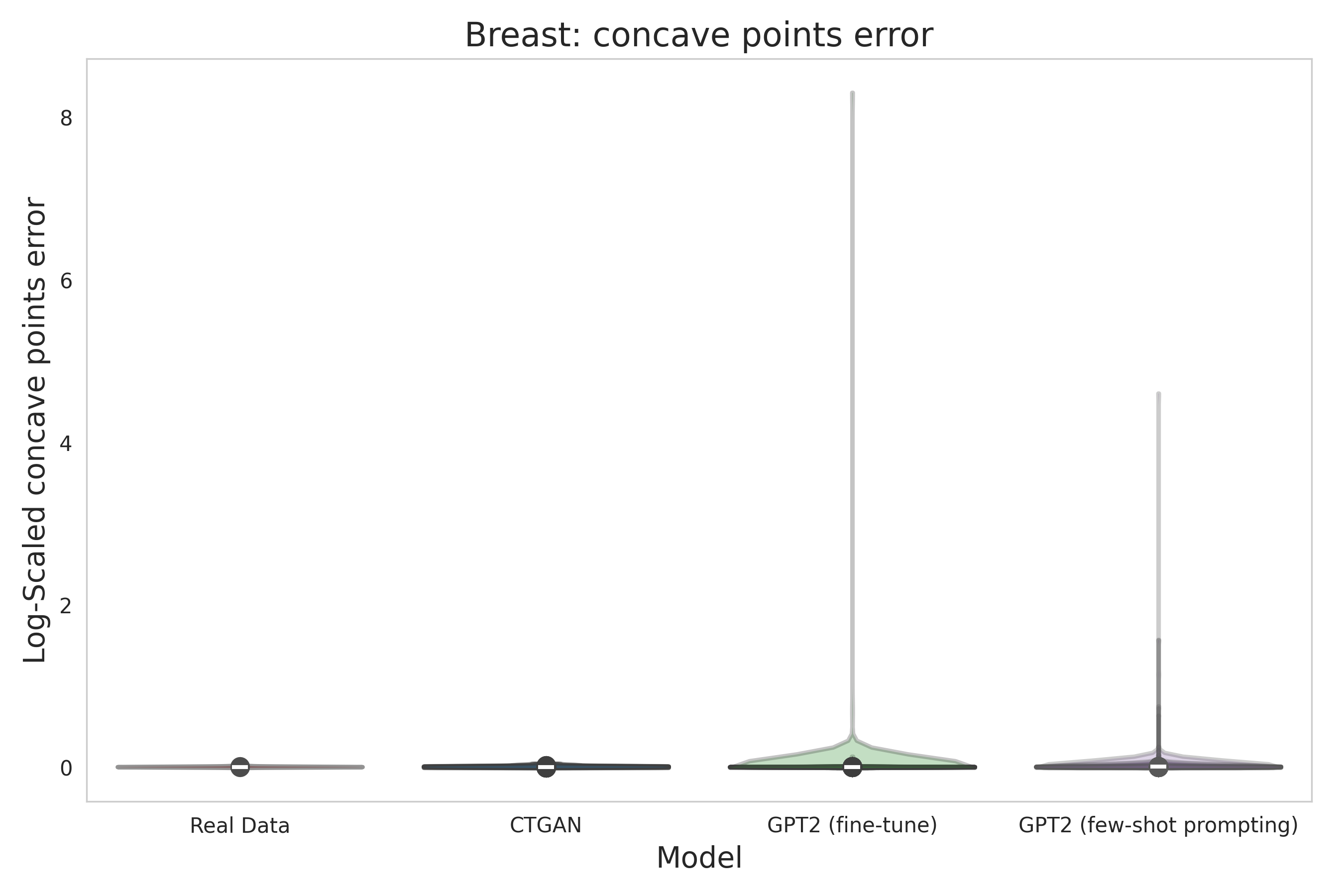}
    \includegraphics[width=0.49\linewidth]{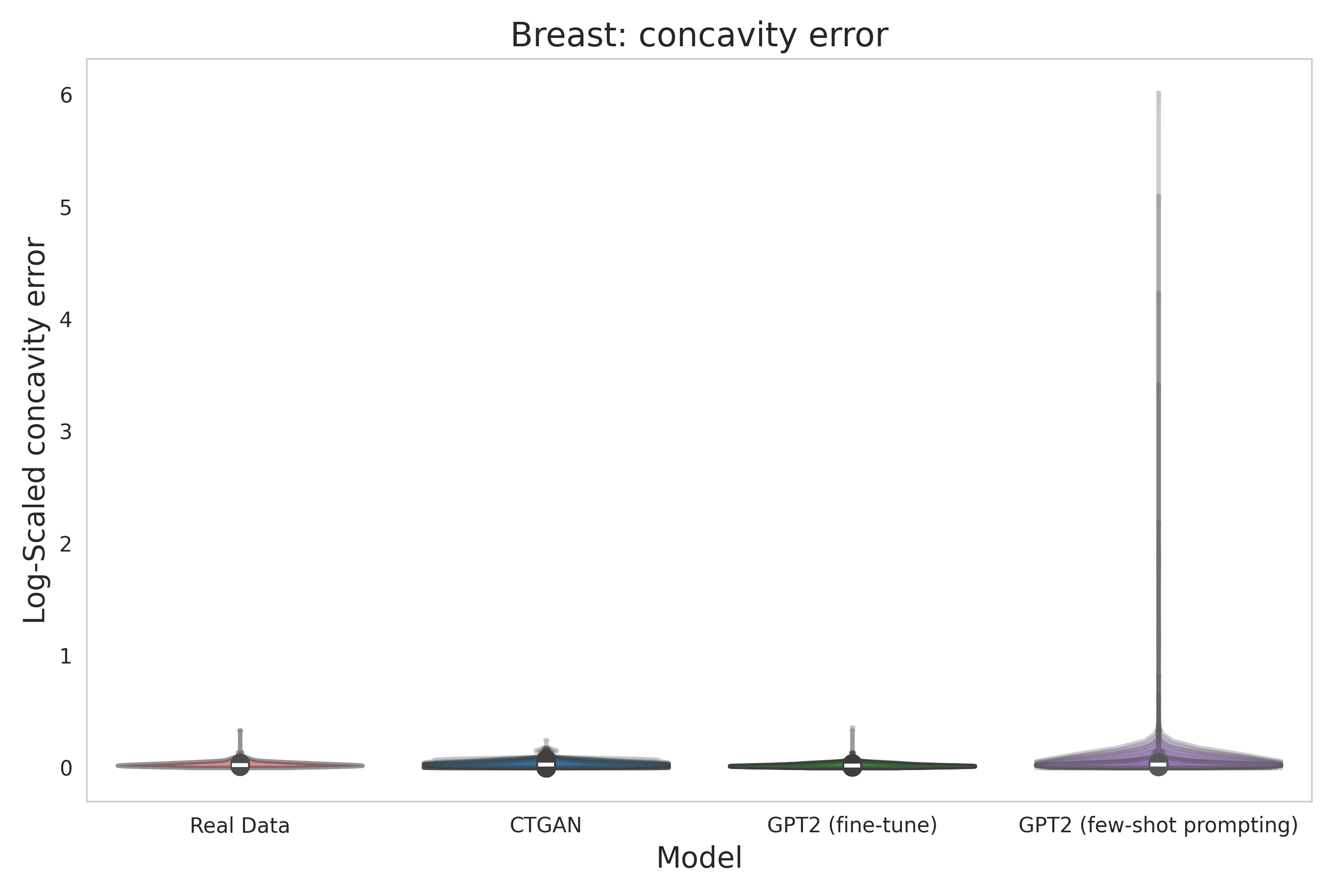}\\
    \includegraphics[width=0.49\linewidth]{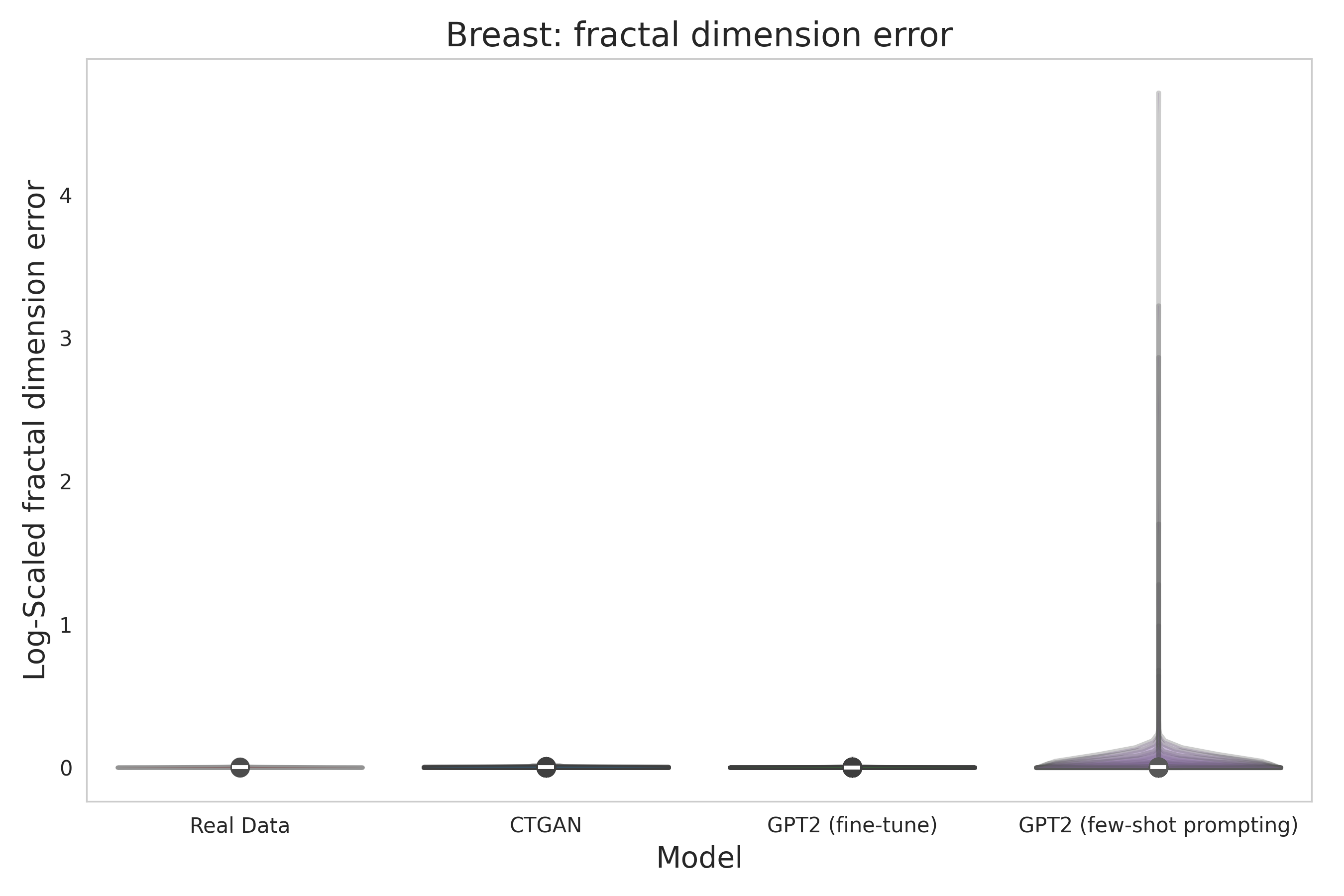}
    \includegraphics[width=0.49\linewidth]{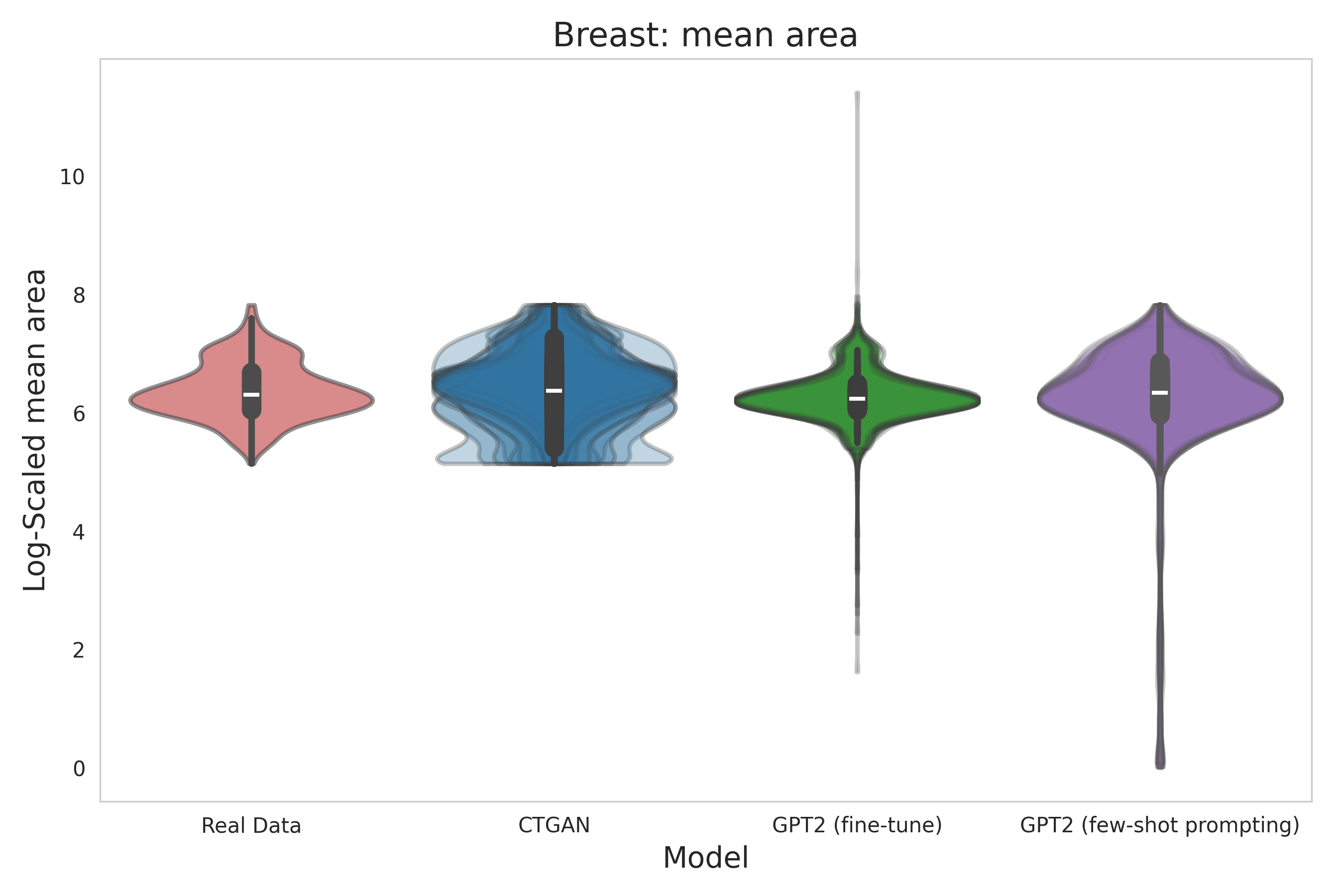}\\
    \caption{Violin plots show the distributions of continuous columns in the Breast dataset.}
    \label{fig:breast1_violins}
\end{figure}

\begin{figure}
    \centering
    \includegraphics[width=0.49\linewidth]{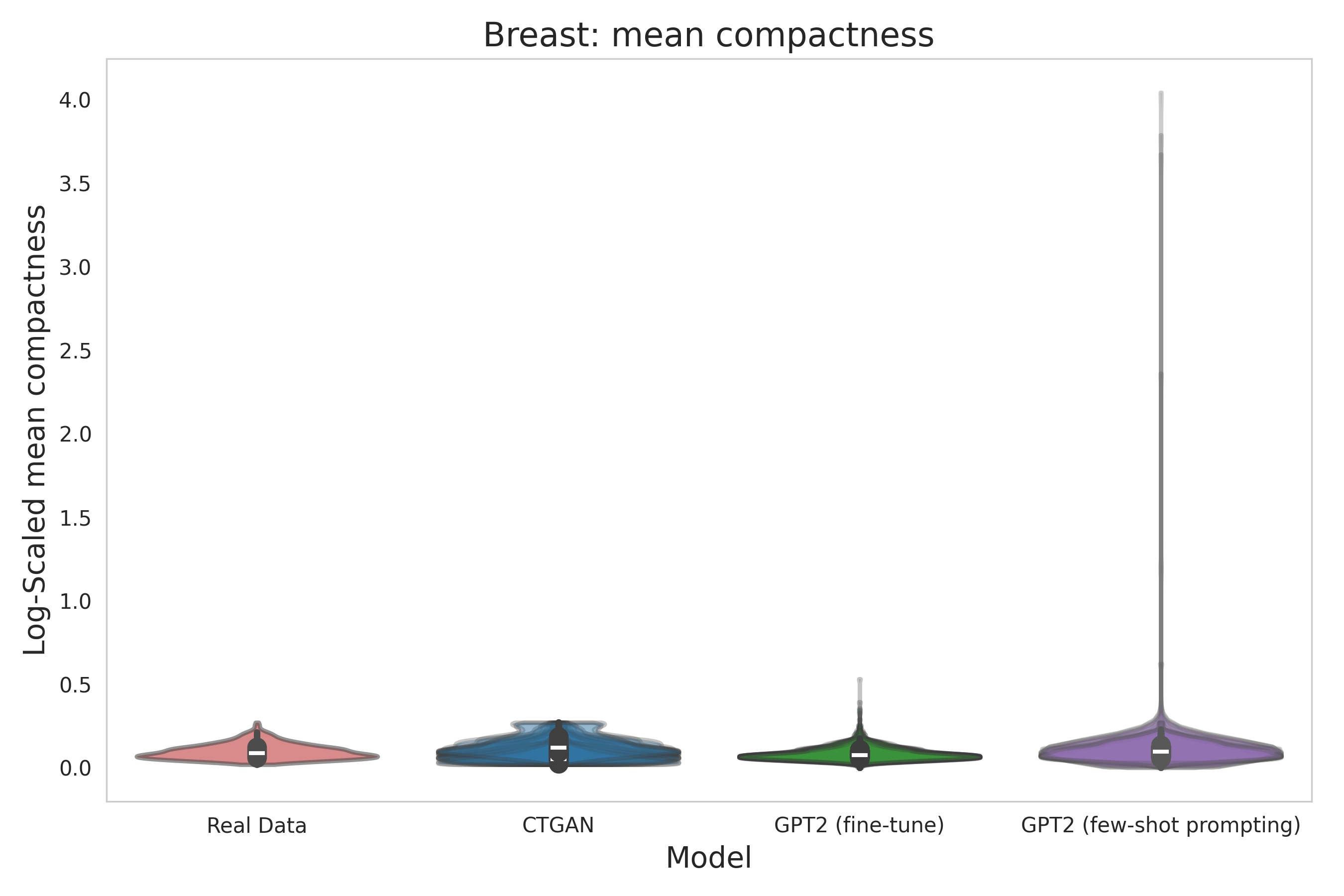}
    \includegraphics[width=0.49\linewidth]{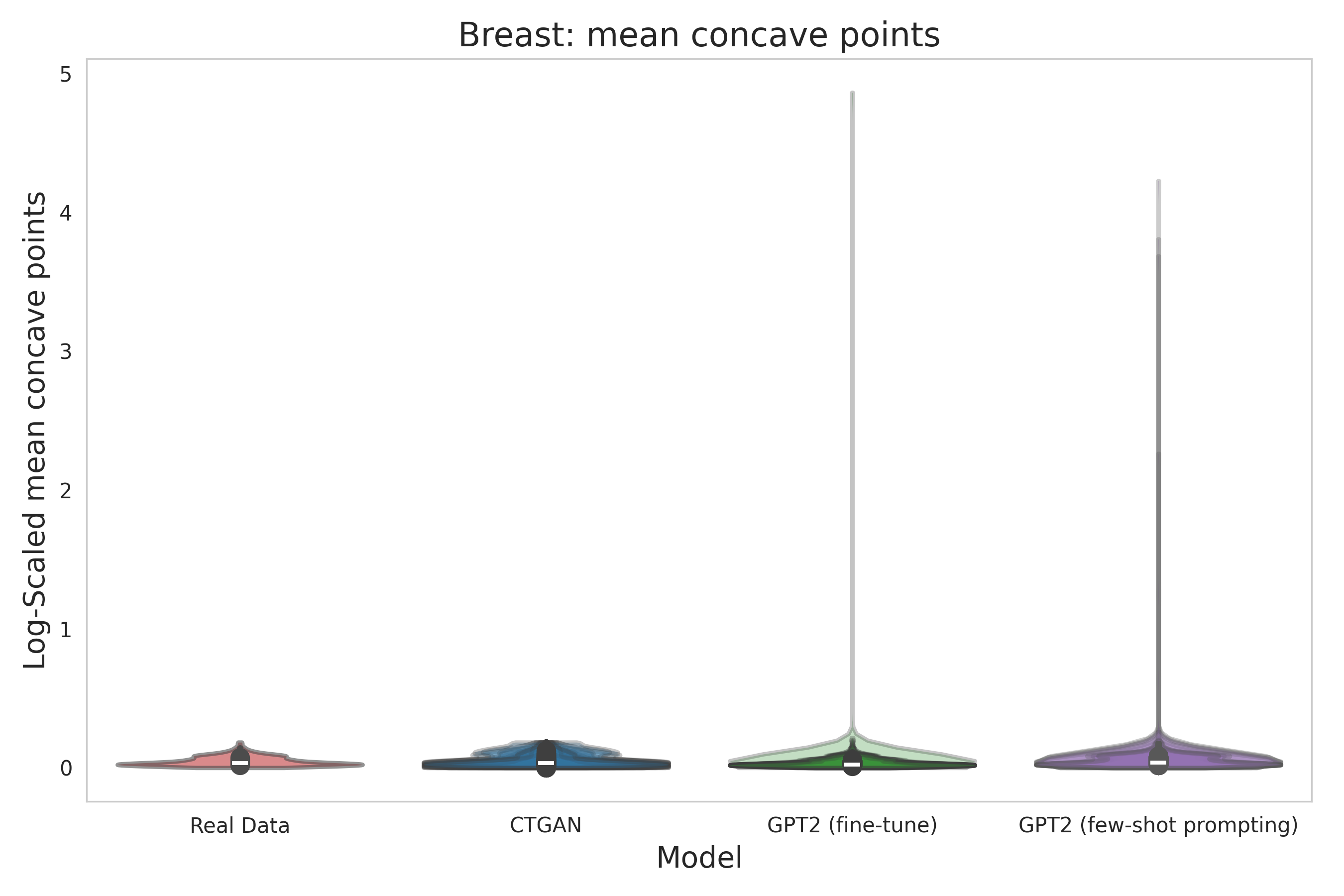}\\
    \includegraphics[width=0.49\linewidth]{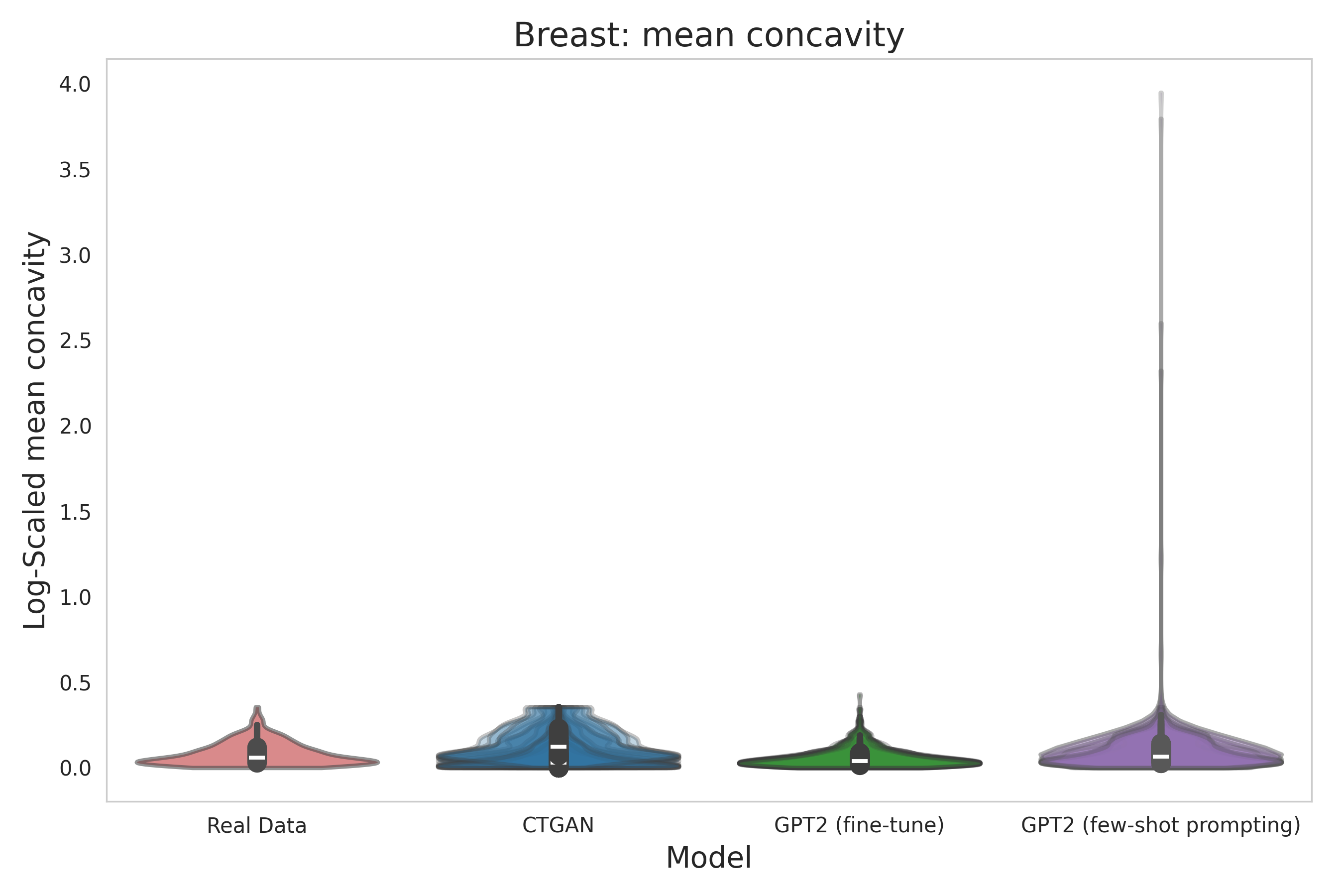}
    \includegraphics[width=0.49\linewidth]{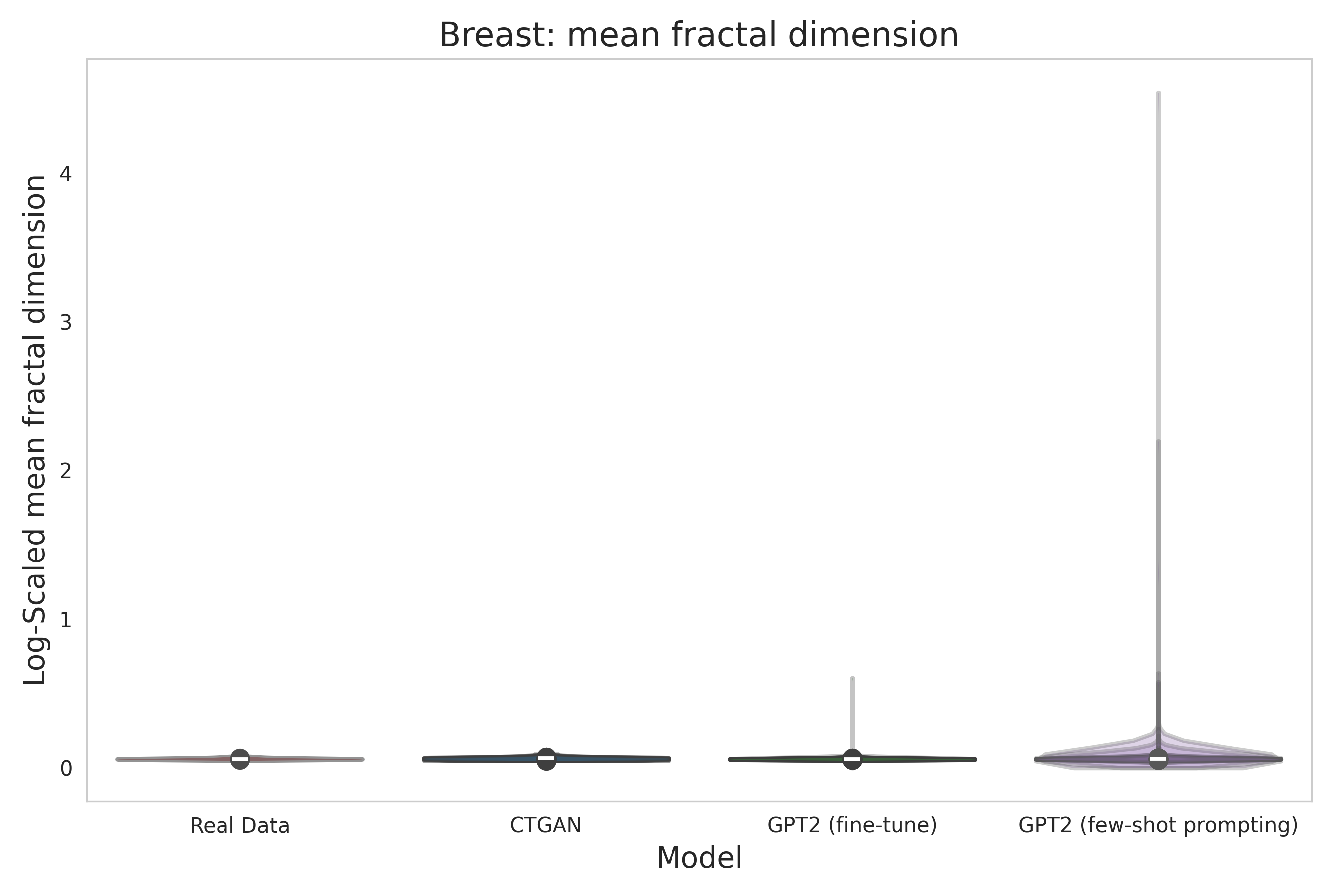}\\
    \includegraphics[width=0.49\linewidth]{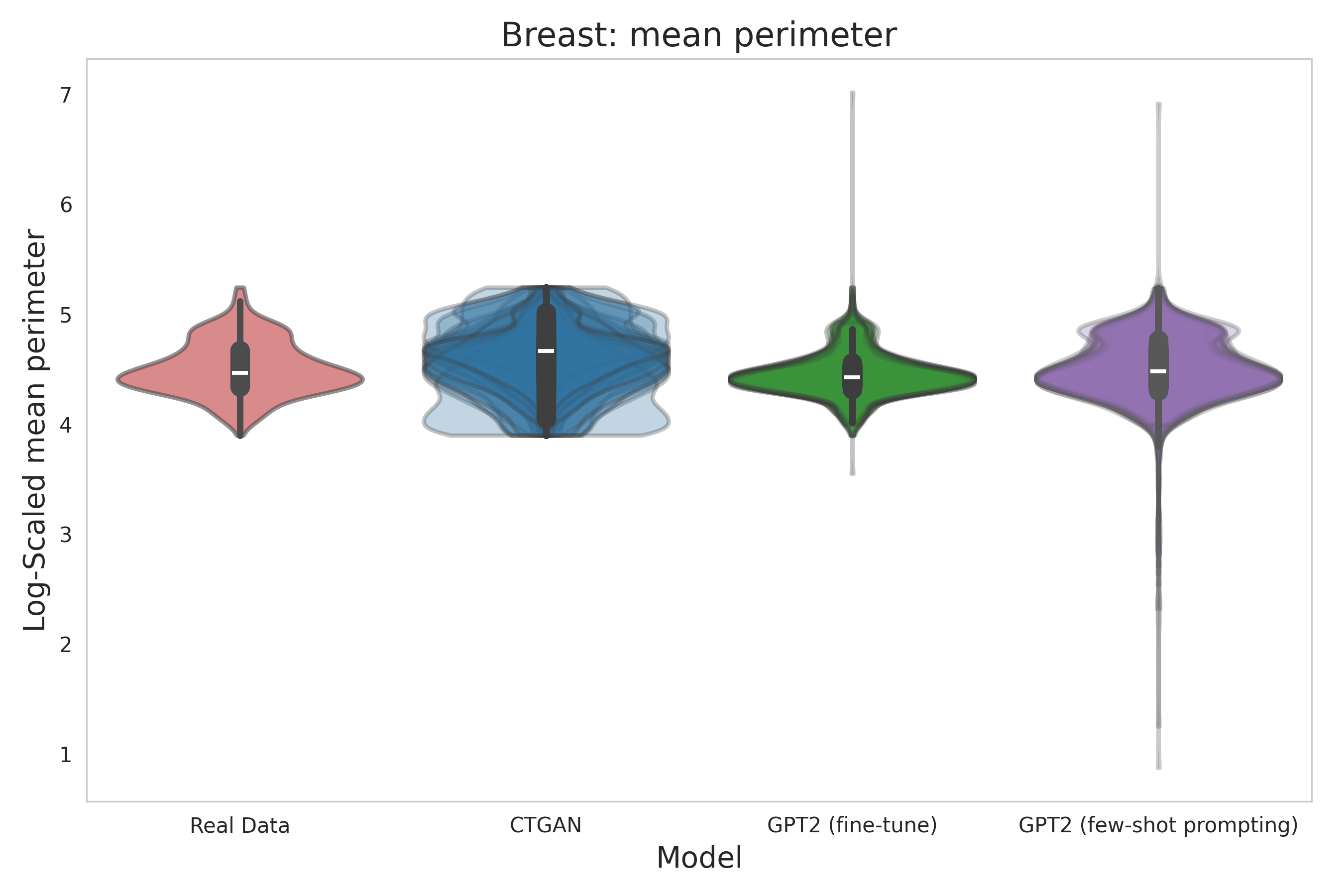}
    \includegraphics[width=0.49\linewidth]{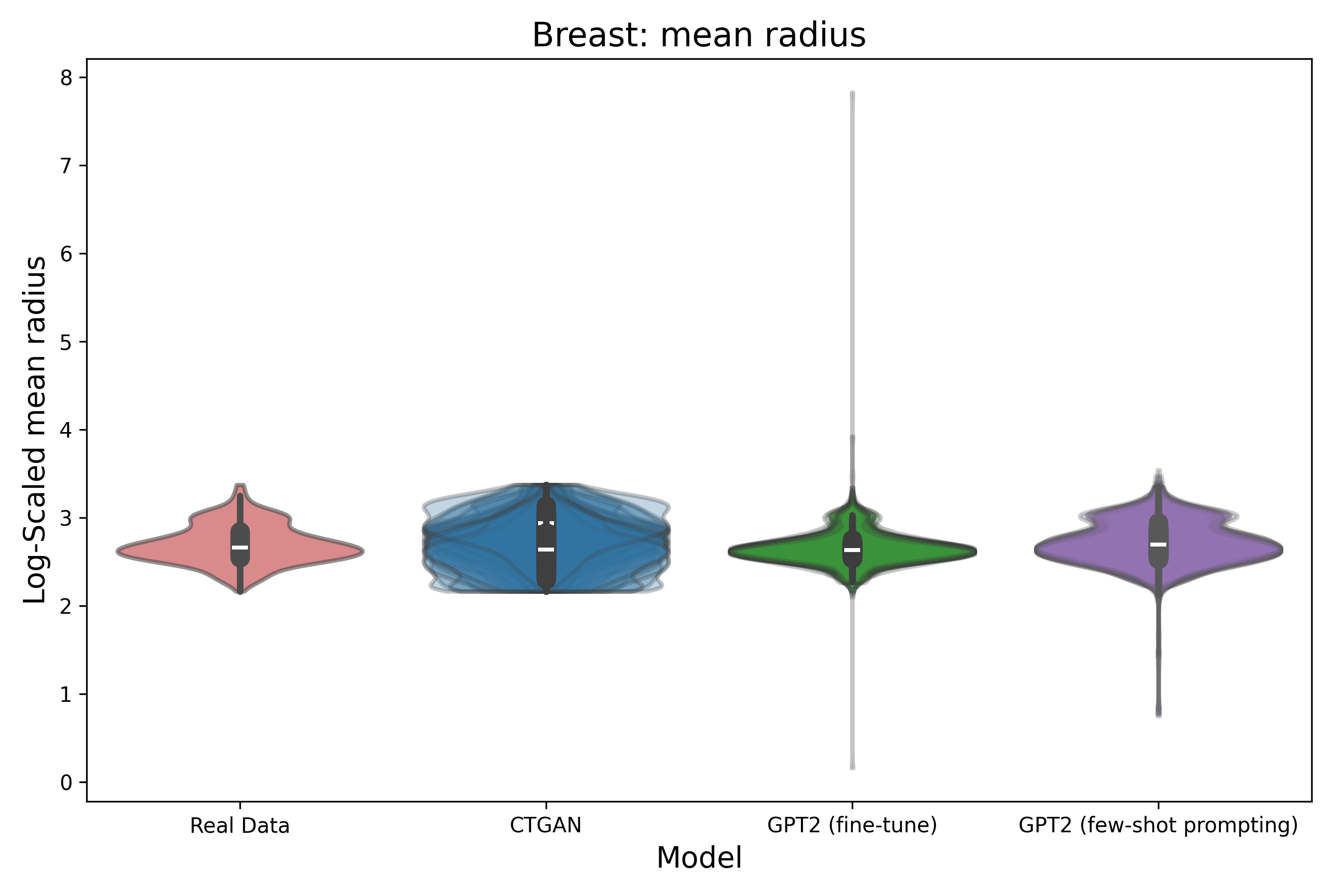}\\
    \caption{Further violin plots show the distributions of continuous columns in the Breast dataset.}
    \label{fig:breast2_violins}
\end{figure}

\begin{figure}
    \centering
    \includegraphics[width=0.49\linewidth]{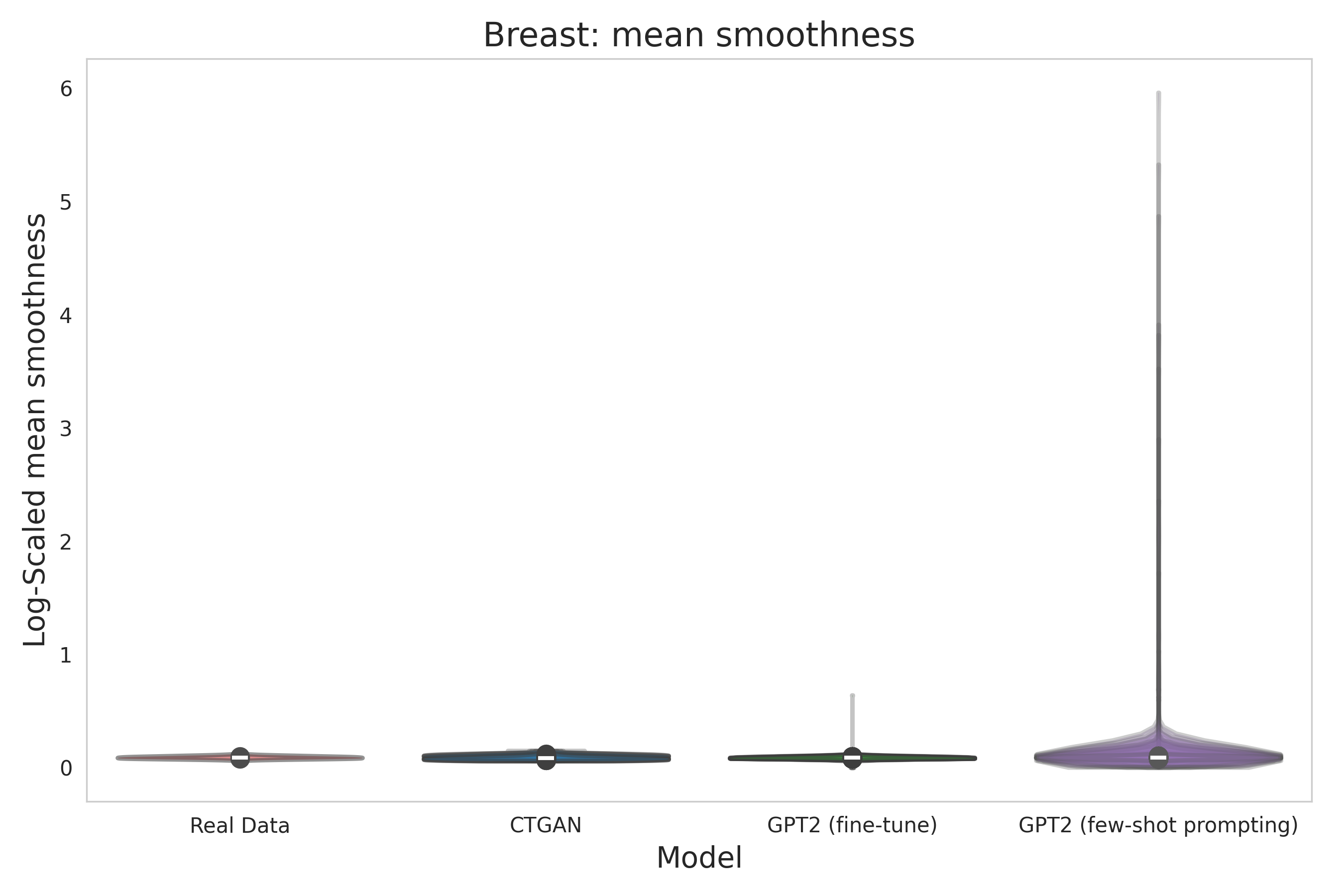}
    \includegraphics[width=0.49\linewidth]{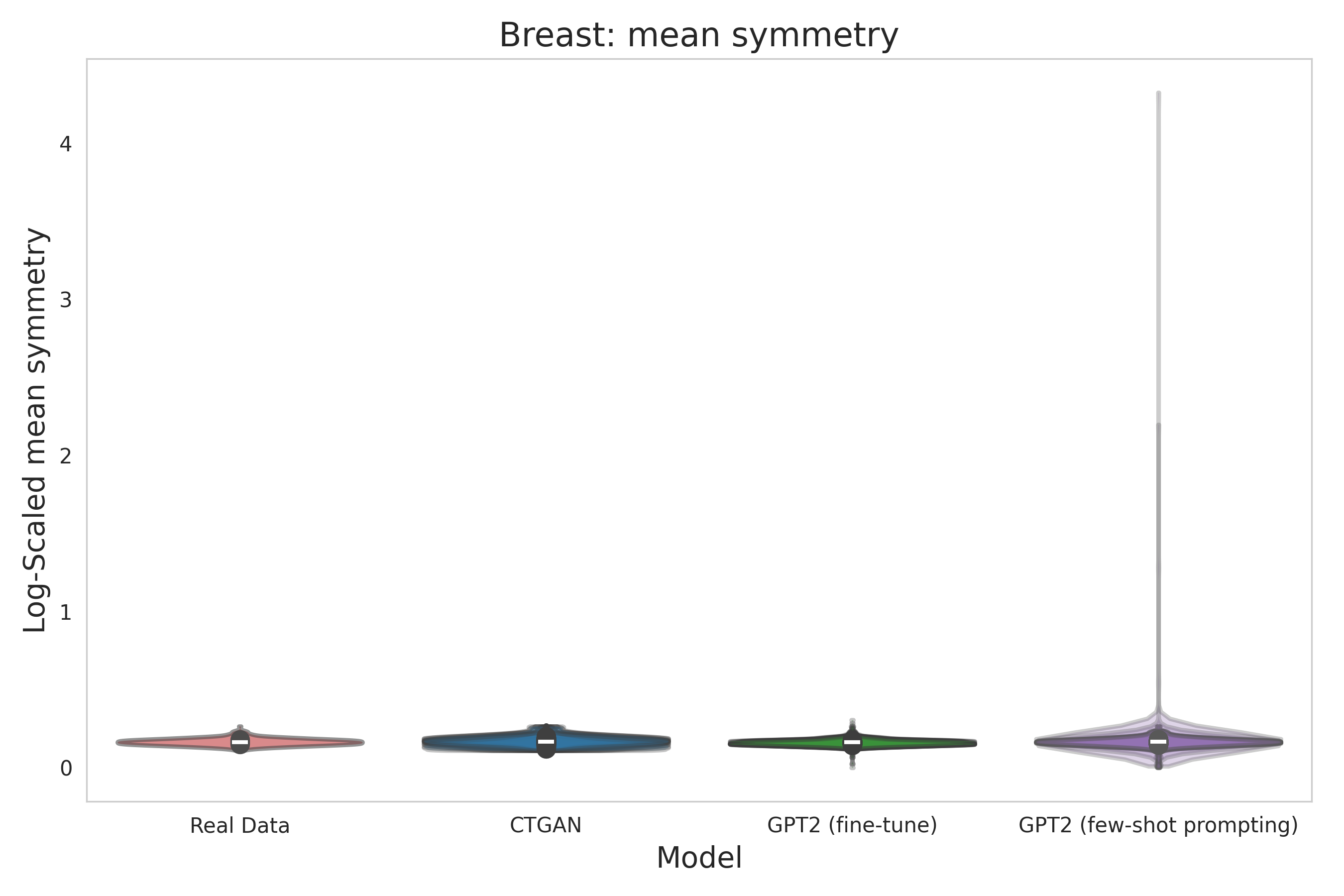}\\
    \includegraphics[width=0.49\linewidth]{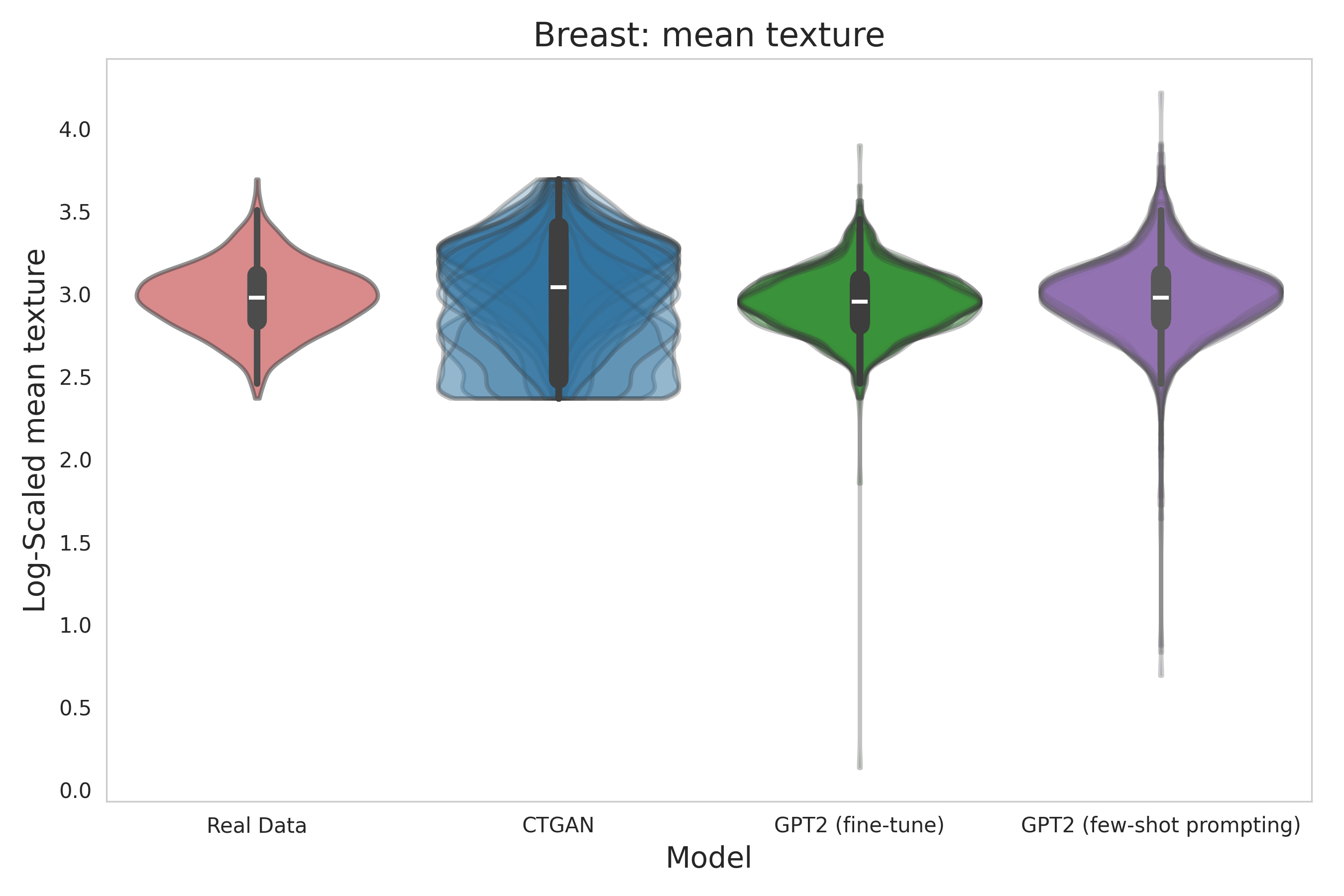}
    \includegraphics[width=0.49\linewidth]{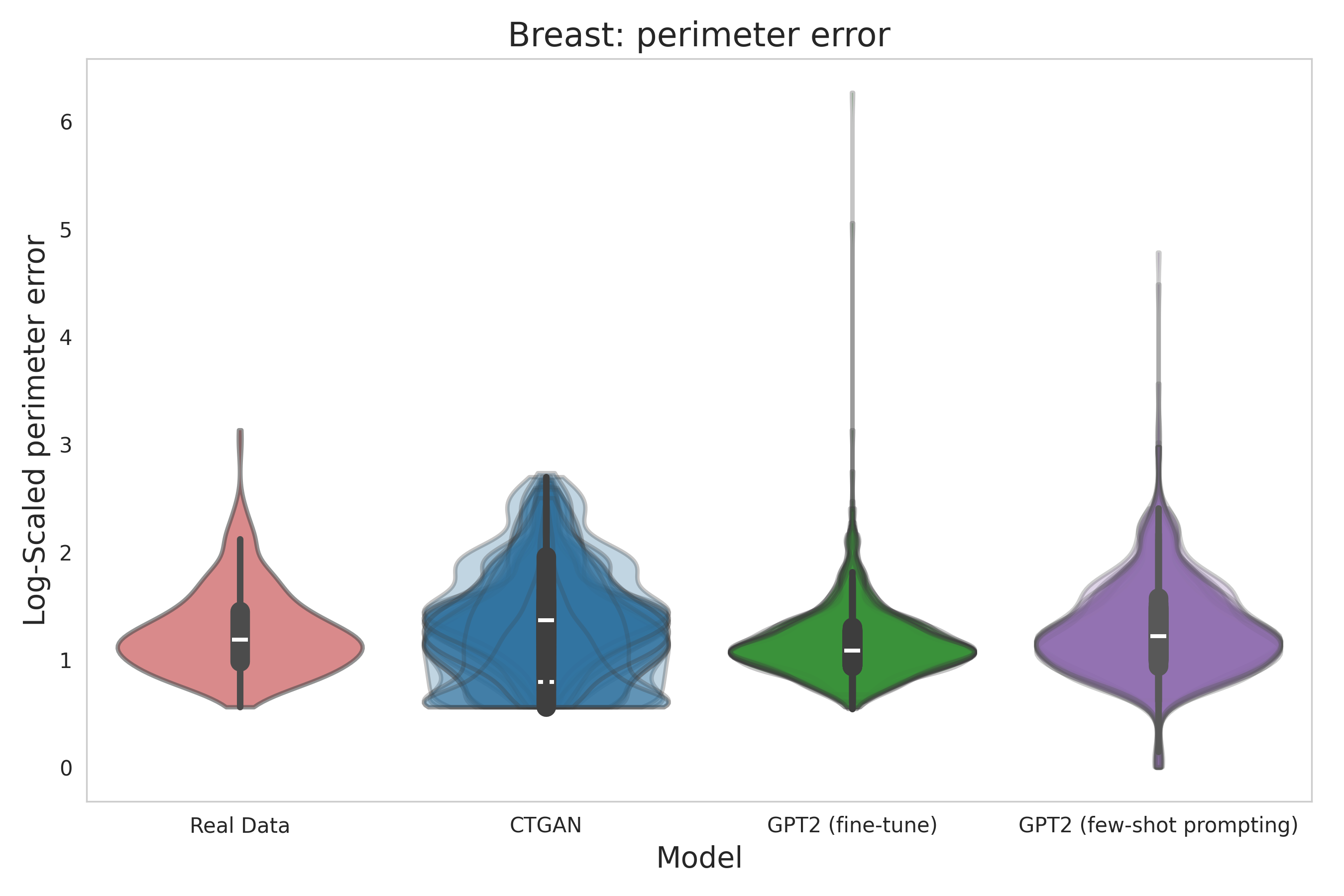}\\
    \includegraphics[width=0.49\linewidth]{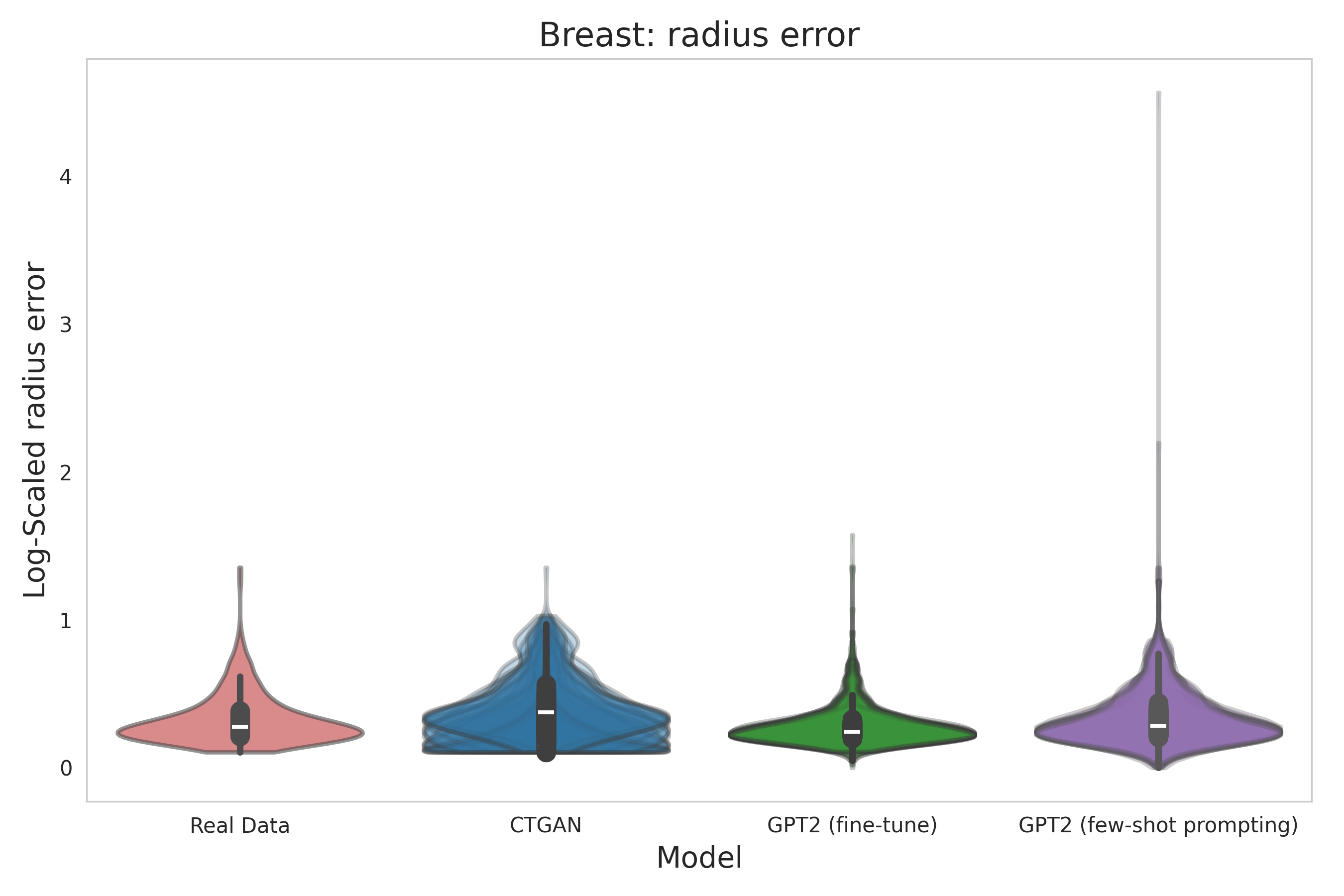}
    \includegraphics[width=0.49\linewidth]{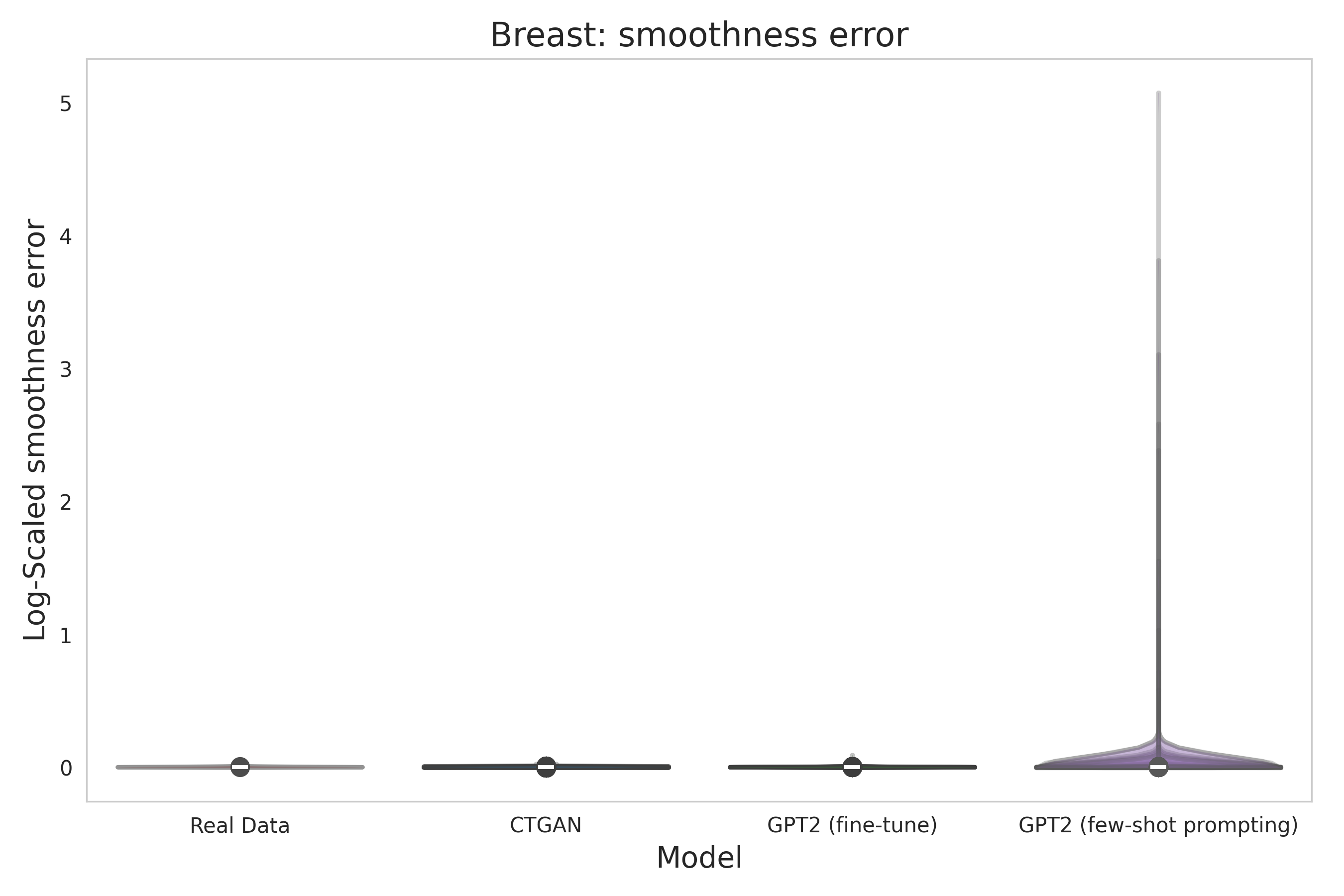}\\
    \caption{Further violin plots show the distributions of continuous columns in the Breast dataset.}
    \label{fig:breast3_violins}
\end{figure}

\begin{figure}
    \centering
    \includegraphics[width=0.49\linewidth]{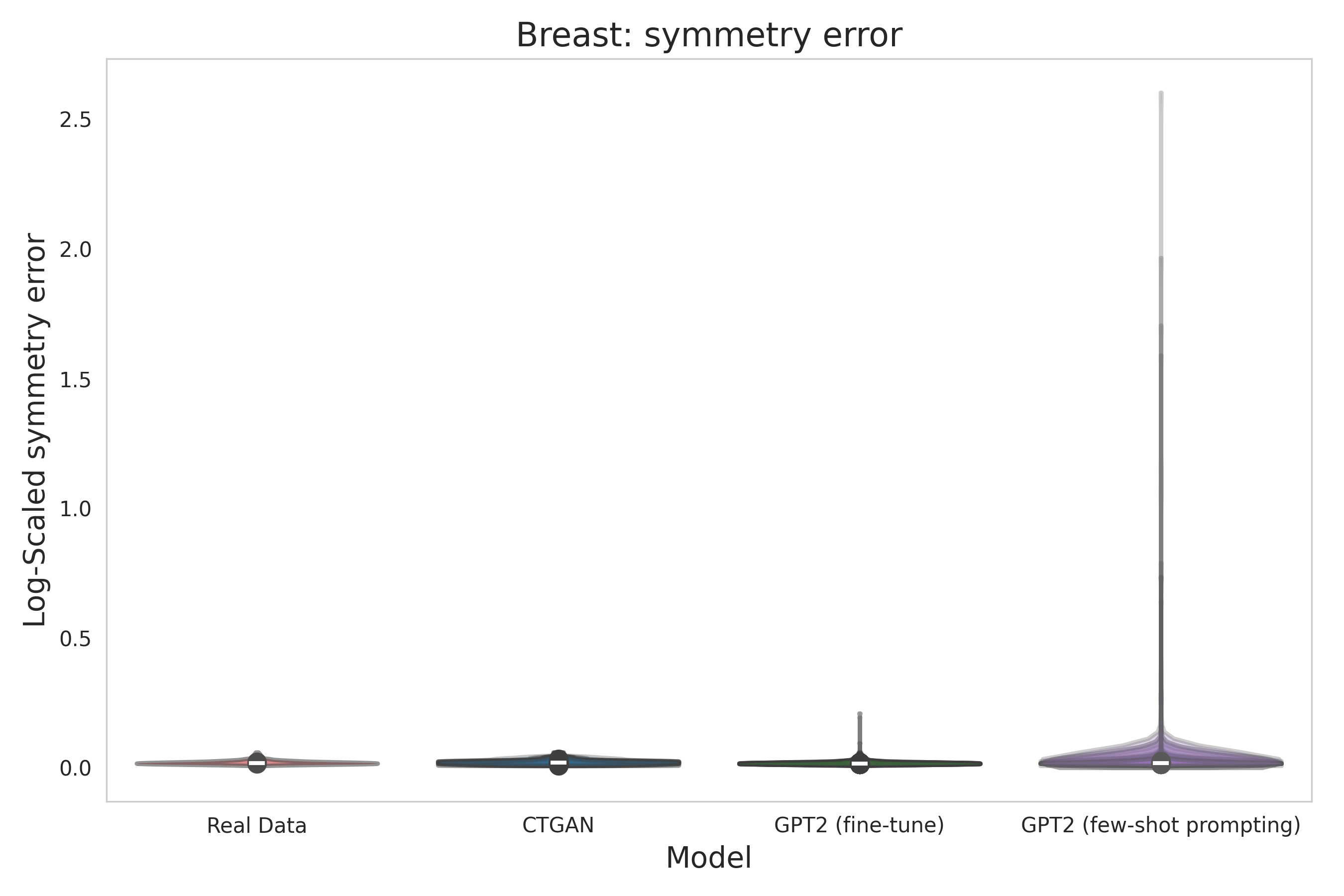}
    \includegraphics[width=0.49\linewidth]{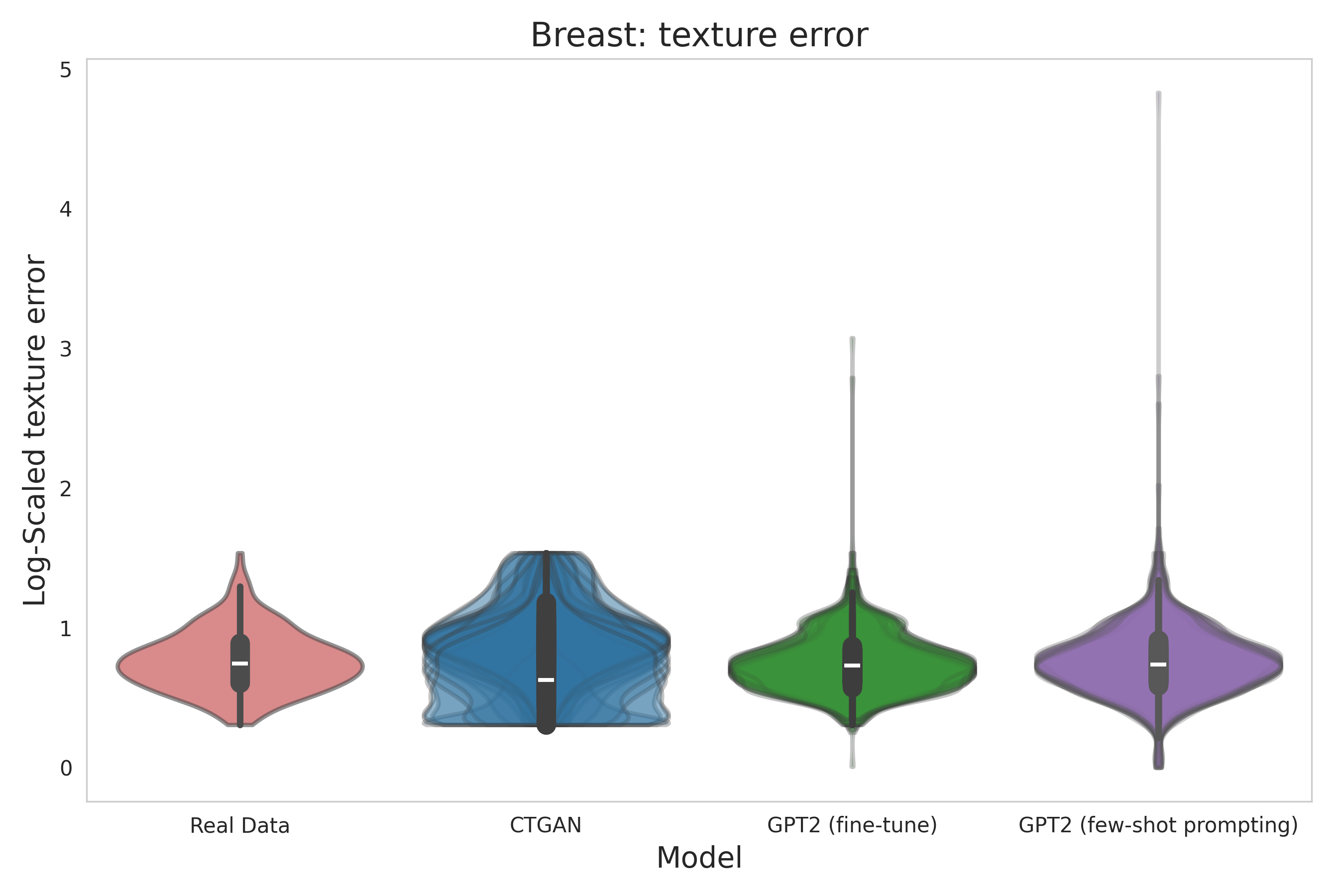}\\
    \includegraphics[width=0.49\linewidth]{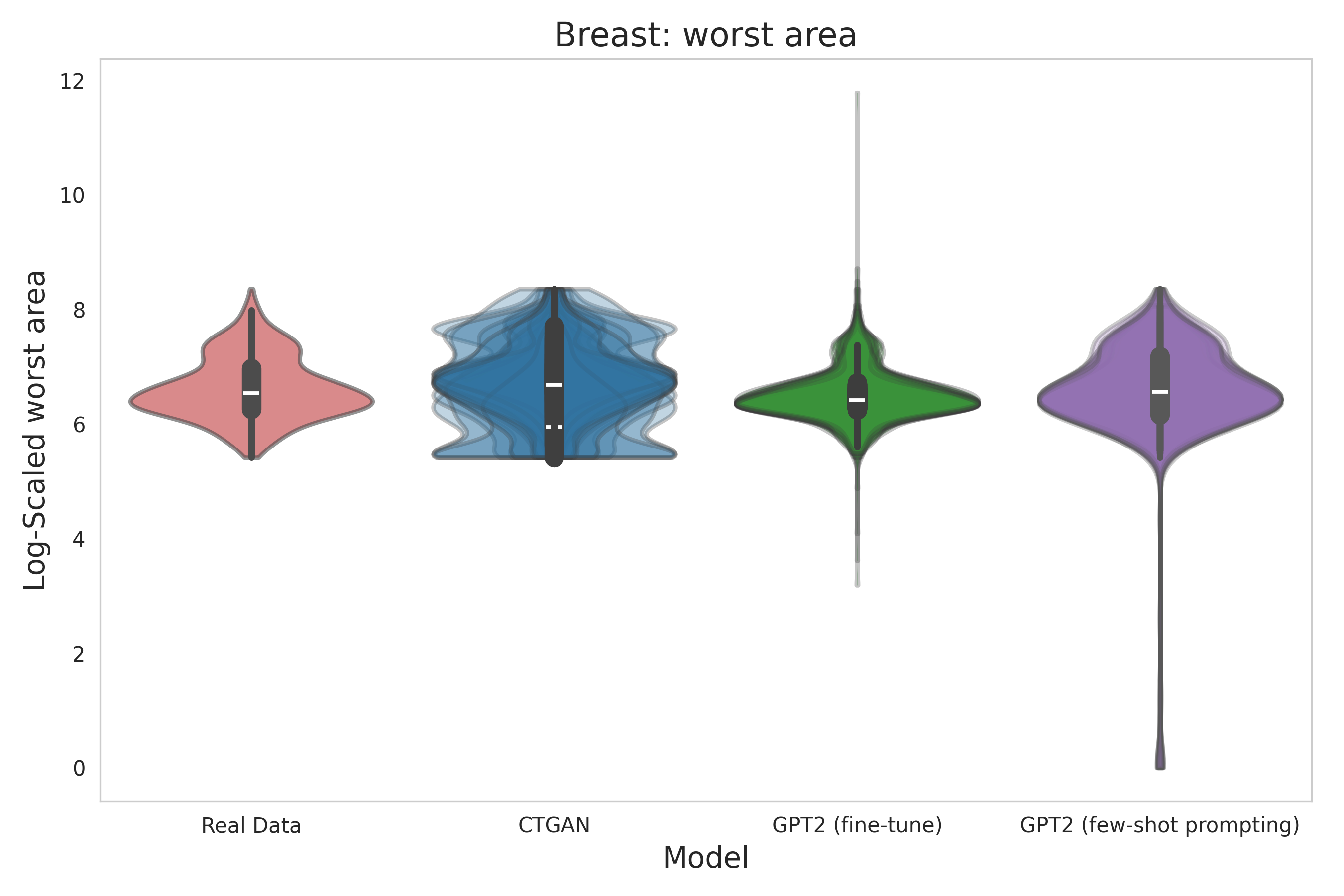}
    \includegraphics[width=0.49\linewidth]{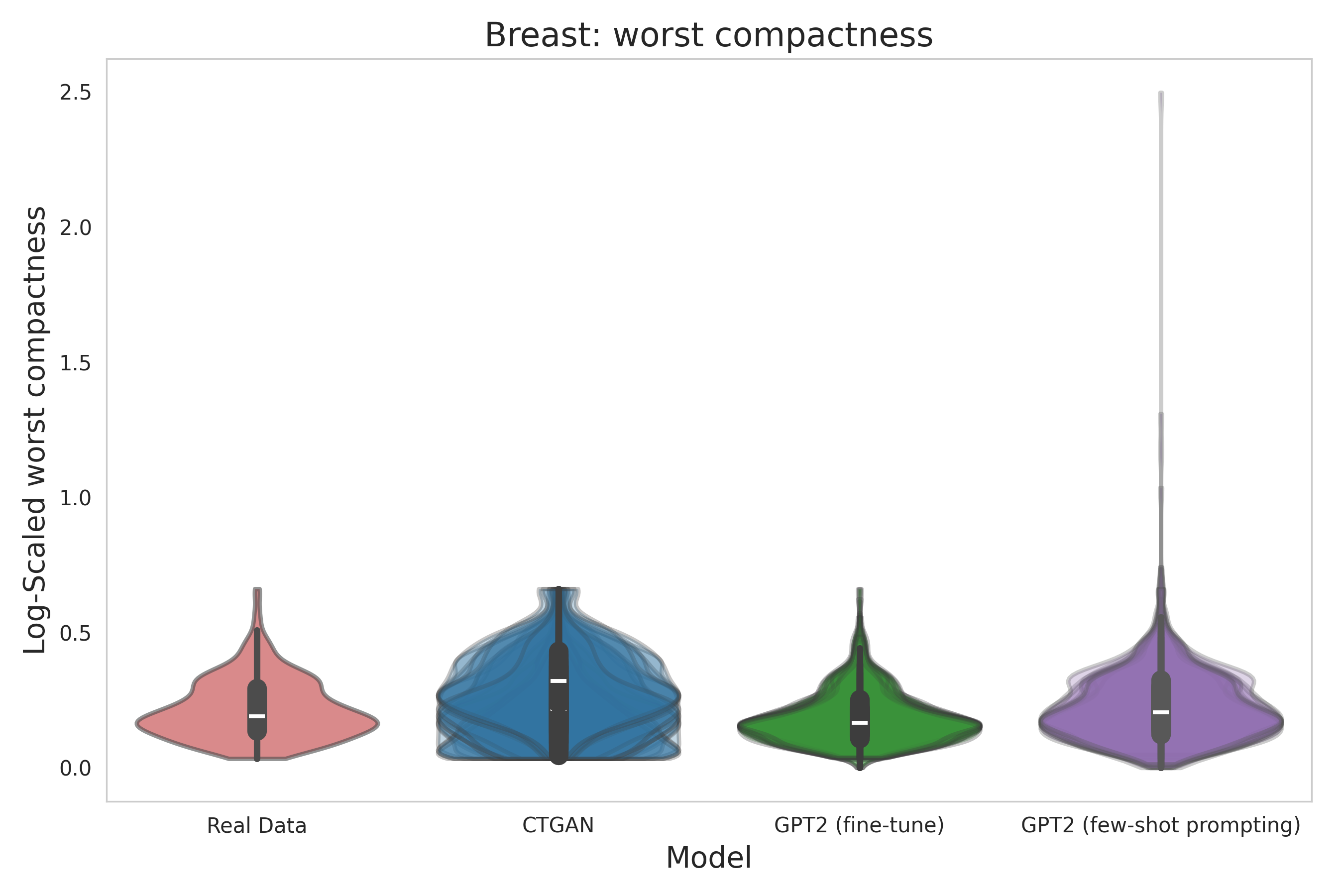}\\
    \includegraphics[width=0.49\linewidth]{figures/violins/breast/breast_cancer_worst_concave_points.png}
    \includegraphics[width=0.49\linewidth]{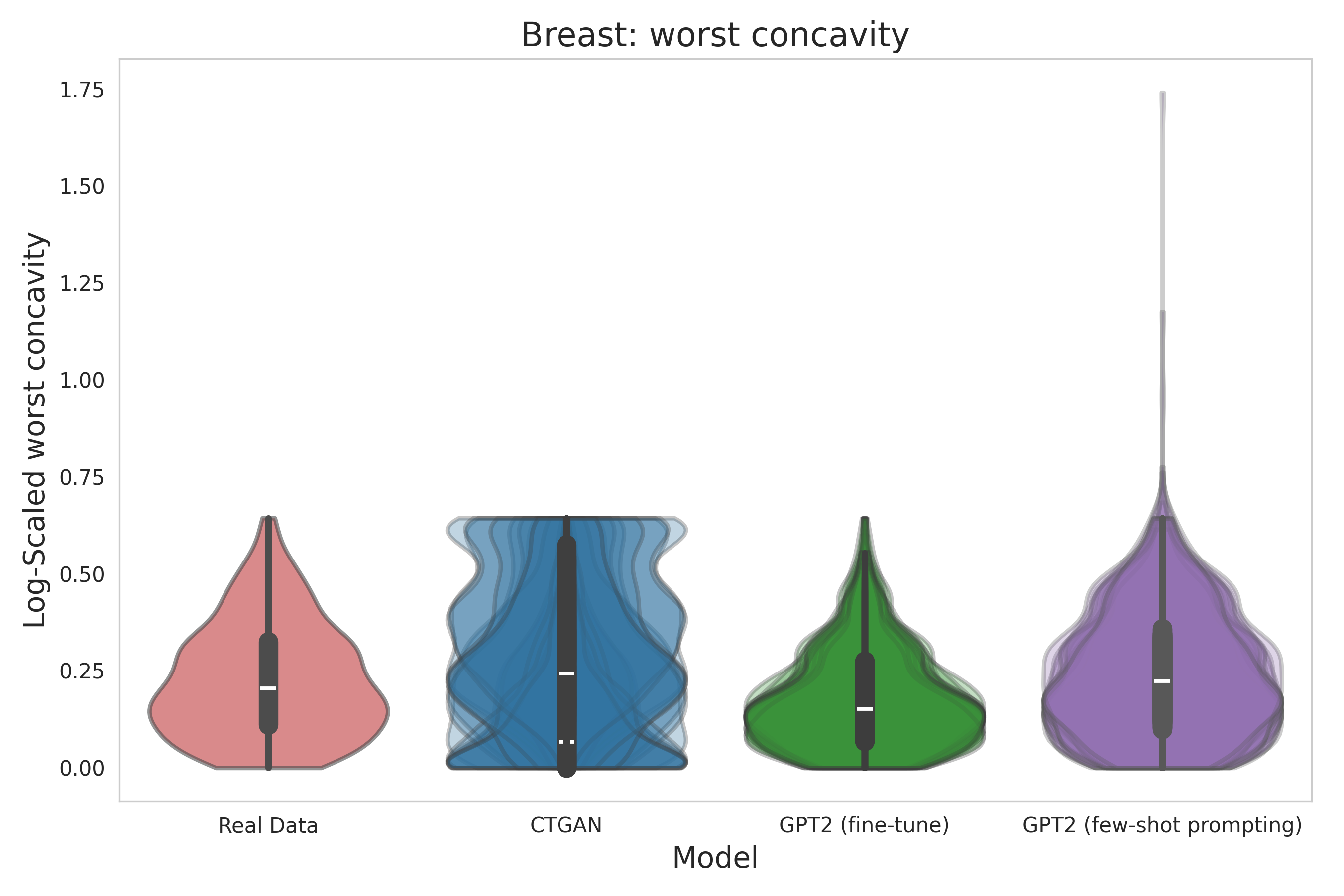}\\
    \caption{Further violin plots show the distributions of continuous columns in the Breast dataset.}
    \label{fig:breast4_violins}
\end{figure}

\begin{figure}
    \centering
    \includegraphics[width=0.49\linewidth]{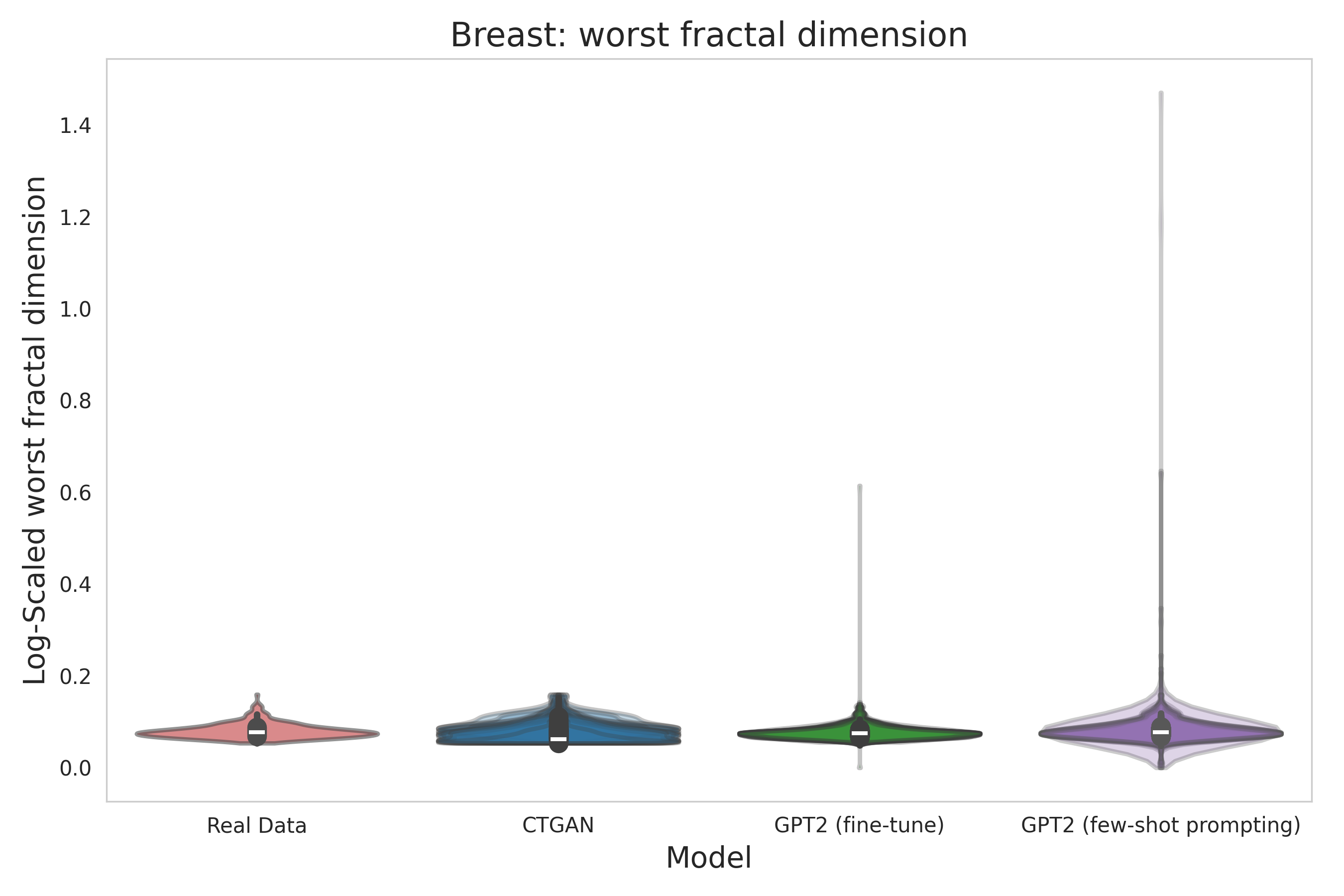}
    \includegraphics[width=0.49\linewidth]{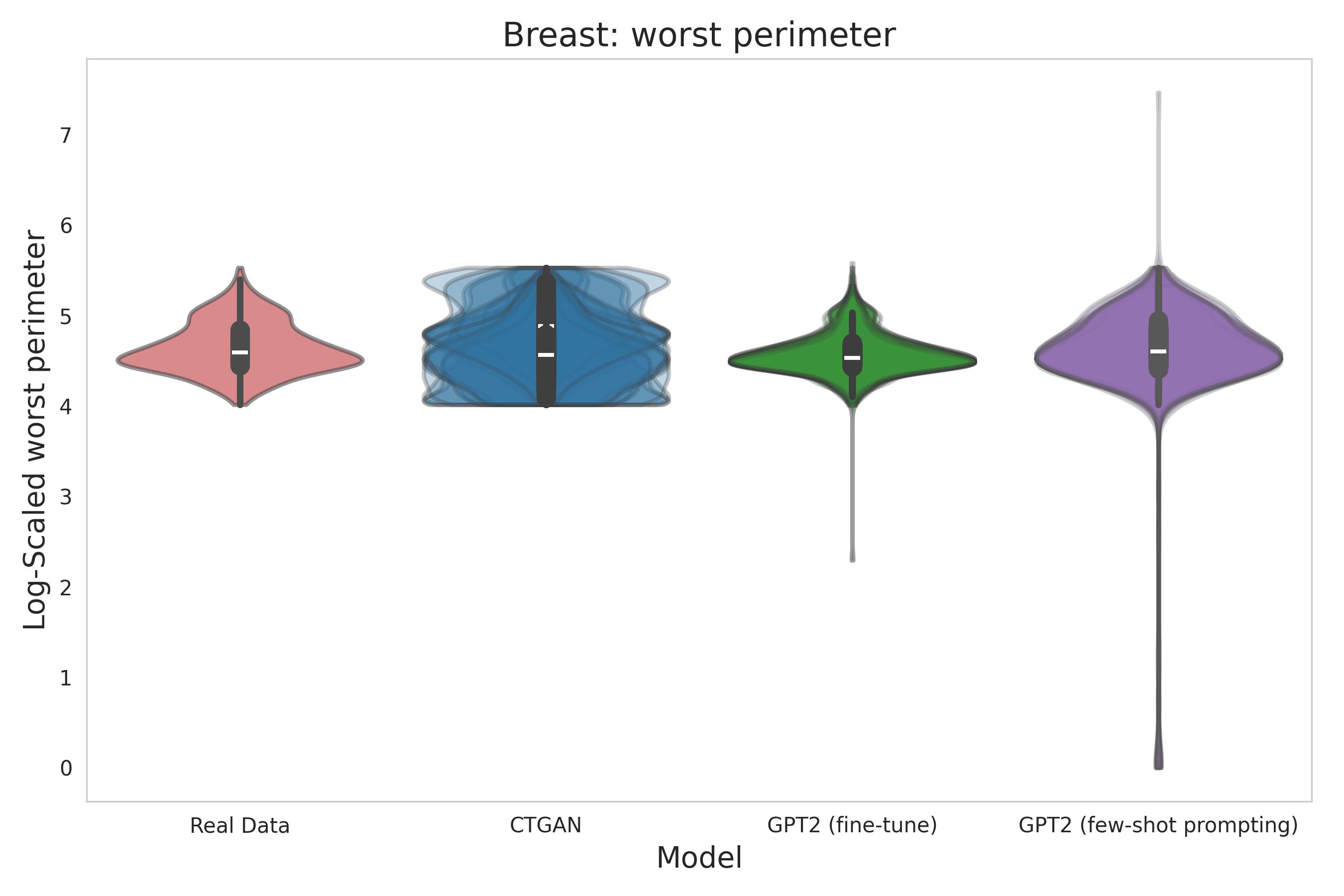}\\
    \includegraphics[width=0.49\linewidth]{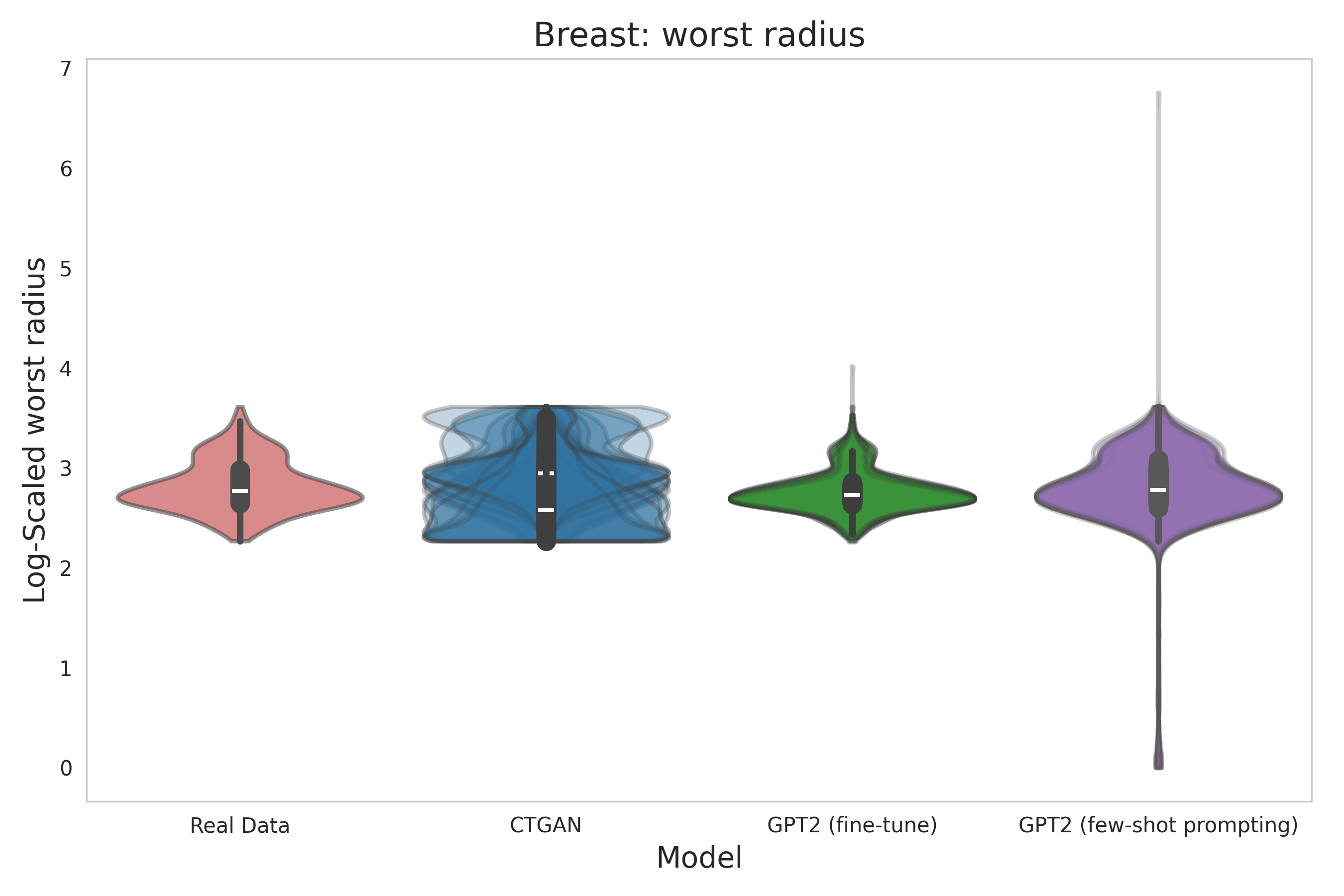}
    \includegraphics[width=0.49\linewidth]{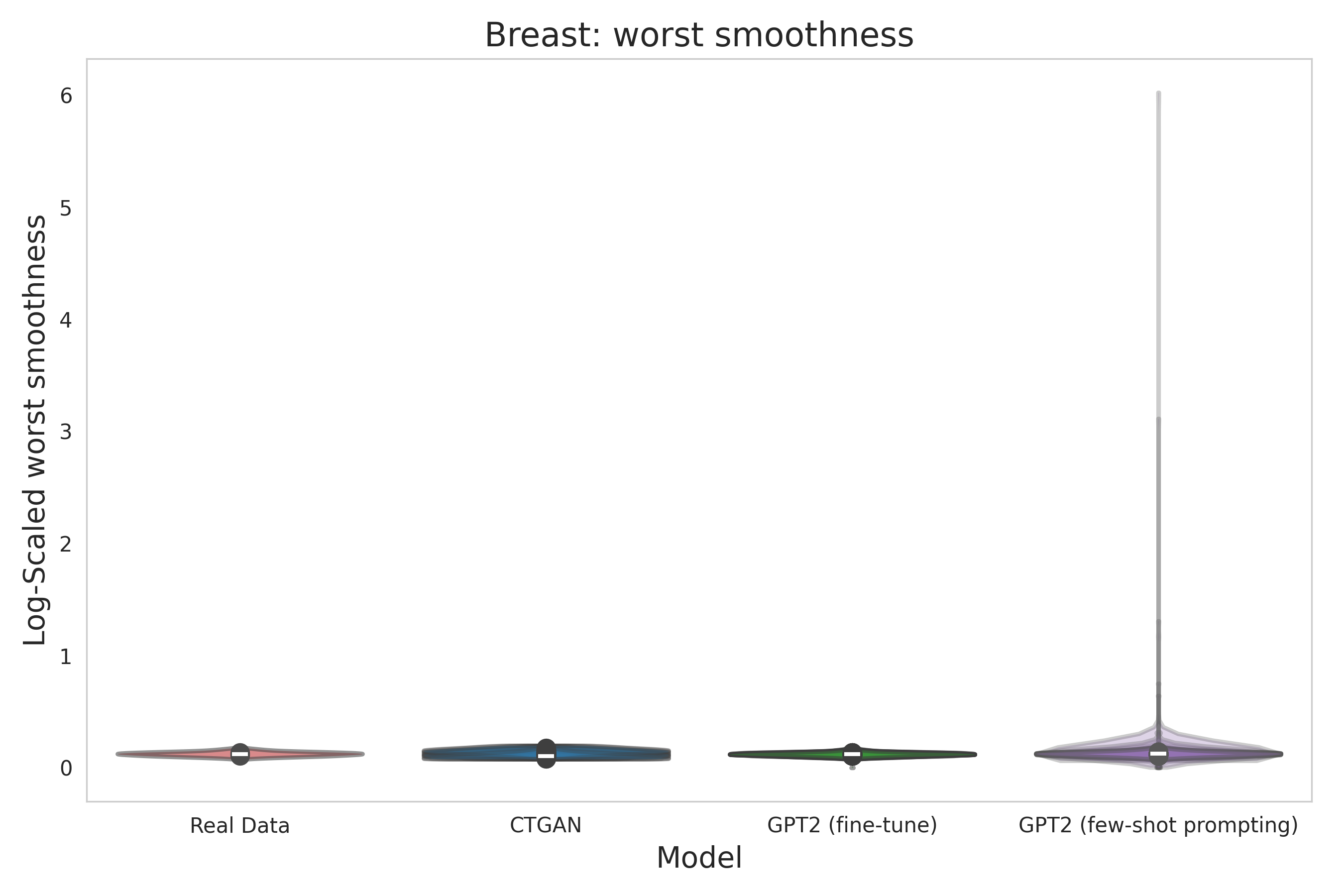}\\
    \includegraphics[width=0.49\linewidth]{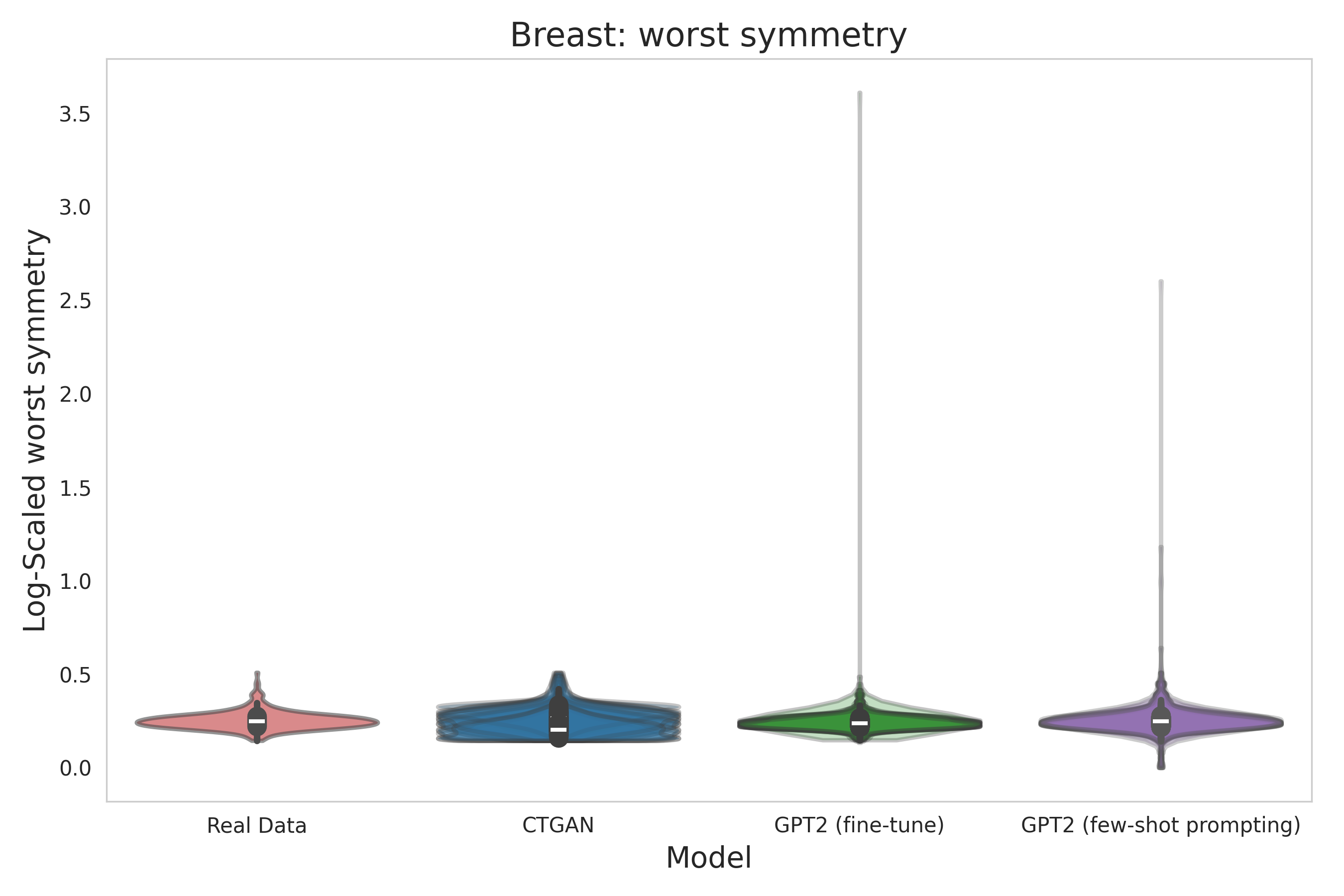}
    \includegraphics[width=0.49\linewidth]{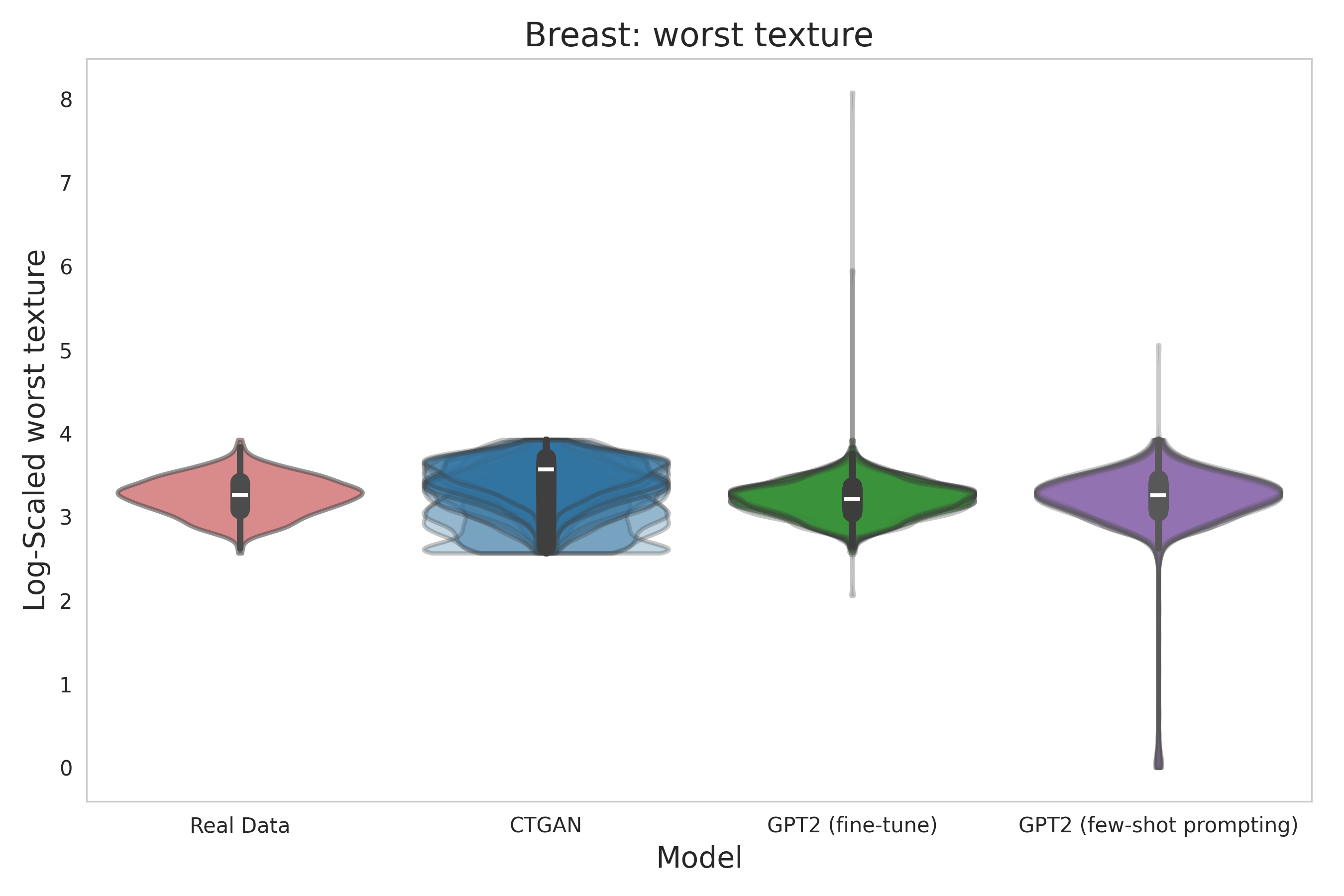}\\
    \caption{Further violin plots show the distributions of continuous columns in the Breast dataset.}
    \label{fig:breast5_violins}
\end{figure}

\begin{figure}
    \centering
    \includegraphics[width=0.49\linewidth]{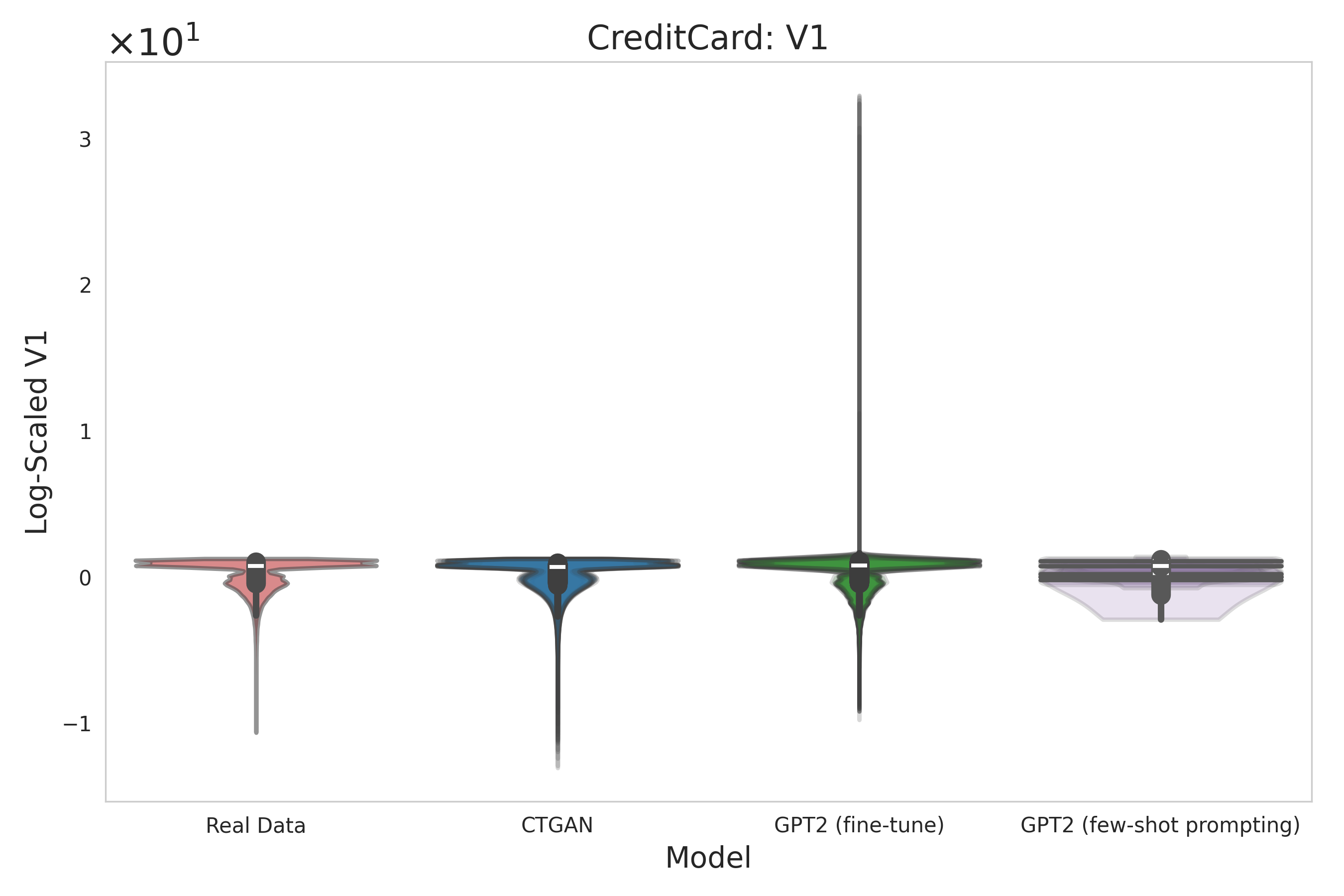}
    \includegraphics[width=0.49\linewidth]{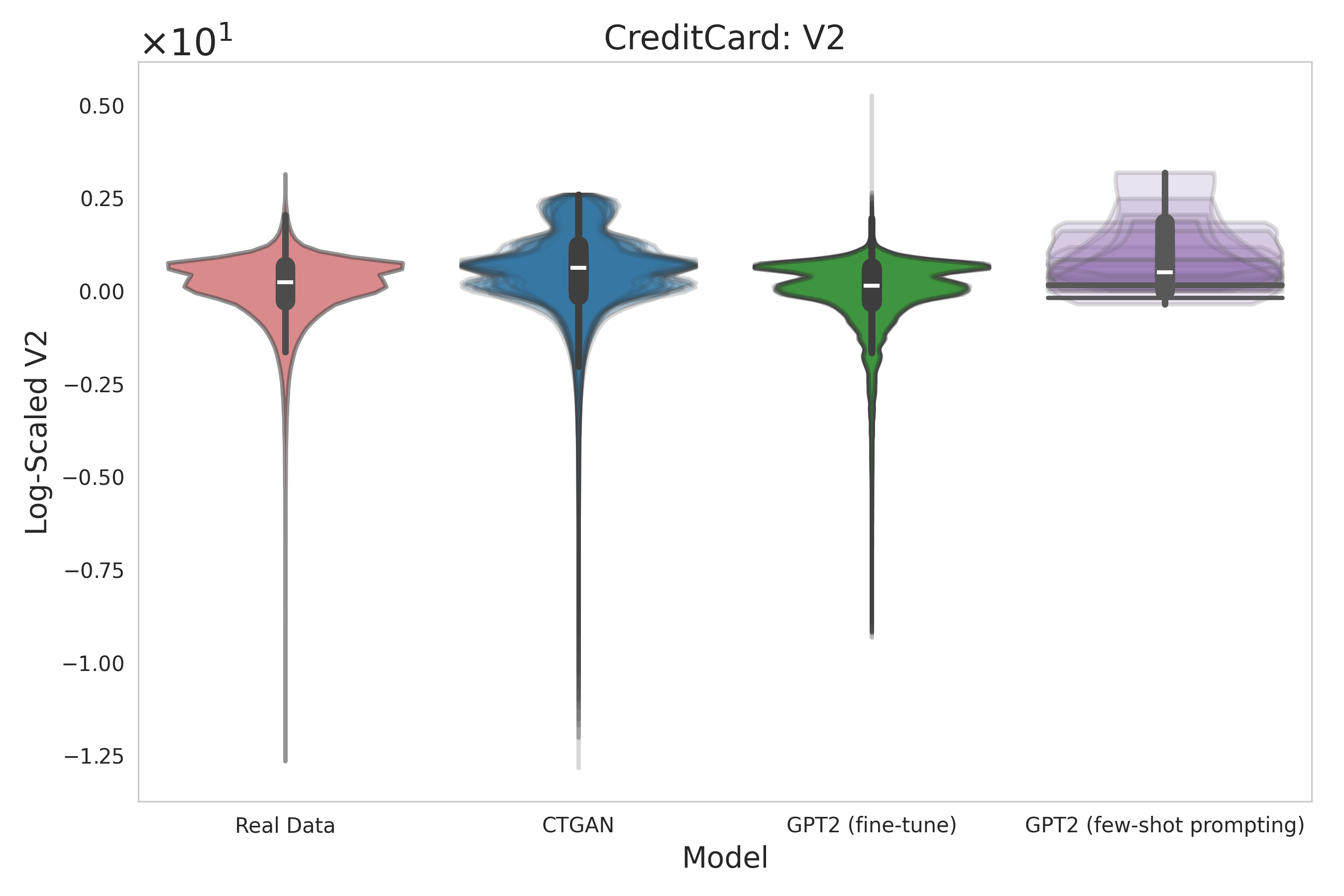}\\
    \includegraphics[width=0.49\linewidth]{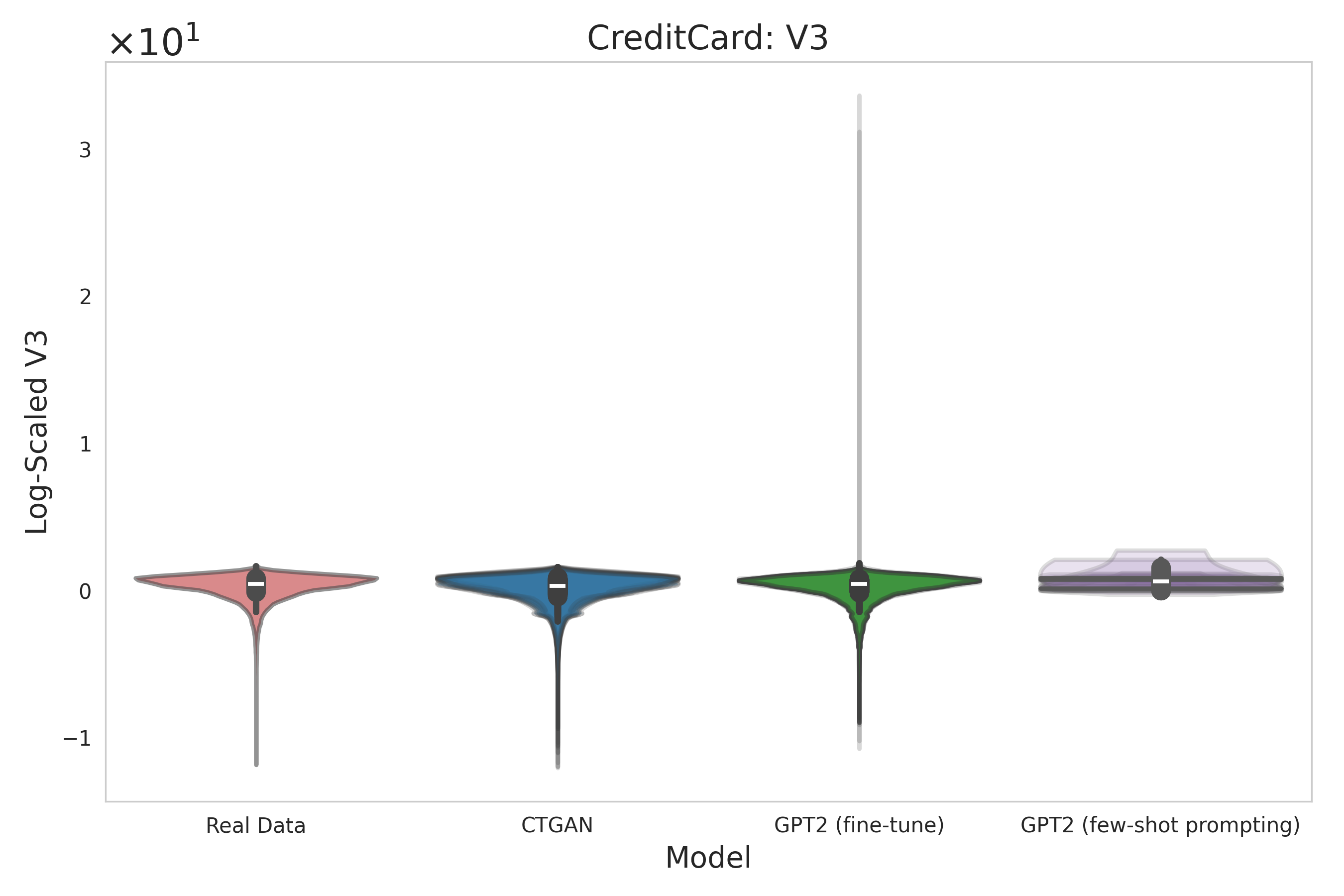}
    \includegraphics[width=0.49\linewidth]{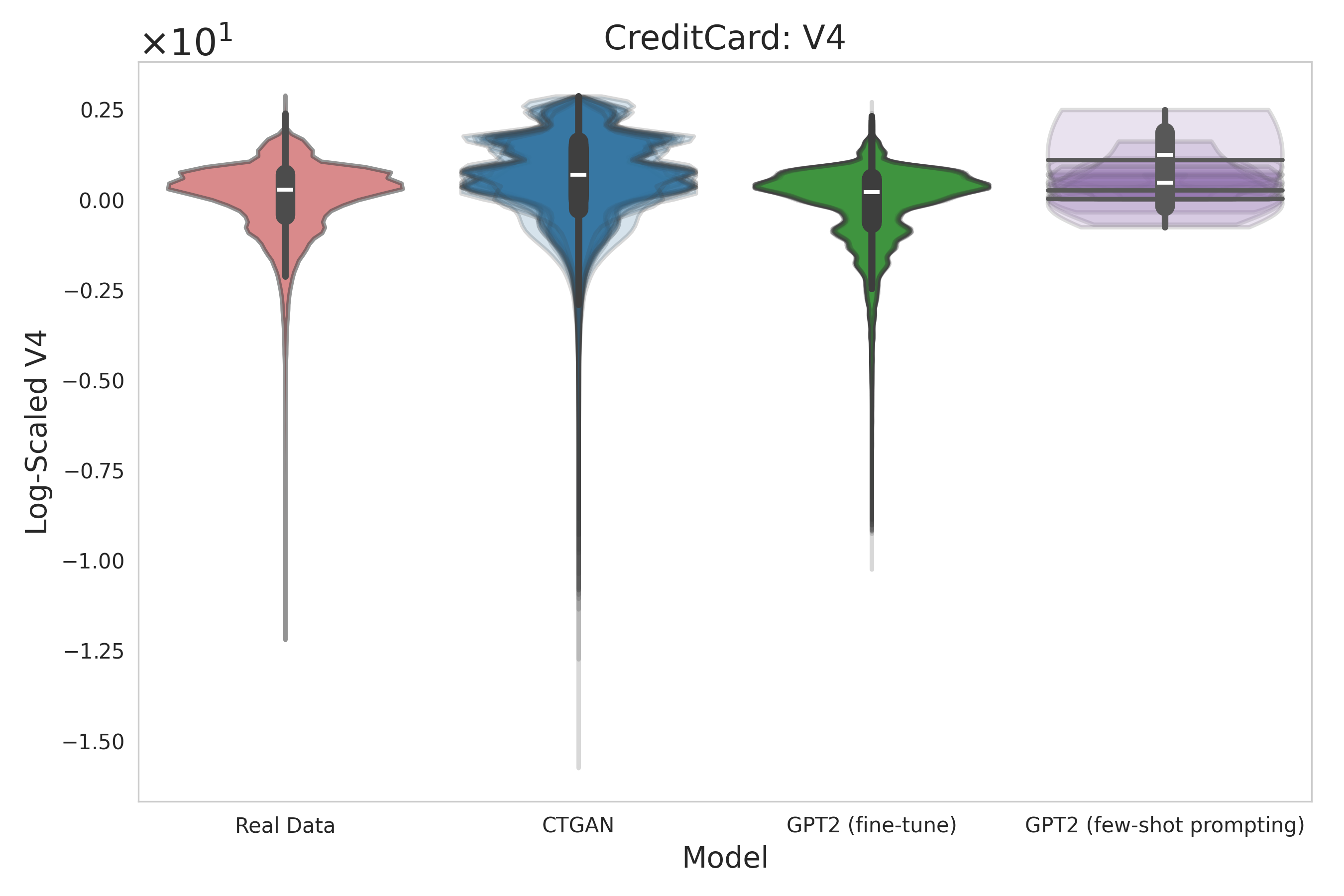}\\
    \includegraphics[width=0.49\linewidth]{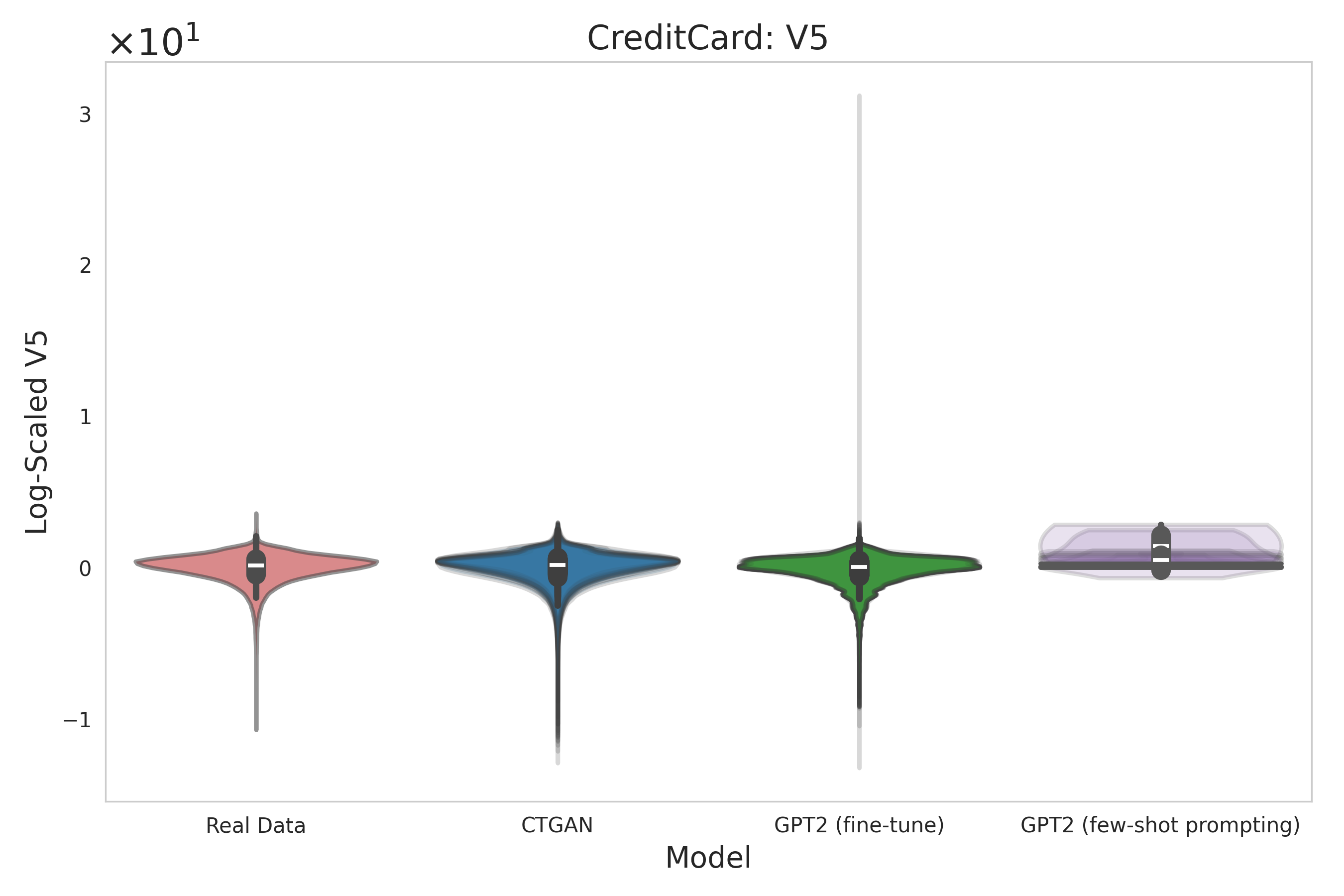}
    \includegraphics[width=0.49\linewidth]{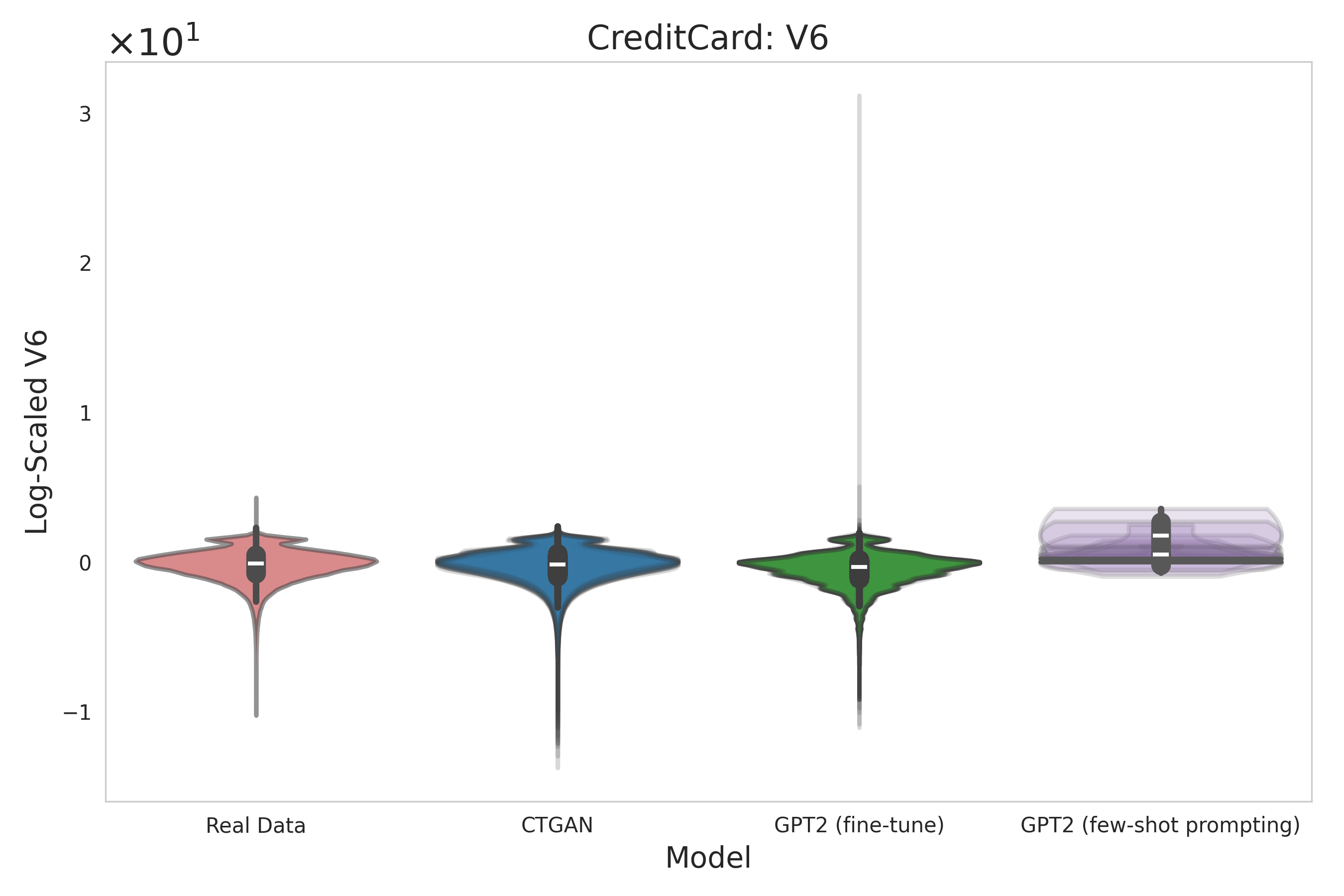}\\
    \caption{Violin plots show the distributions of continuous columns in the Credit dataset.}
    \label{fig:credit1_violins}
\end{figure}

\begin{figure}
    \centering
    
    \includegraphics[width=0.49\linewidth]{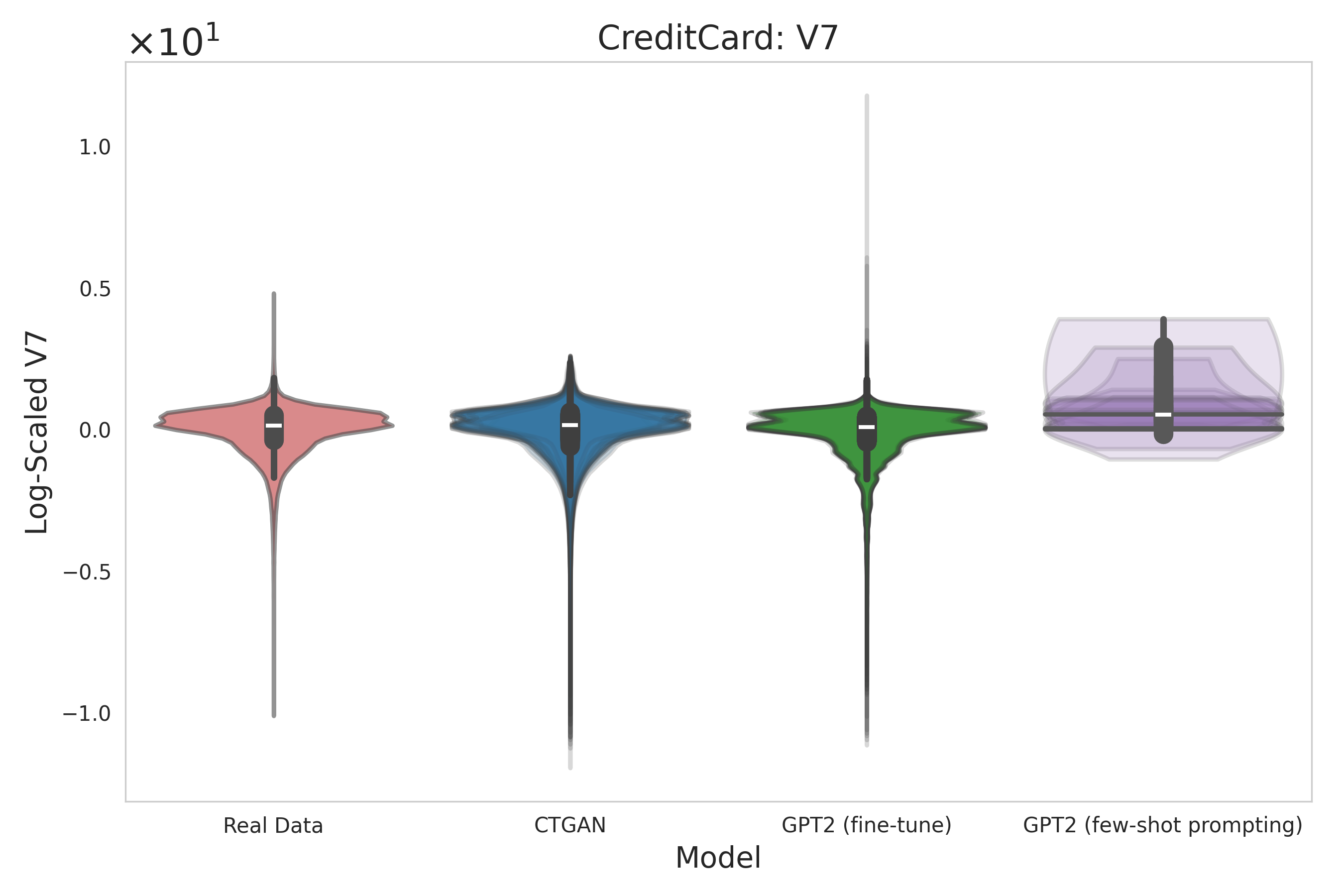}
    \includegraphics[width=0.49\linewidth]{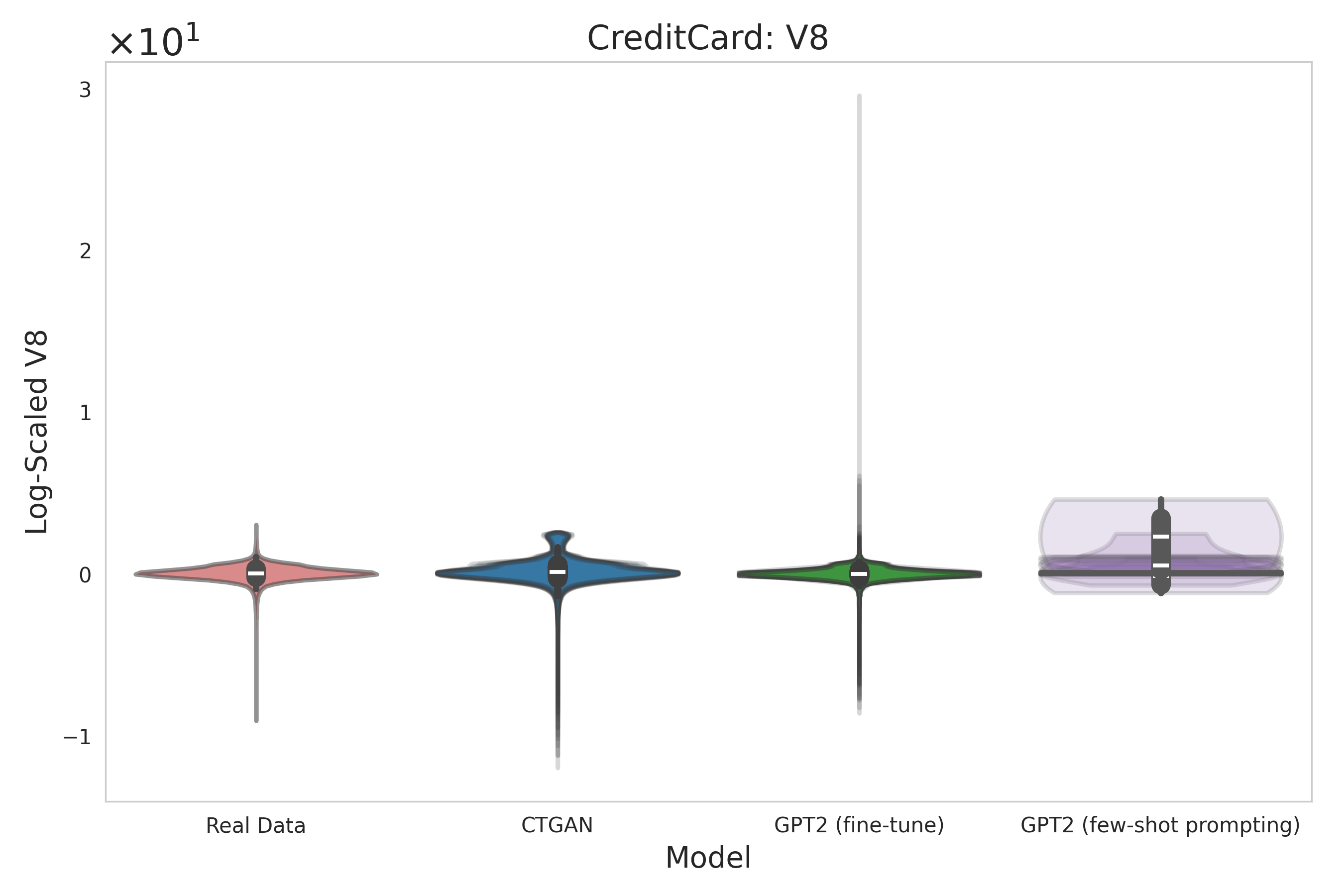}\\
    \includegraphics[width=0.49\linewidth]{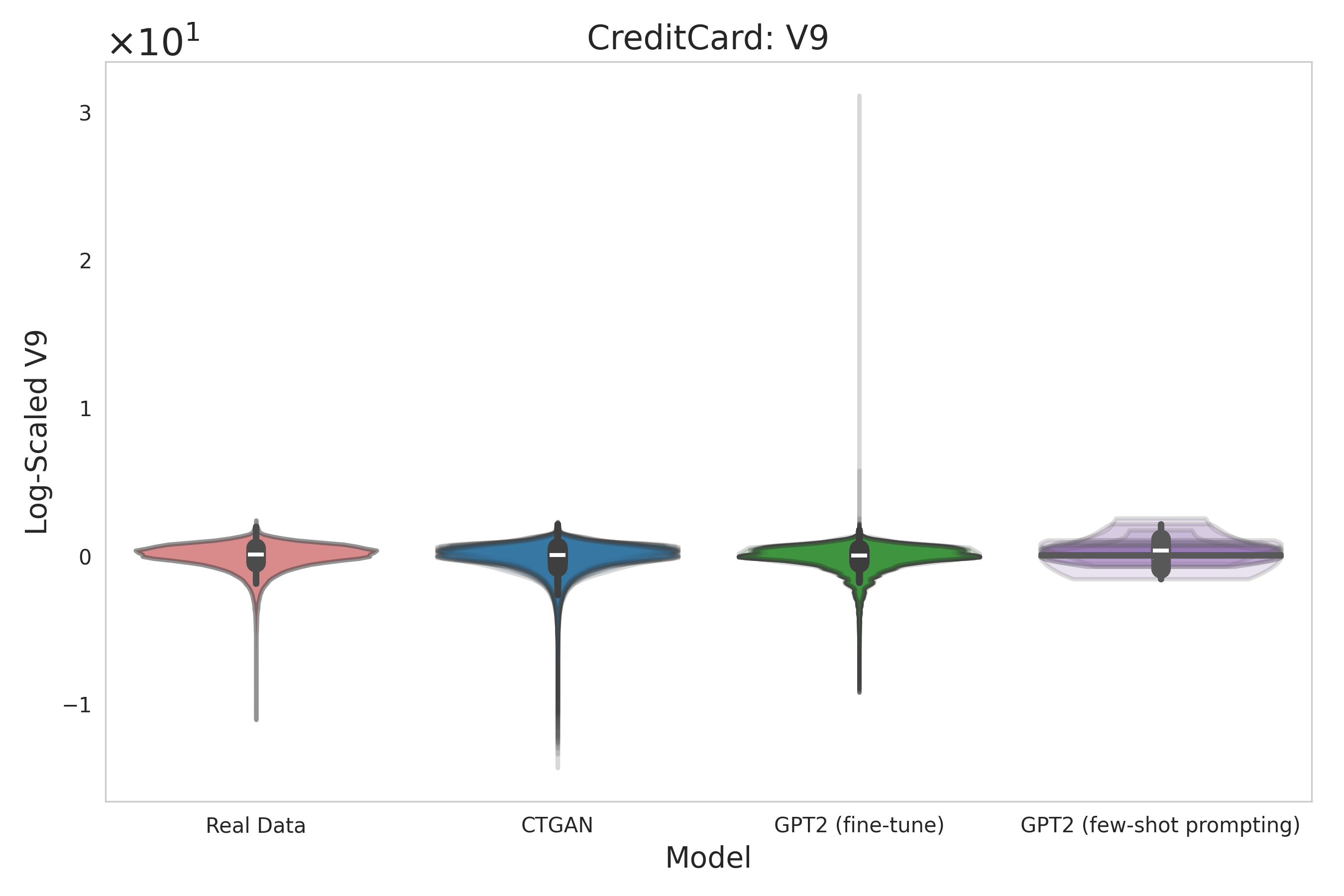}
    \includegraphics[width=0.49\linewidth]{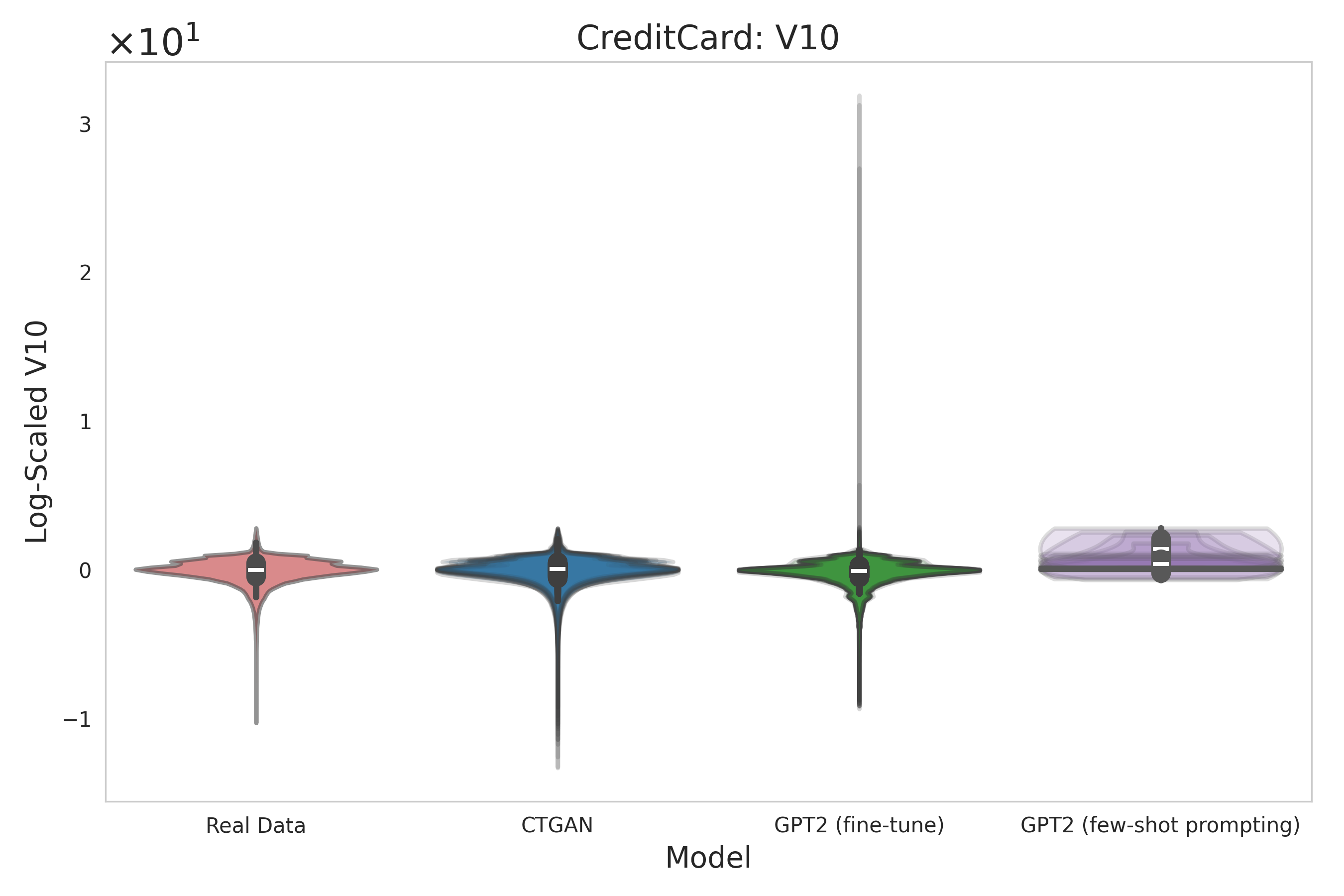}\\
    \includegraphics[width=0.49\linewidth]{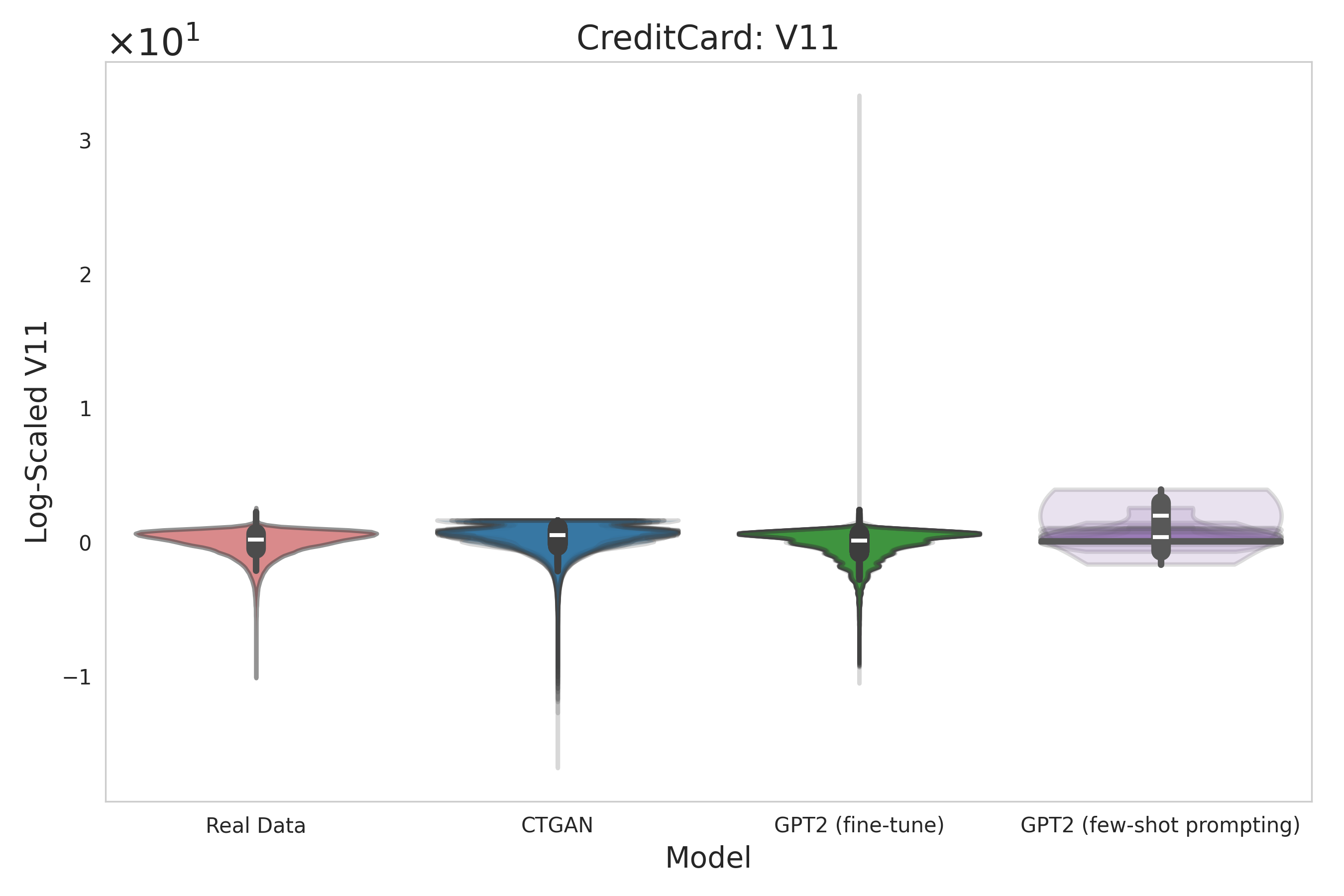}
    \includegraphics[width=0.49\linewidth]{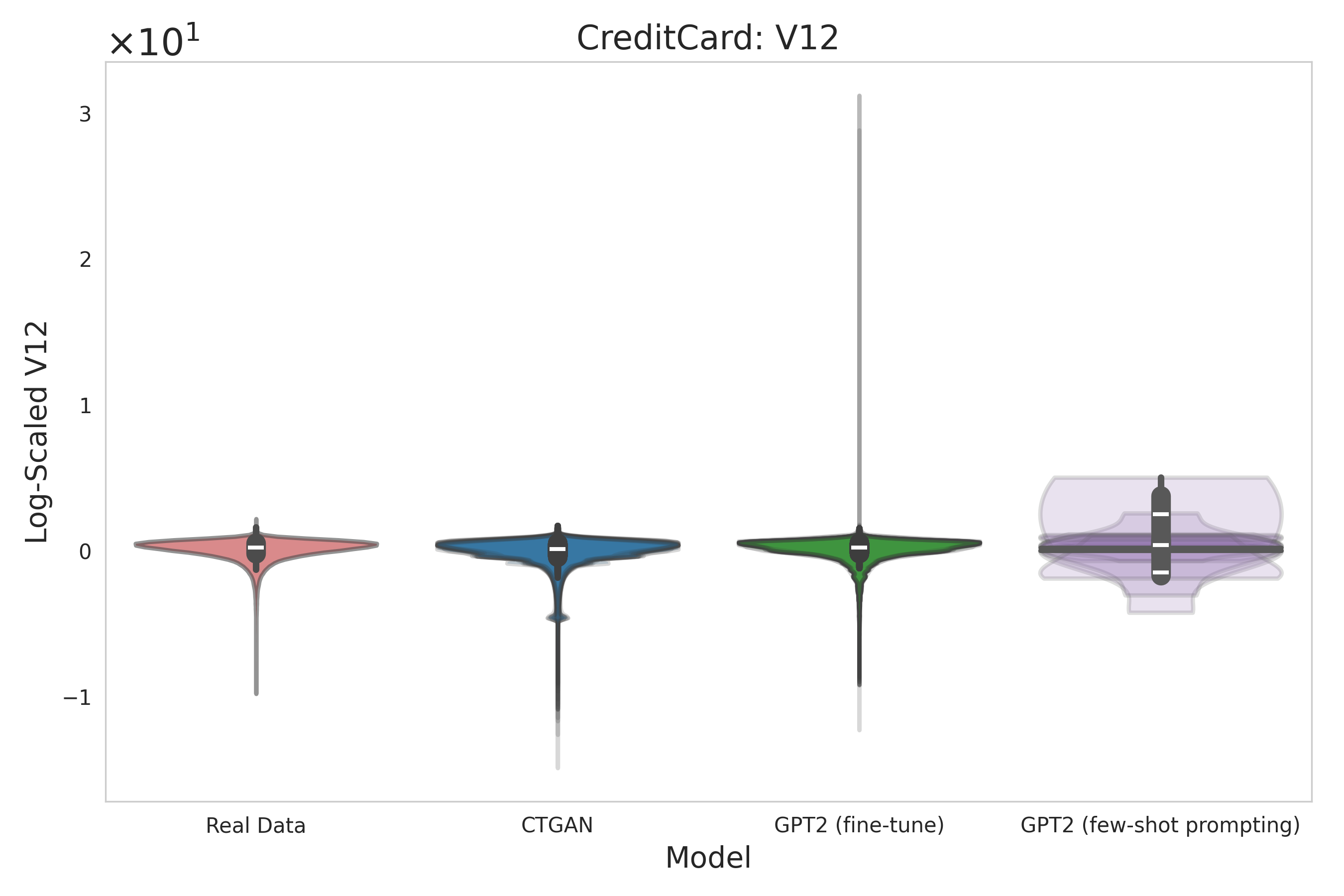}\\
    \caption{Further violin plots show the distributions of continuous columns in the Credit dataset.}
    \label{fig:credit2_violins}
\end{figure}

\begin{figure}
    \centering
    \includegraphics[width=0.49\linewidth]{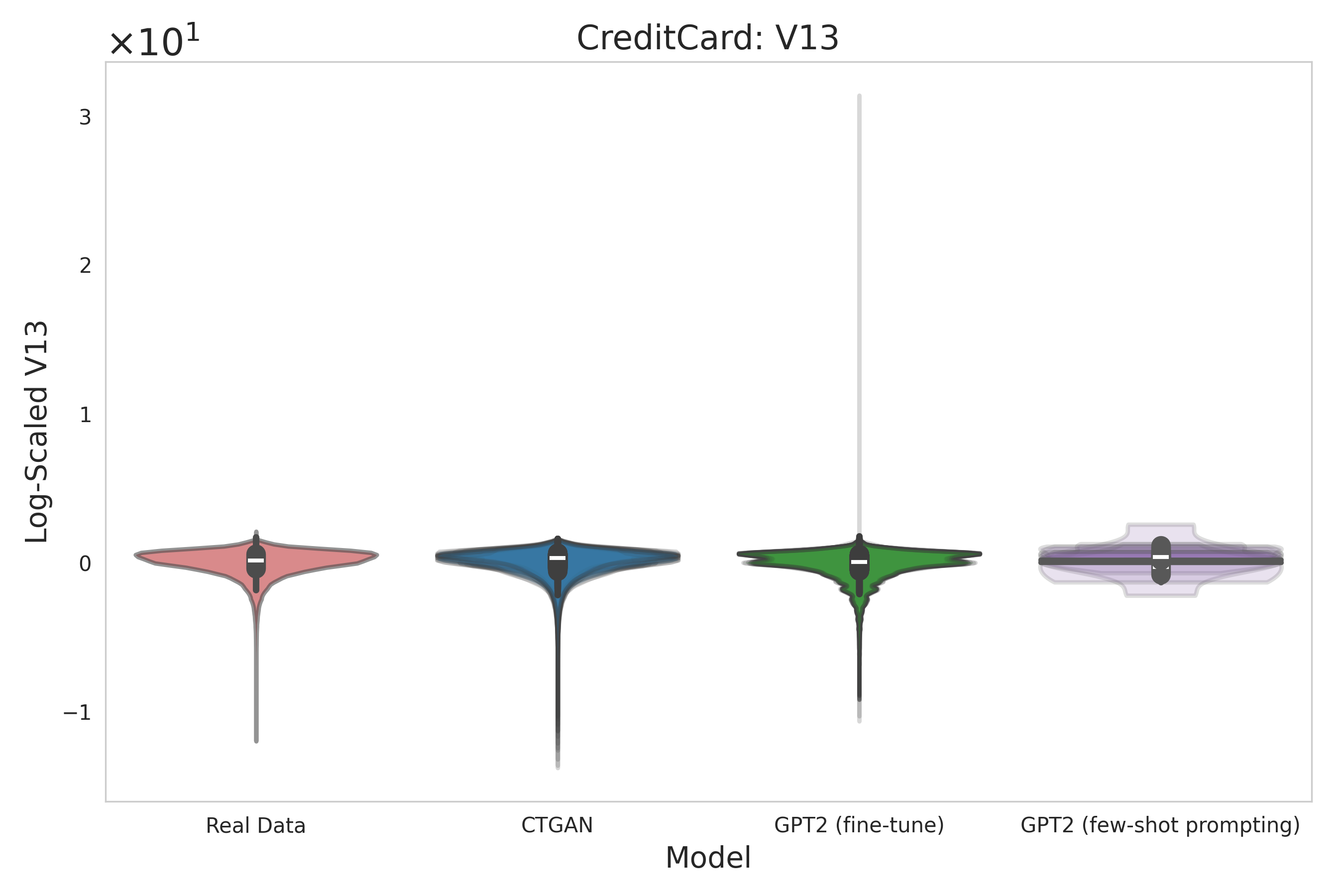}
    \includegraphics[width=0.49\linewidth]{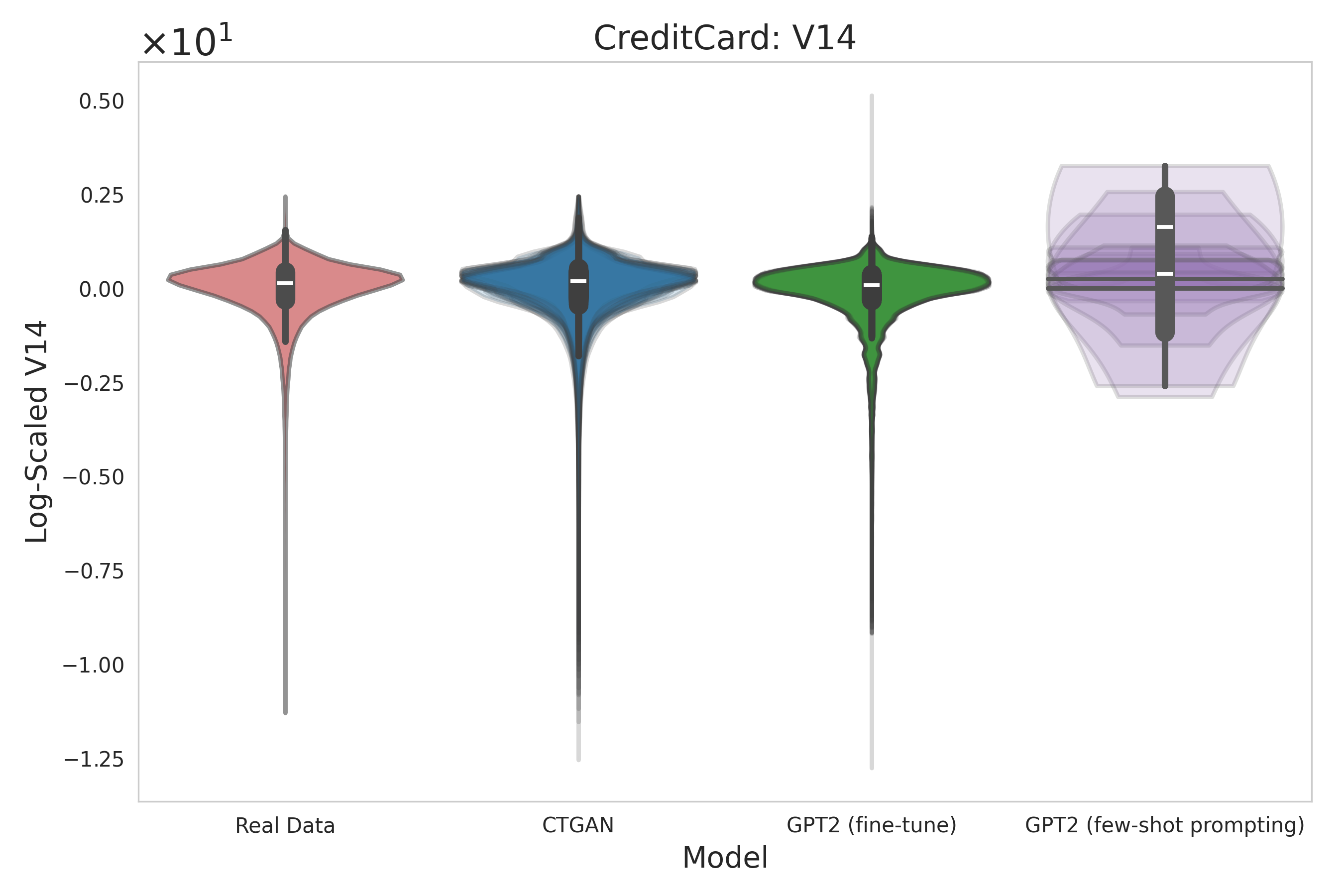}\\
    \includegraphics[width=0.49\linewidth]{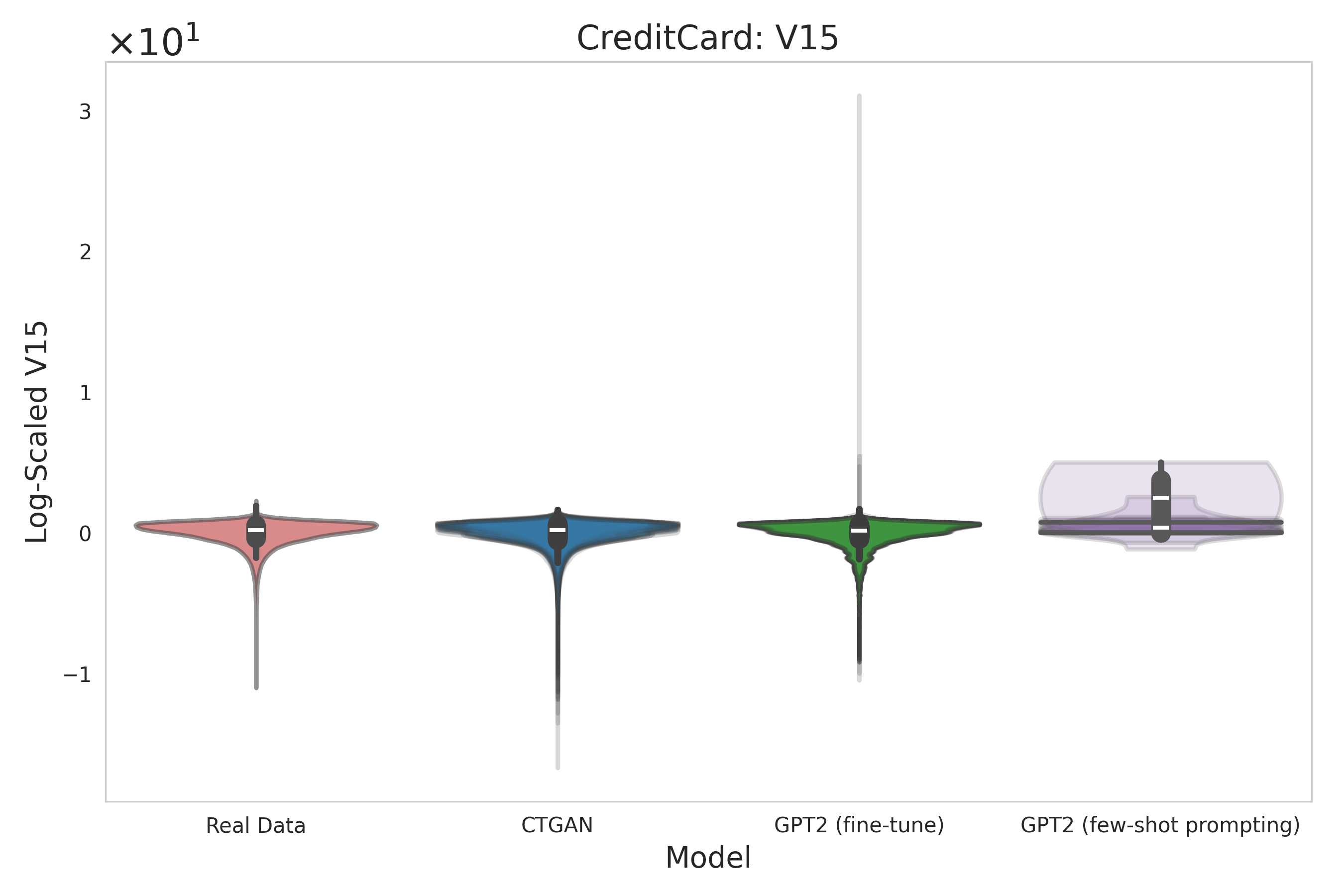}
    \includegraphics[width=0.49\linewidth]{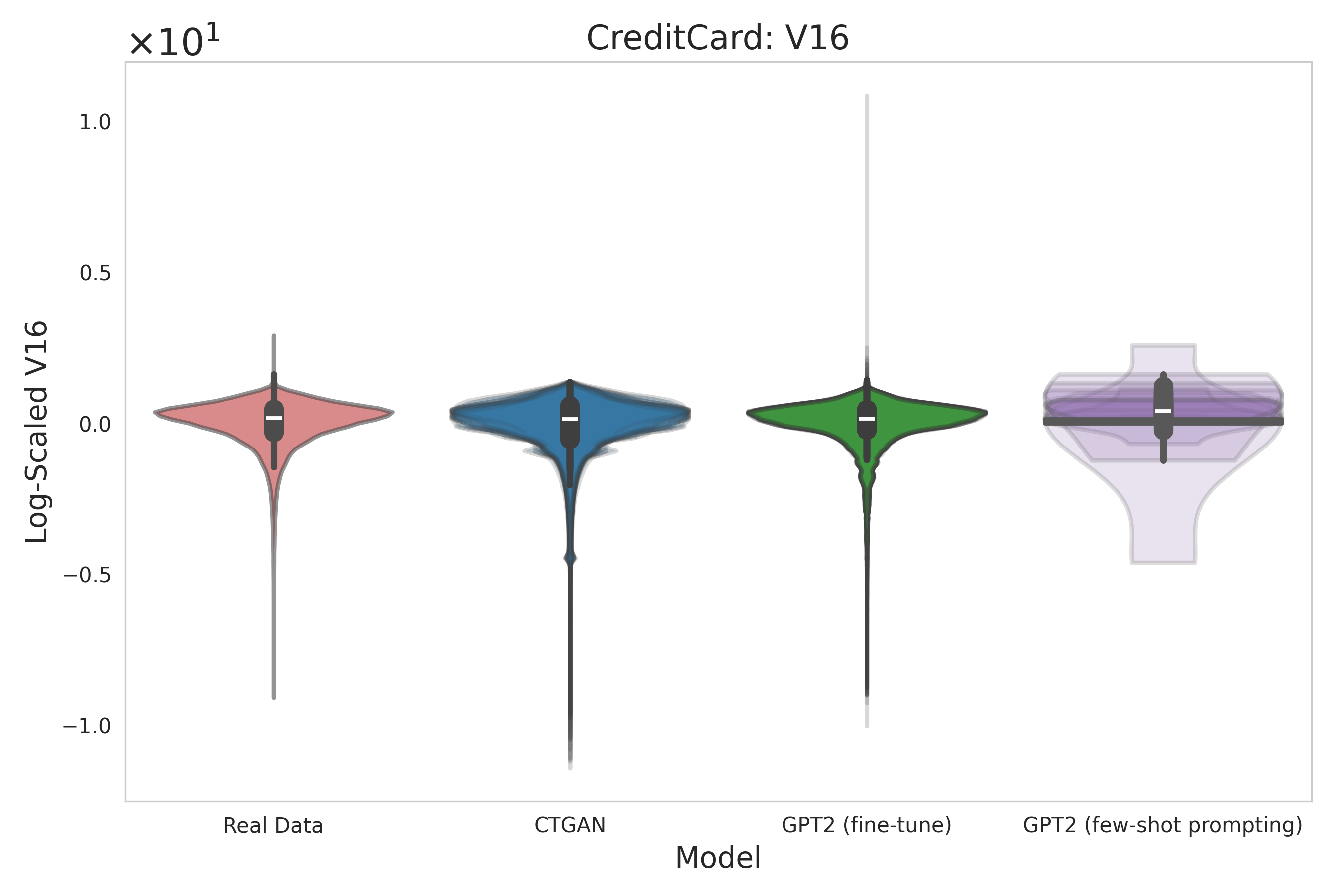}\\
    \includegraphics[width=0.49\linewidth]{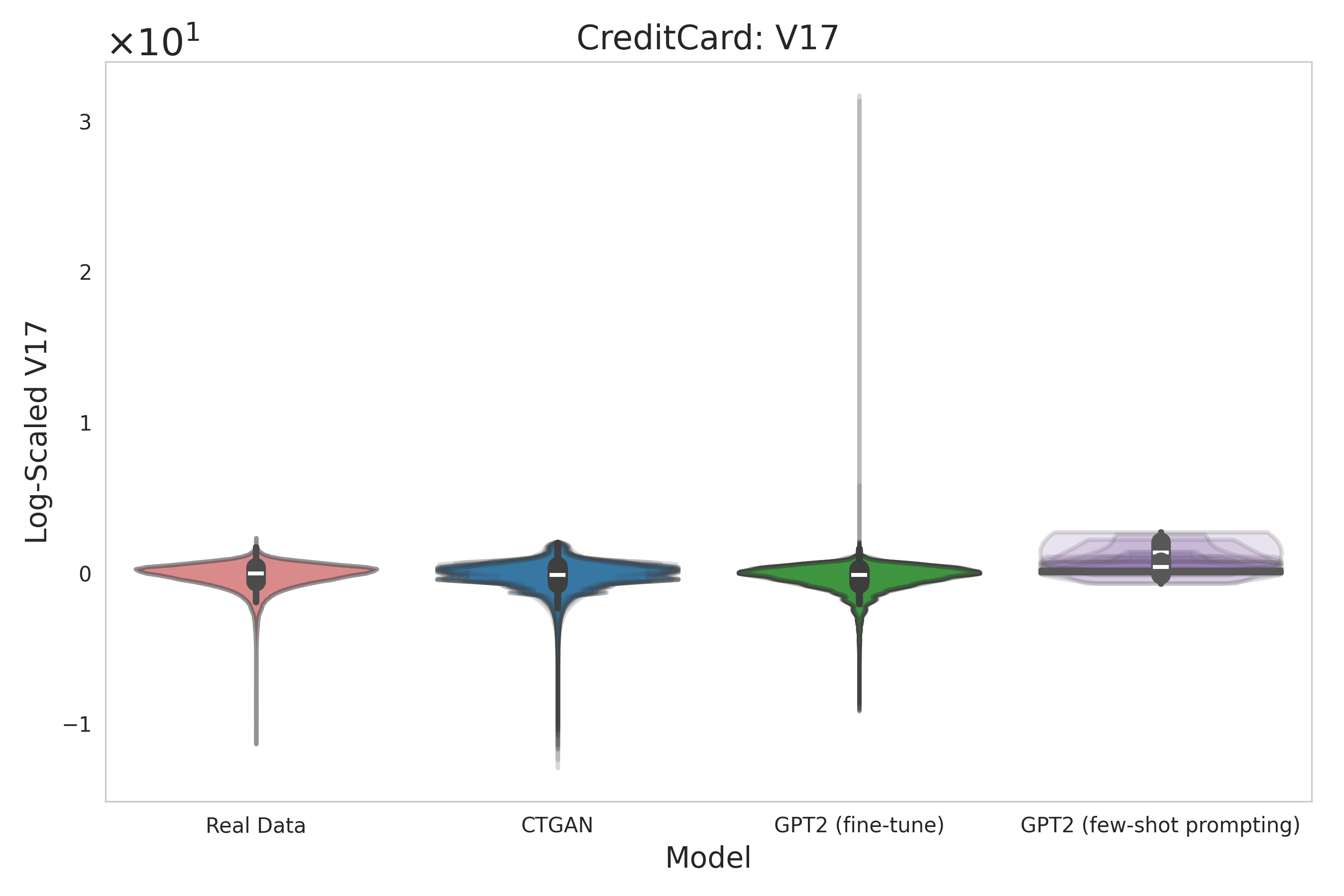}
    \includegraphics[width=0.49\linewidth]{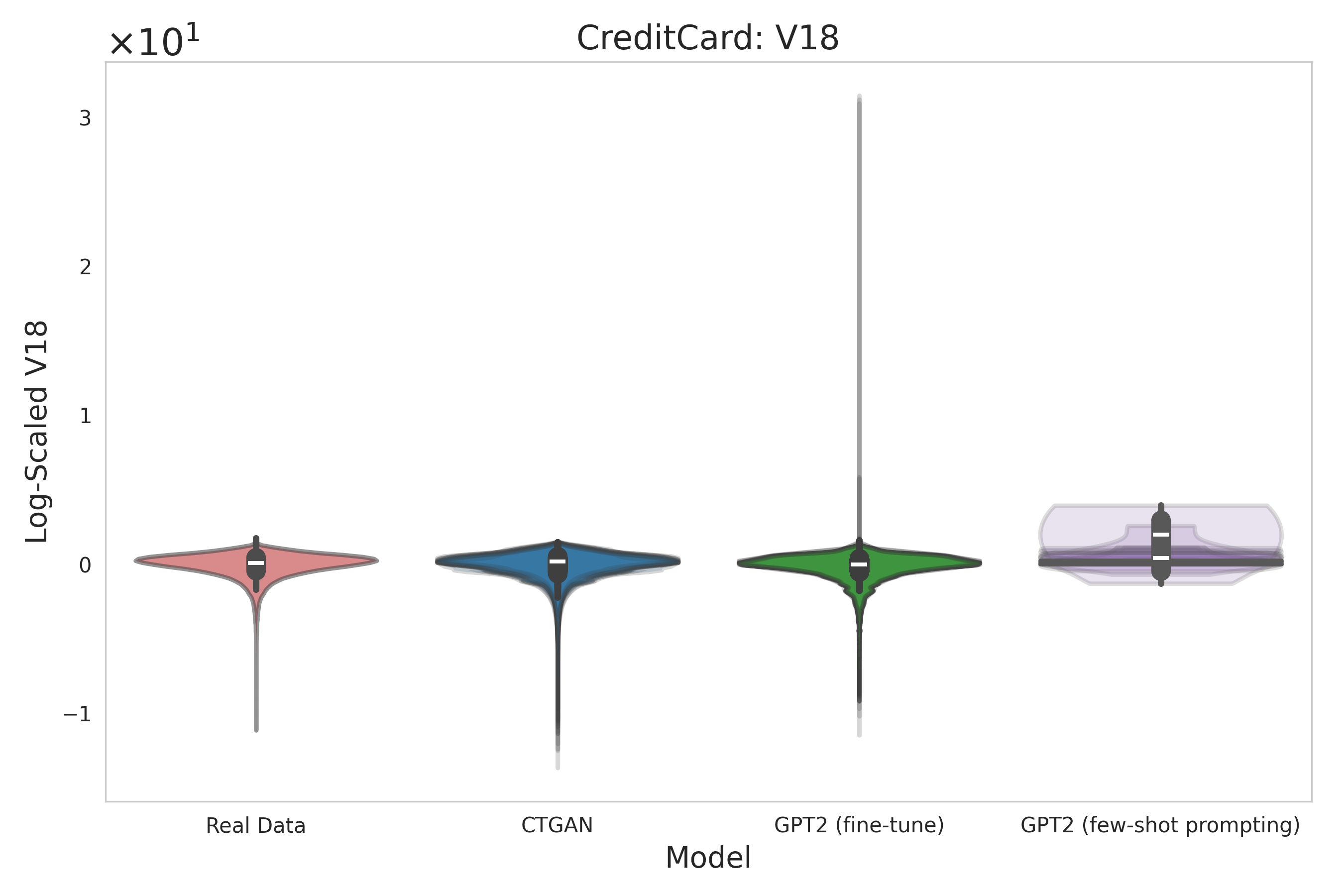}\\
    \caption{Further violin plots show the distributions of continuous columns in the Credit dataset.}
    \label{fig:credit3_violins}
\end{figure}

\begin{figure}
    \centering
    \includegraphics[width=0.49\linewidth]{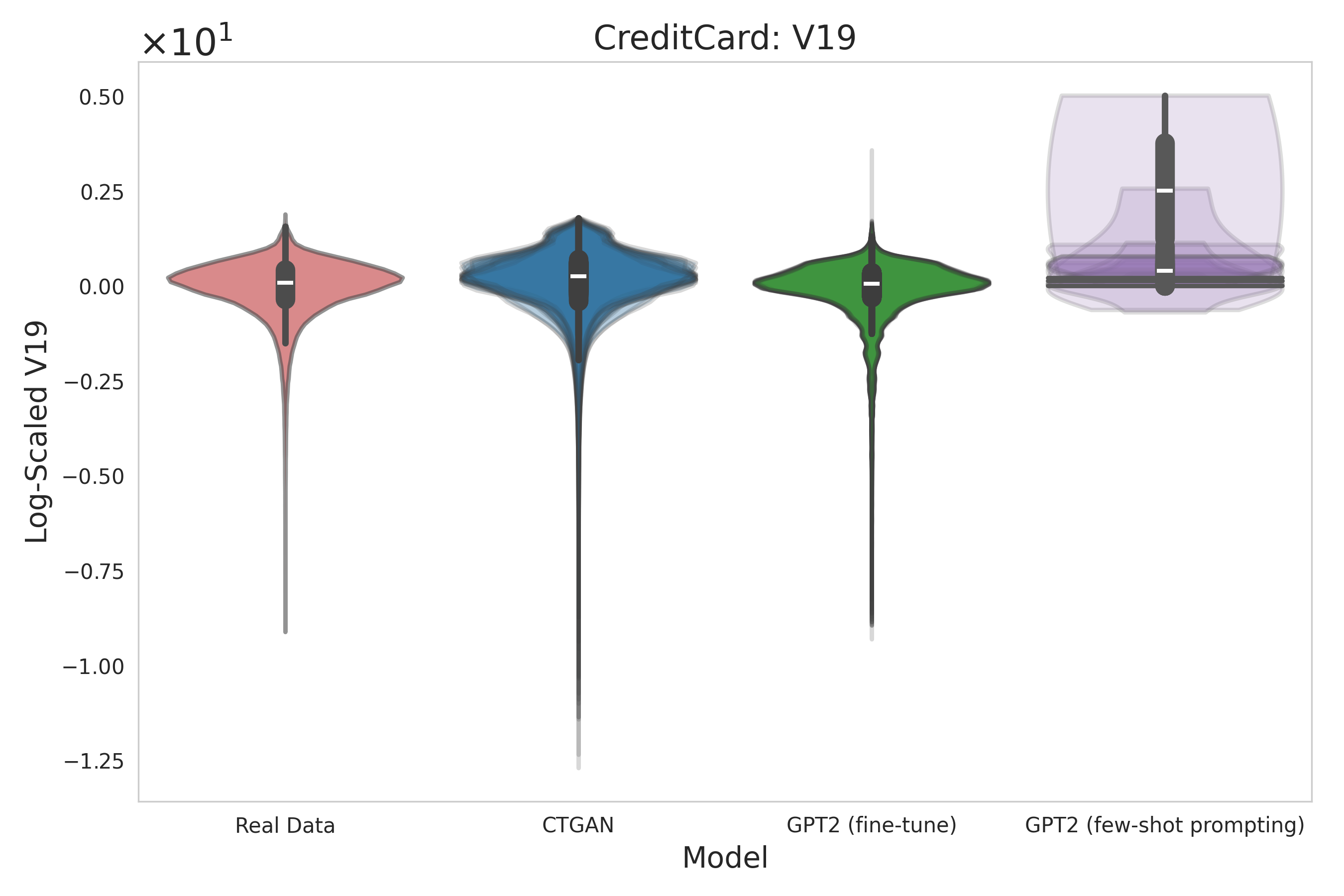}
    \includegraphics[width=0.49\linewidth]{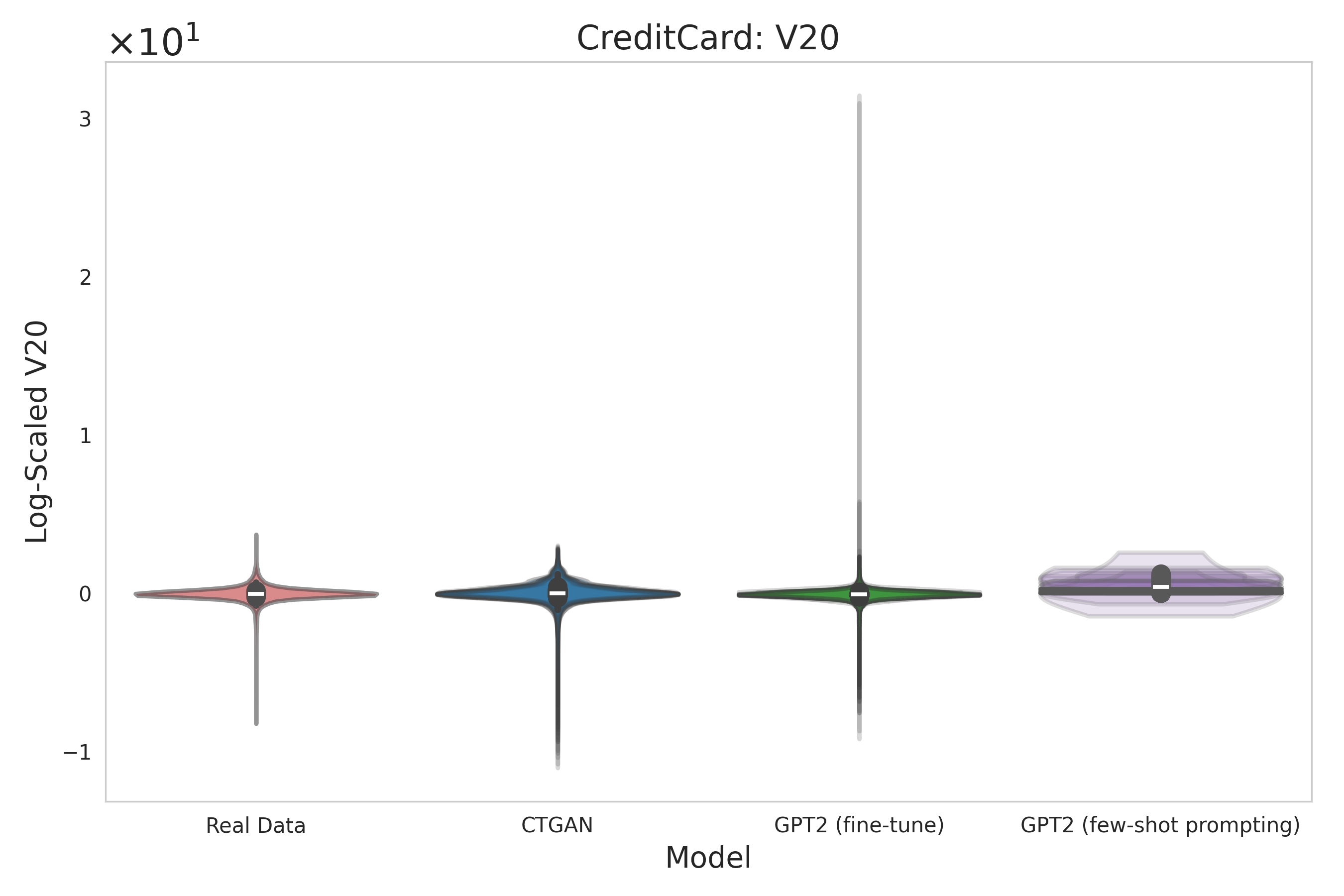}\\
    \includegraphics[width=0.49\linewidth]{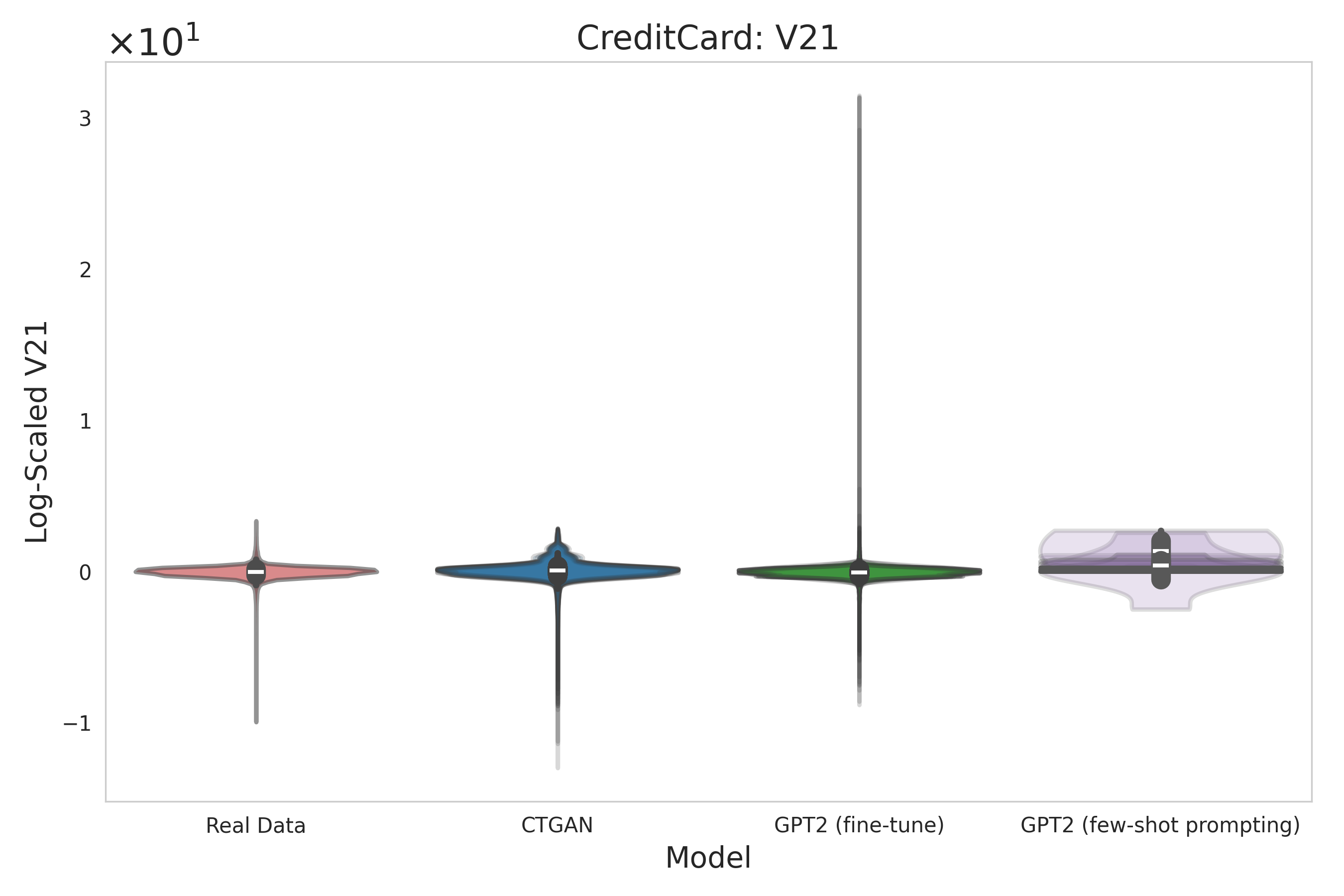}
    \includegraphics[width=0.49\linewidth]{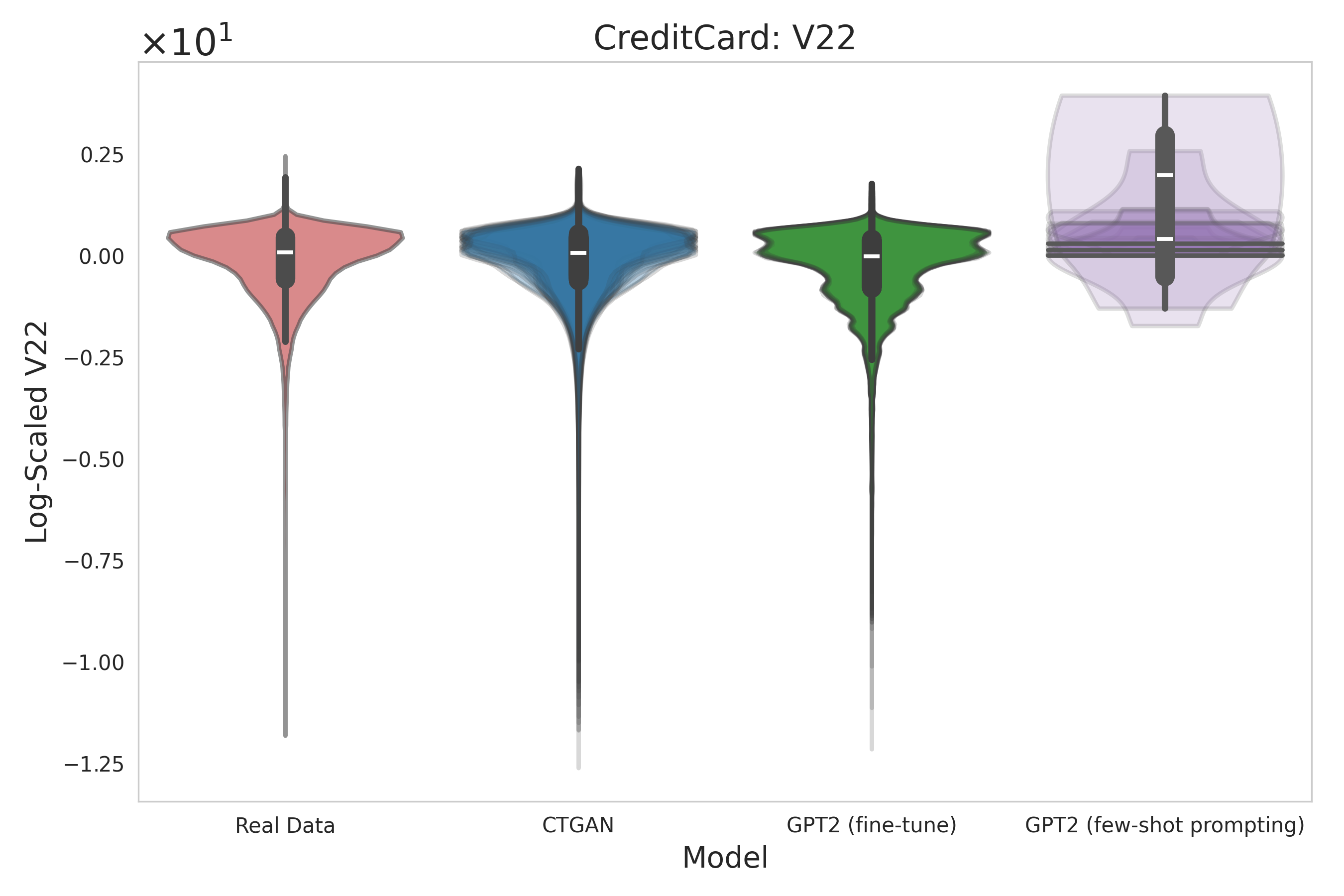}\\
    \includegraphics[width=0.49\linewidth]{figures/violins/credit/creditcard_V23.png}
    \includegraphics[width=0.49\linewidth]{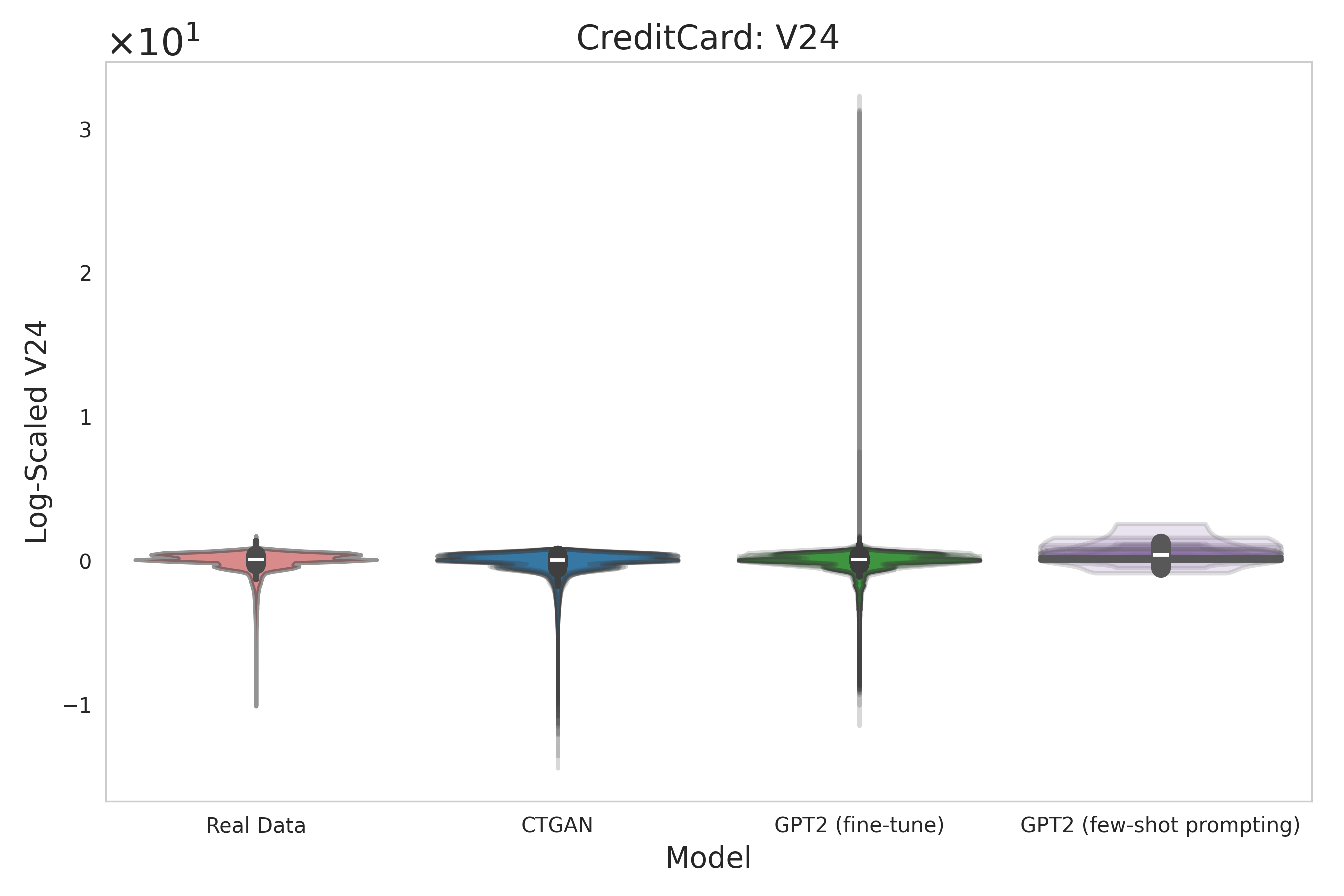}\\
    \caption{Further violin plots show the distributions of continuous columns in the Credit dataset.}
    \label{fig:credit4_violins}
\end{figure}

\begin{figure}
    \centering
    \includegraphics[width=0.49\linewidth]{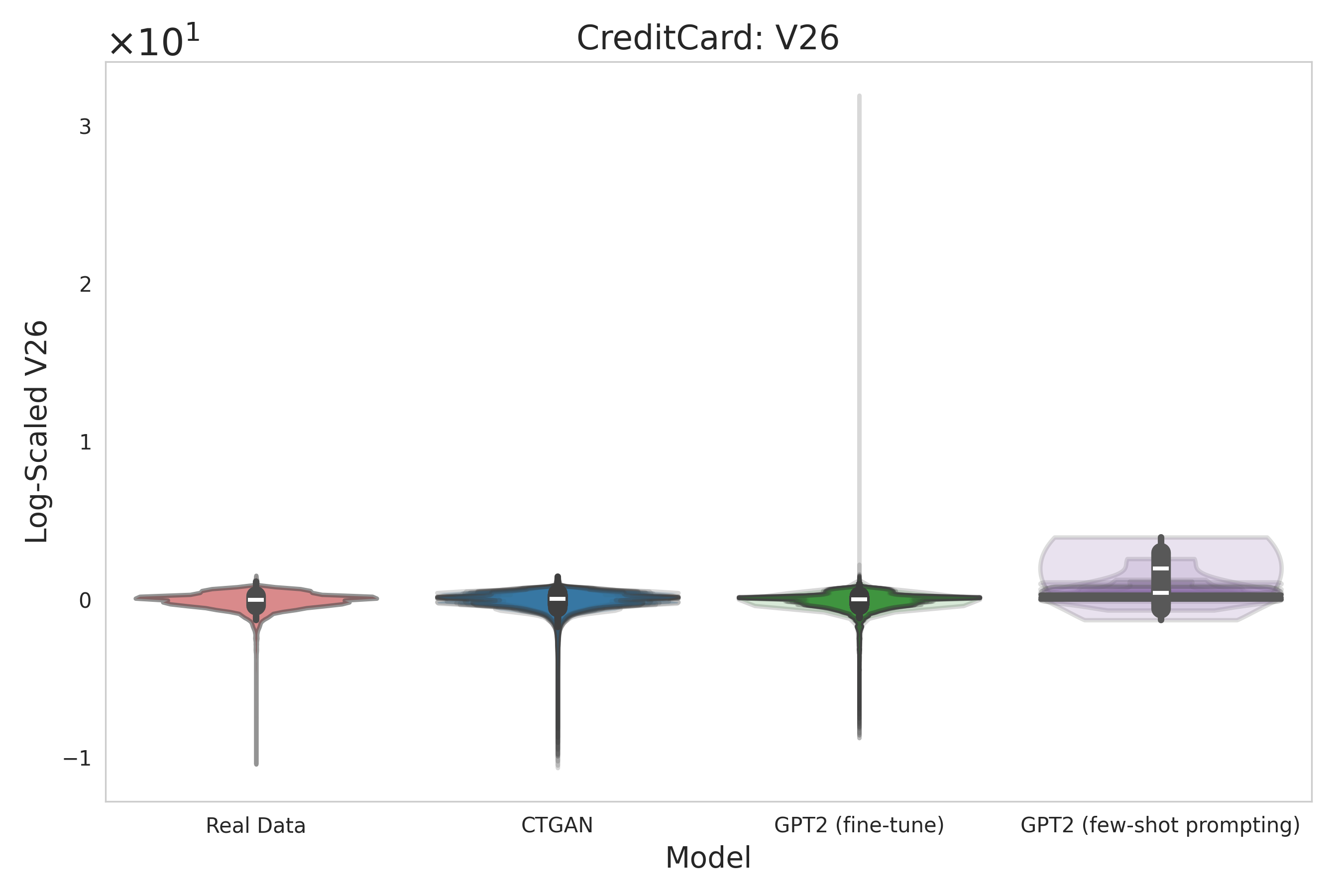}
    \includegraphics[width=0.49\linewidth]{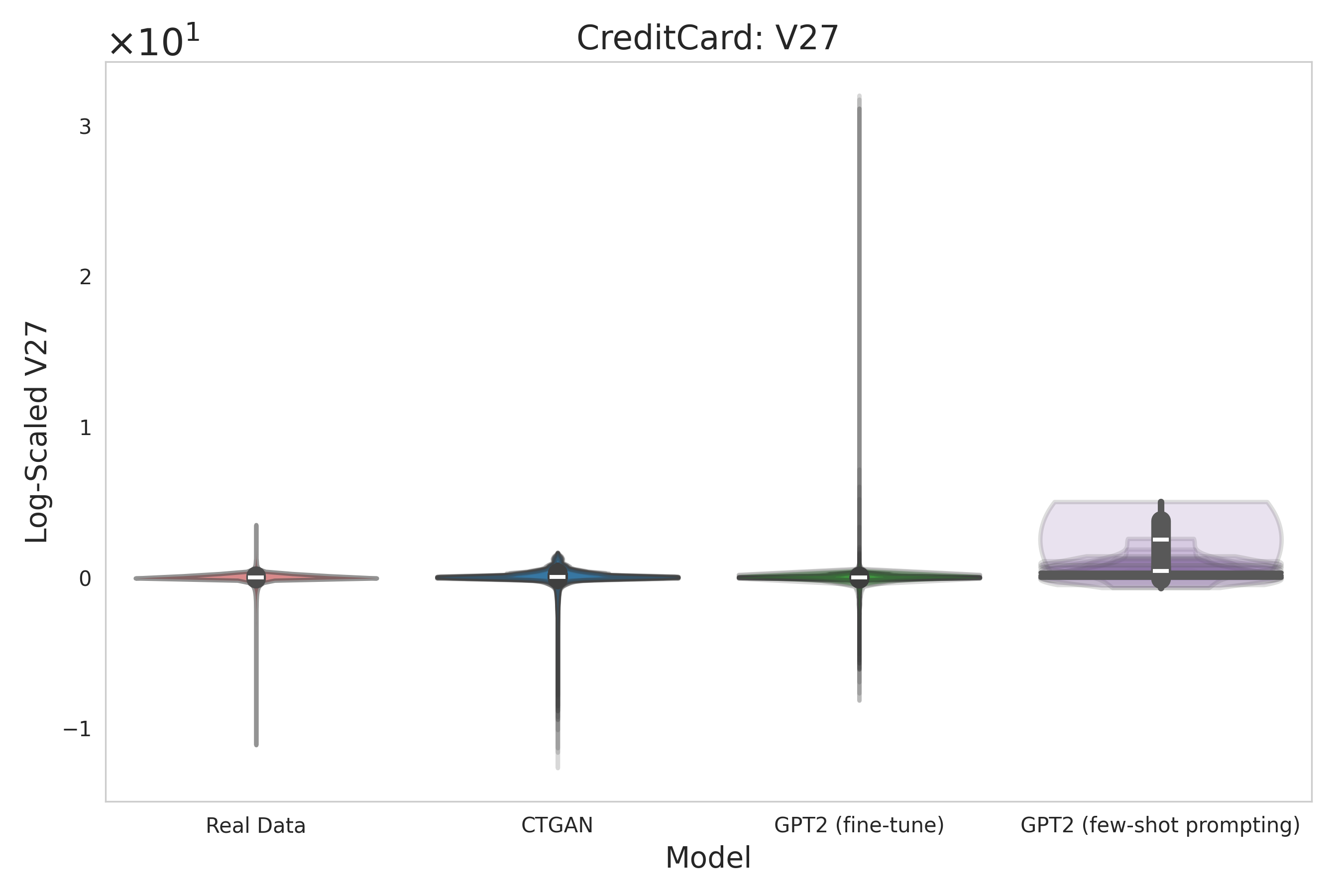}\\
    \includegraphics[width=0.49\linewidth]{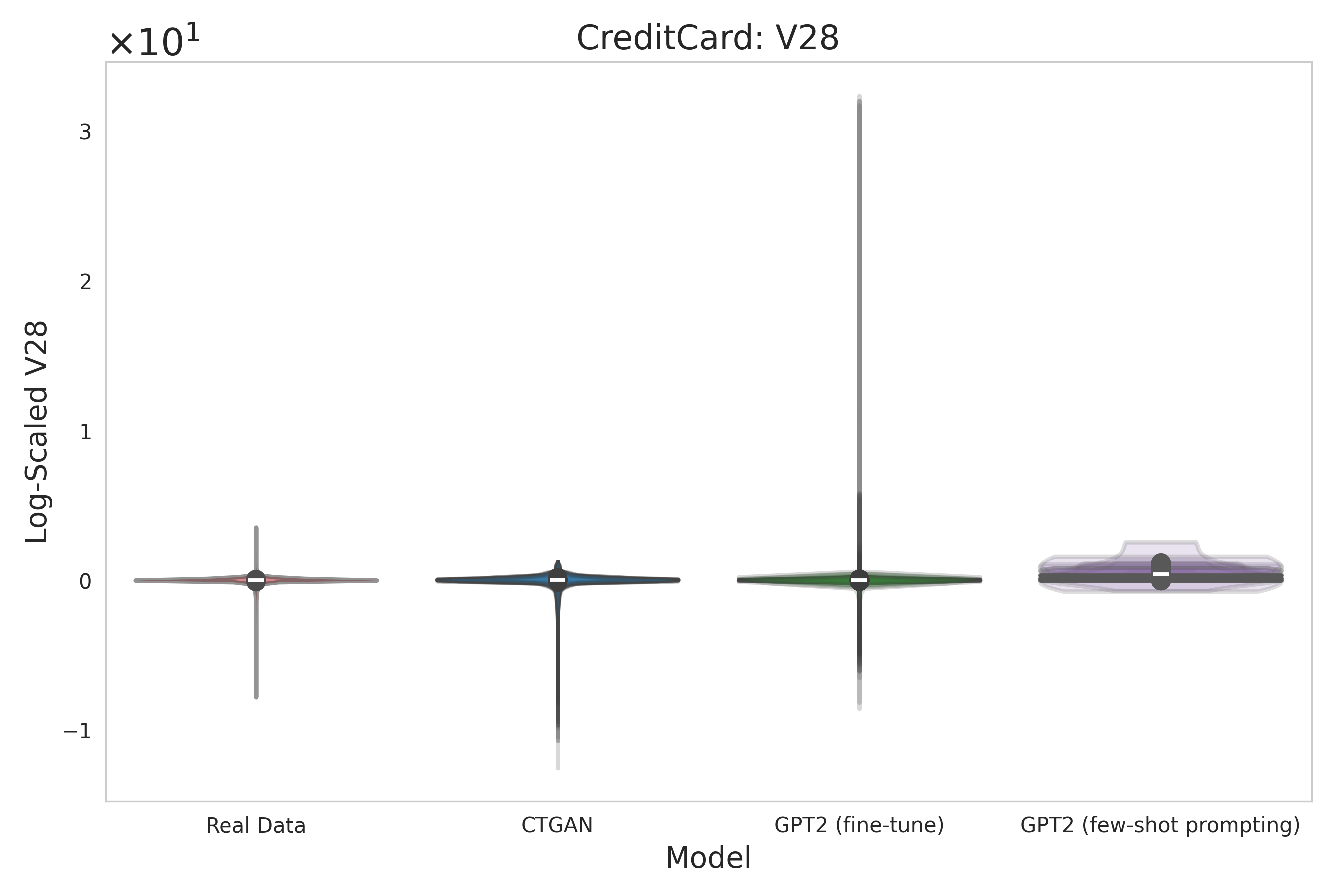}
    \includegraphics[width=0.49\linewidth]{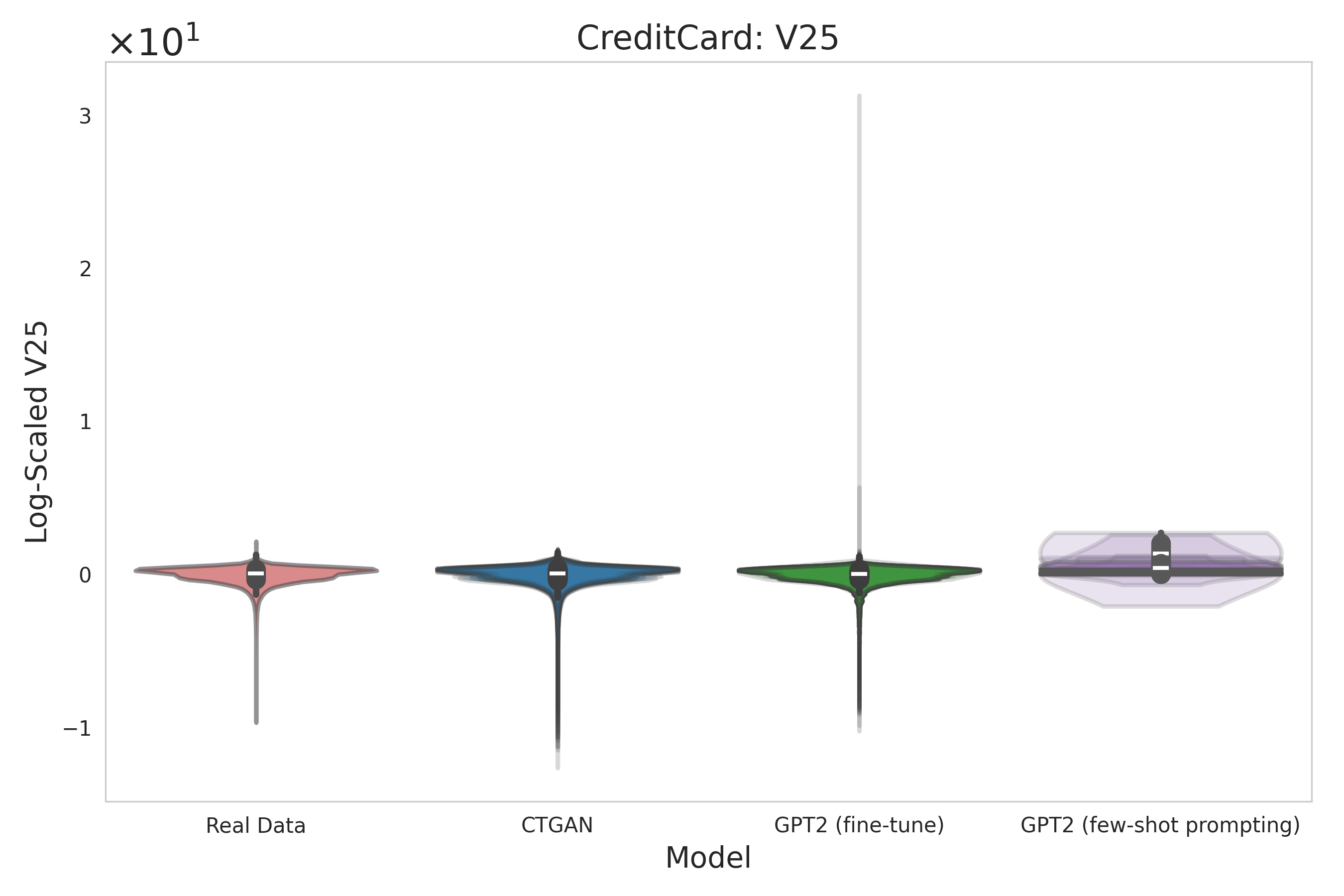}\\
    \caption{Further violin plots show the distributions of continuous columns in the Credit dataset.}
    \label{fig:credit5_violins}
\end{figure}

\begin{figure}
\centering
    \begin{tabular}{ll}
        a & b \\
        \includegraphics[width=0.49\linewidth]{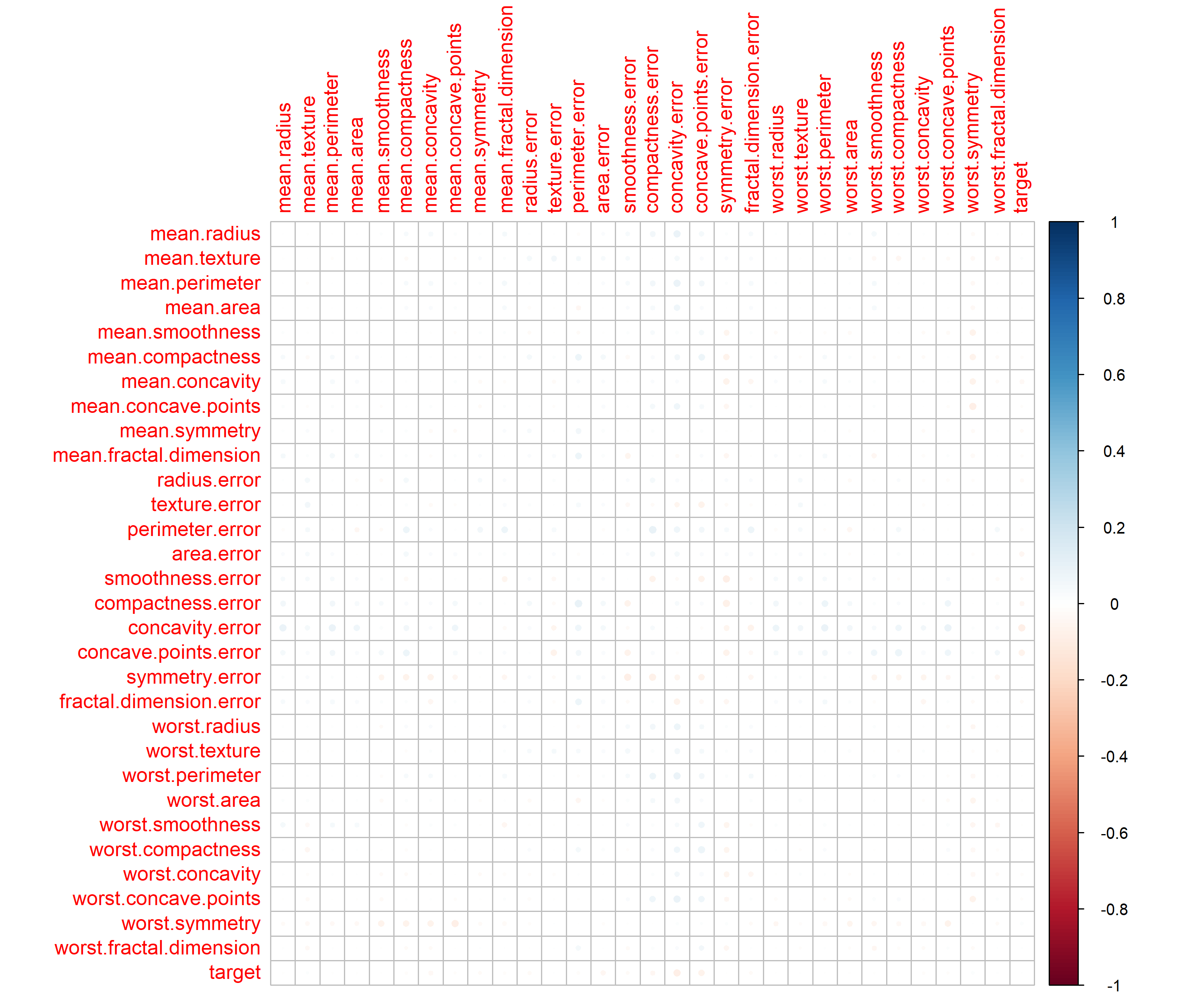} &    \includegraphics[width=0.49\linewidth]{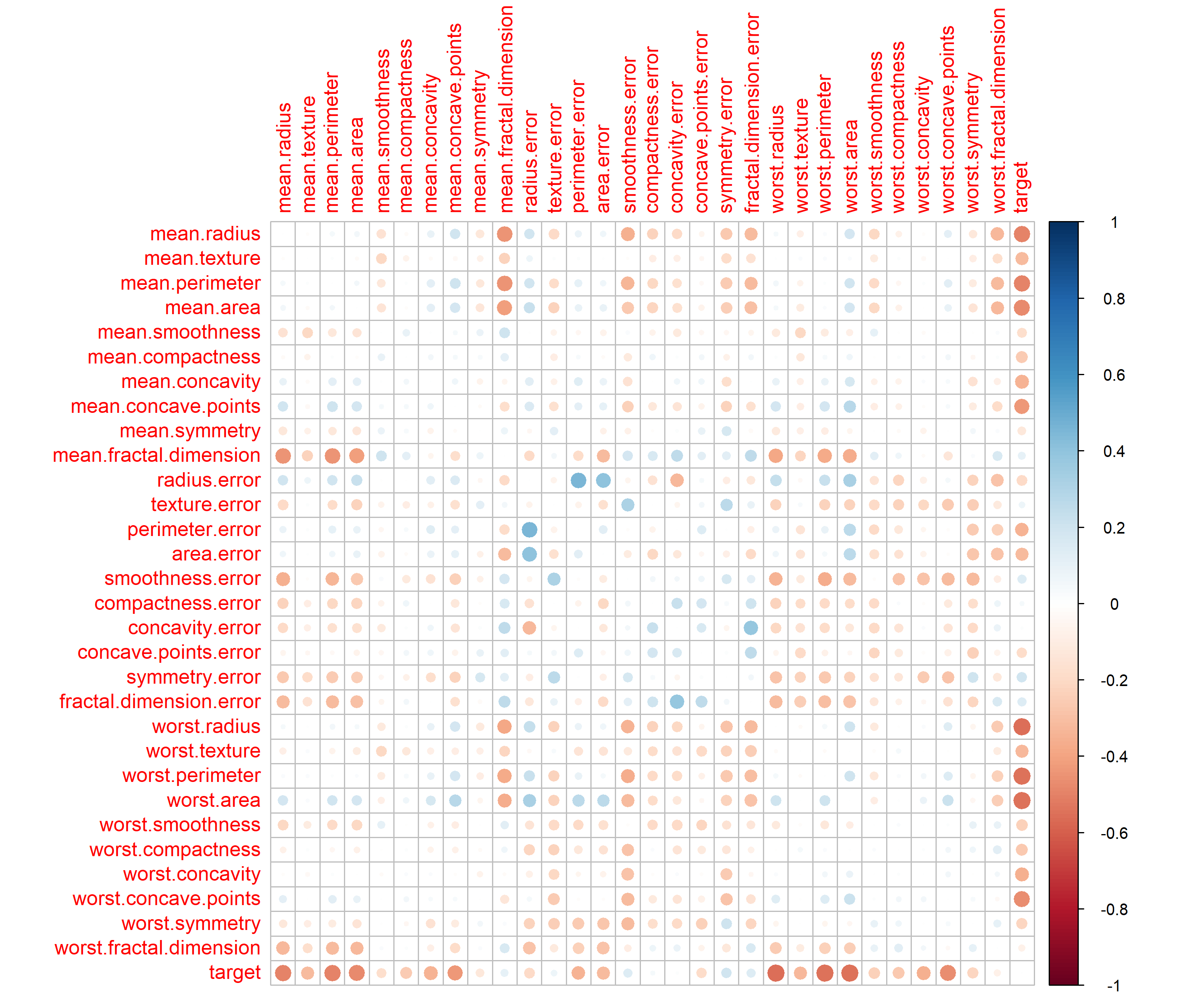}\\
        c & d \\
        \includegraphics[width=0.49\linewidth]{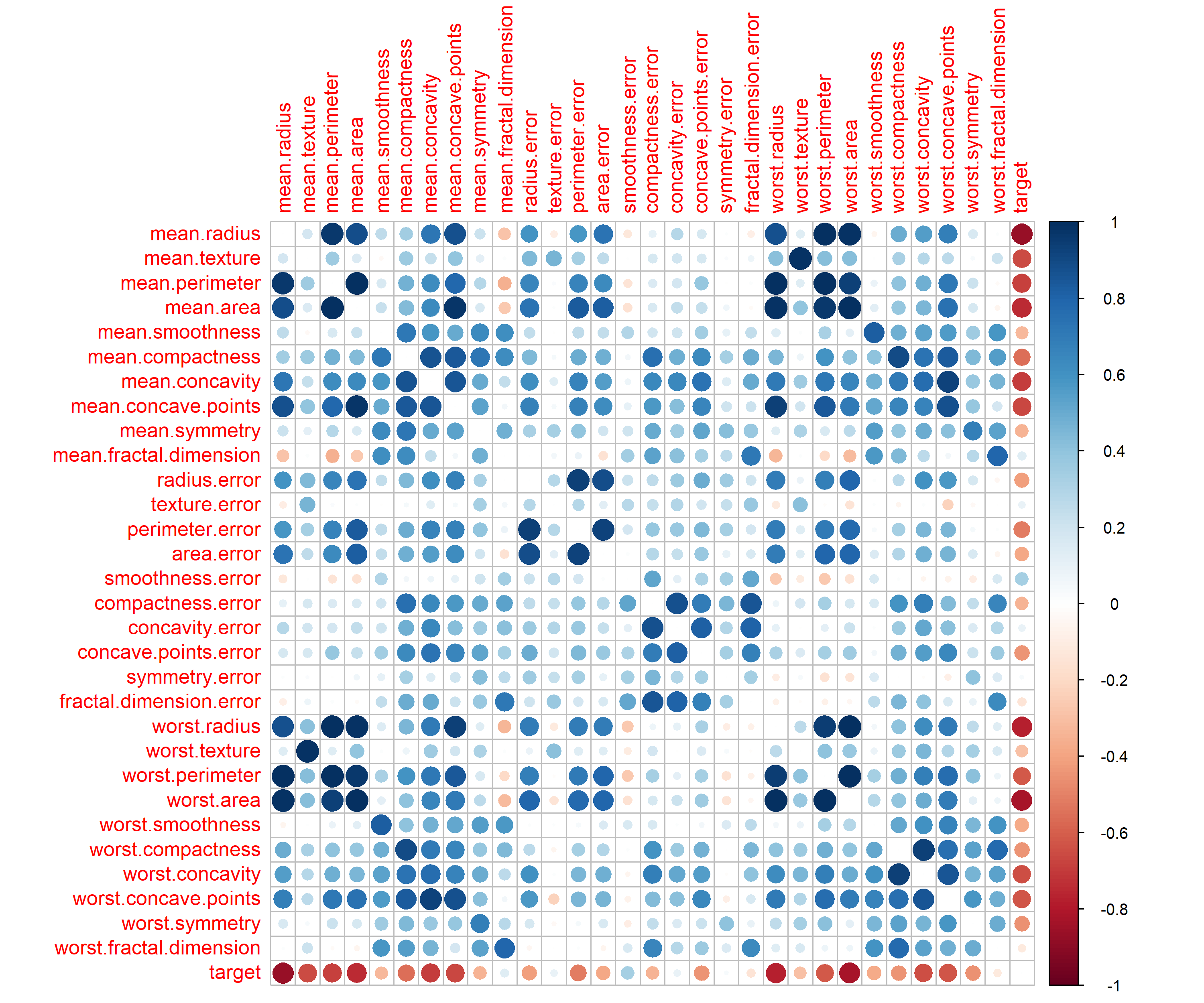} &
        \includegraphics[width=0.49\linewidth]{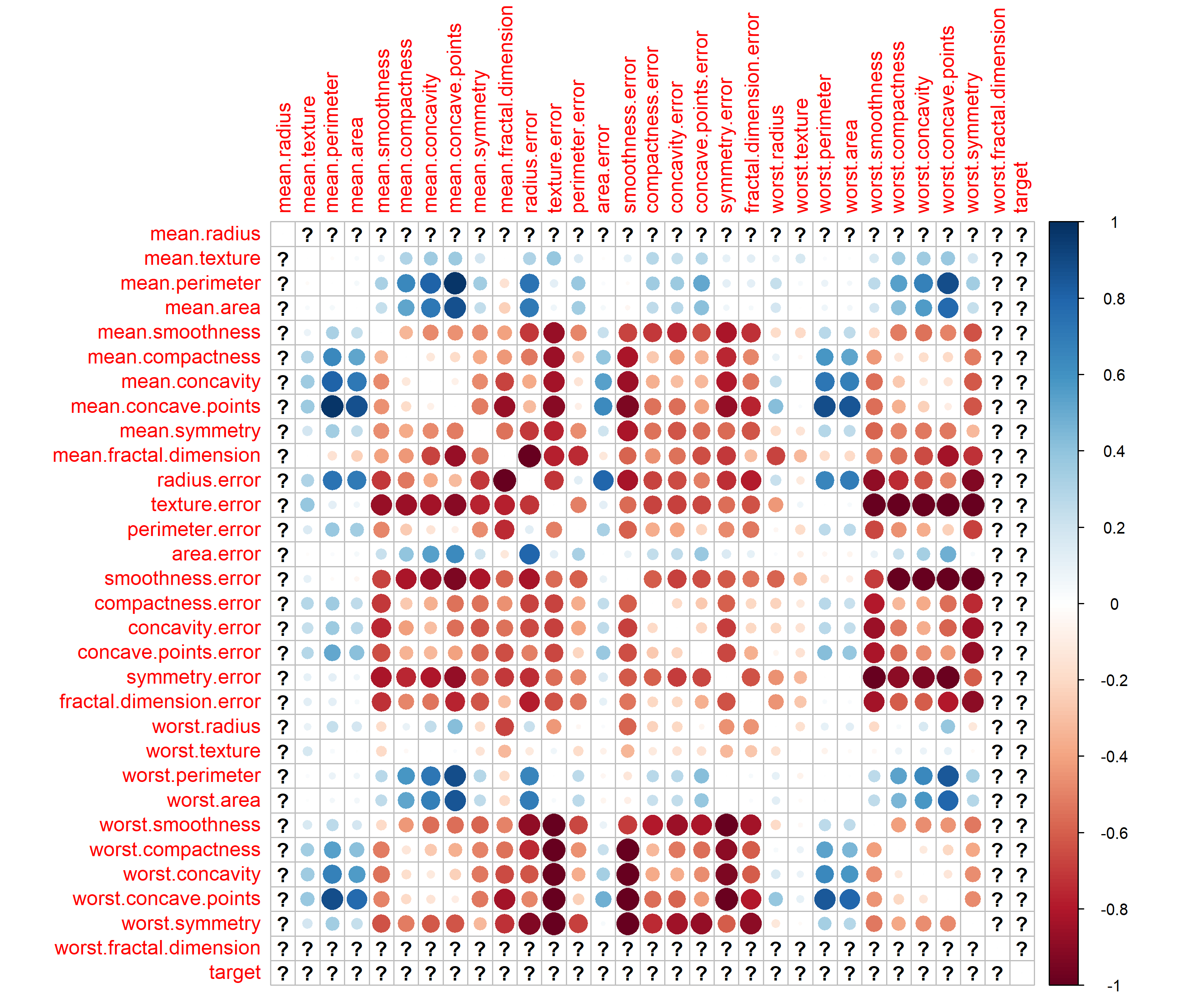}\\
    \end{tabular}
    \caption{Heatmap for the Breast Dataset: (a) Resampling of the train set, (b) Fine-tuned GPT-2, (c) CTGAN, (d) GPT-2 few-shot prompting. The models shown here correspond to those with the smallest absolute determinant with the dependencies between the original train set. Questions marks denote situations where the dependence measure was unable to be computed due to zero variation or inconsistent format.}
    \label{fig:breast_heatmap}
\end{figure}

\begin{figure}
    \begin{tabular}{ll}
        a & b \\
        \includegraphics[width=0.49\linewidth]{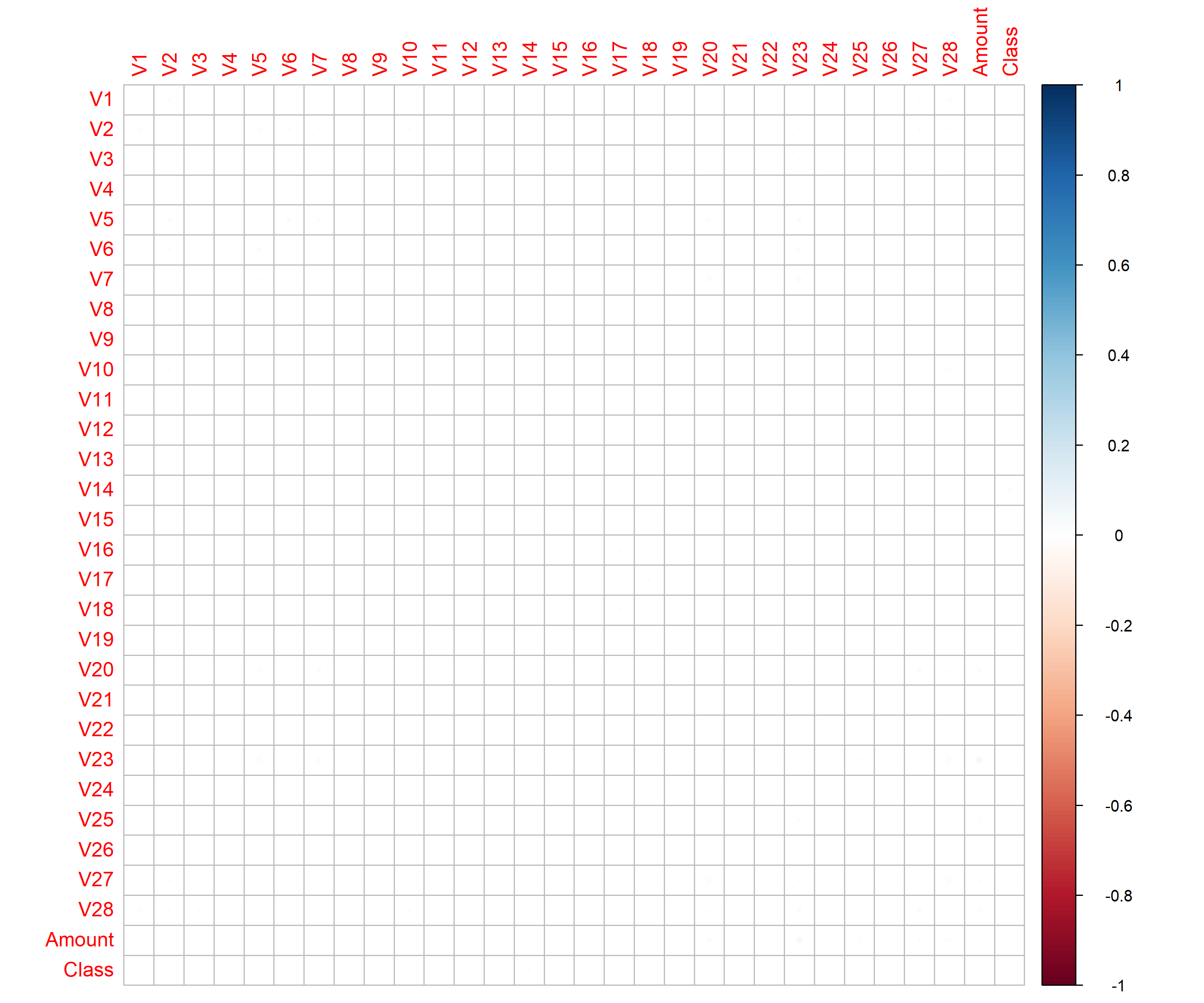} & 
        \includegraphics[width=0.49\linewidth]{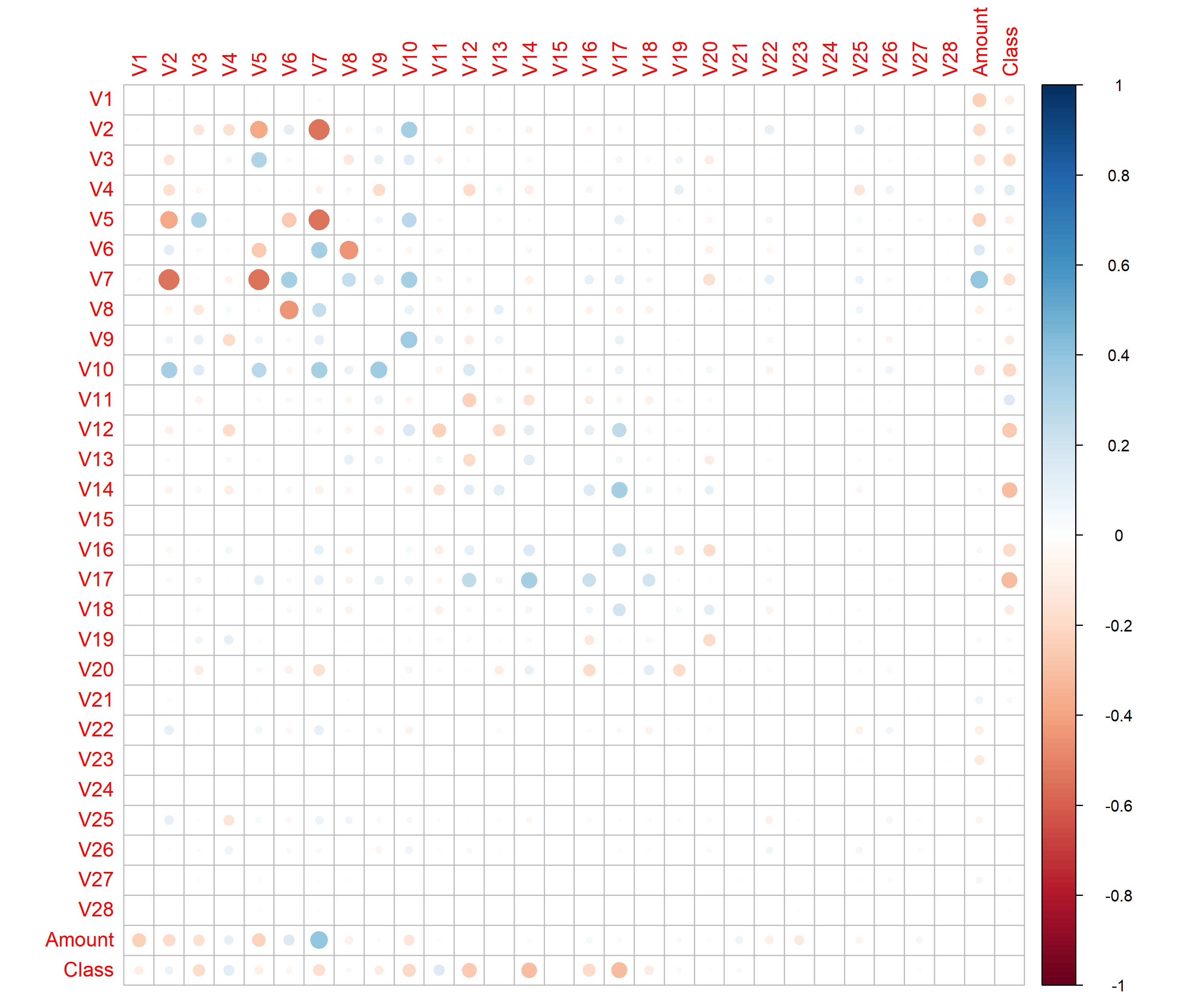}\\
        c & d \\
        \includegraphics[width=0.49\linewidth]{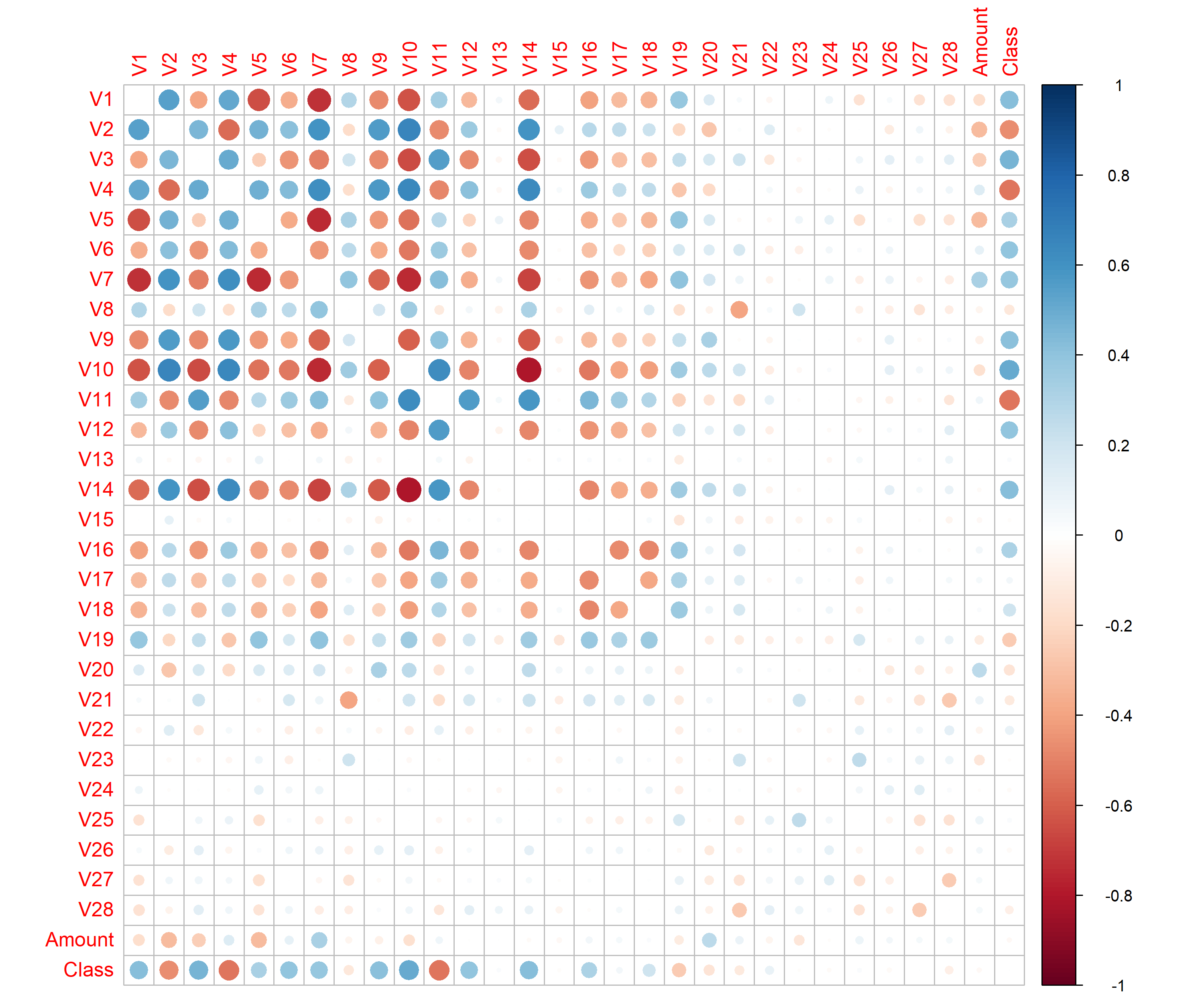} &
        \includegraphics[width=0.49\linewidth]{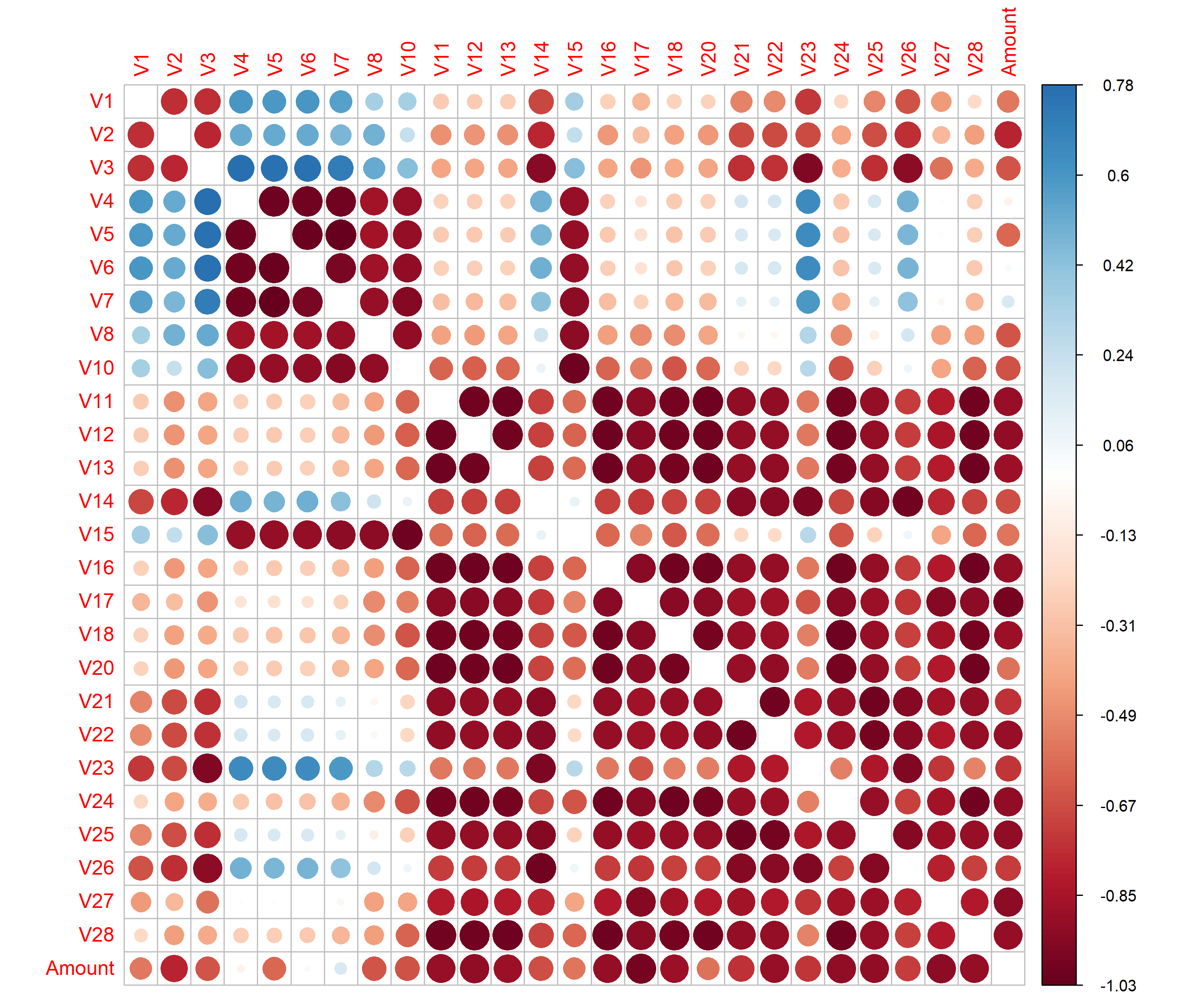}\\
    \end{tabular}
    \caption{Heatmap for the Credit Dataset: (a) Resampling of the train set, (b) Fine-tuned GPT-2, (c) CTGAN, (d) GPT-2 few-shot prompting. The models shown here correspond to those with the smallest absolute determinant with the dependencies between the original train set.}
    \label{fig:credit_heatmap}
\end{figure}

\begin{figure}
    \centering
    \includegraphics[width=0.9\linewidth]{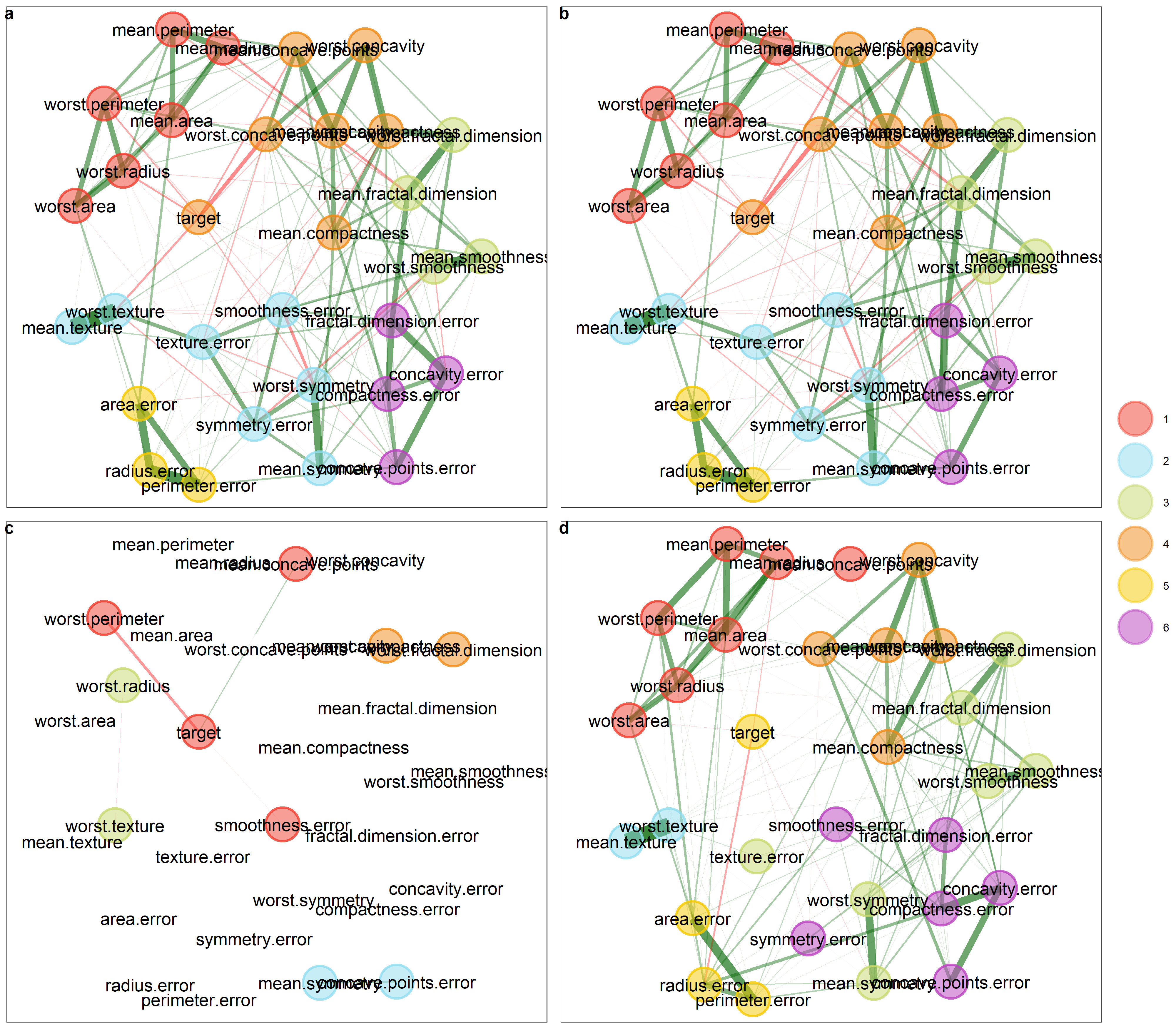}
    \caption{Representative network plots of the dependency between real and synthetic data for the Breast dataset: (a) train set, (b) resampling of the train set, (c) CTGAN, and (d) GPT-2 fine-tuned. The node colors represent the clusters found using the Louvain community detection algorithm. The edge colors reflect the directionality of the relationships (red for negative correlations, green for positive correlations). Displayed models were chosen according to the lowest absolute determinant of the dependency matrix.}
    \label{fig:breast_graph}
\end{figure}

\begin{figure}
    \centering
    \includegraphics[width=0.9\linewidth]{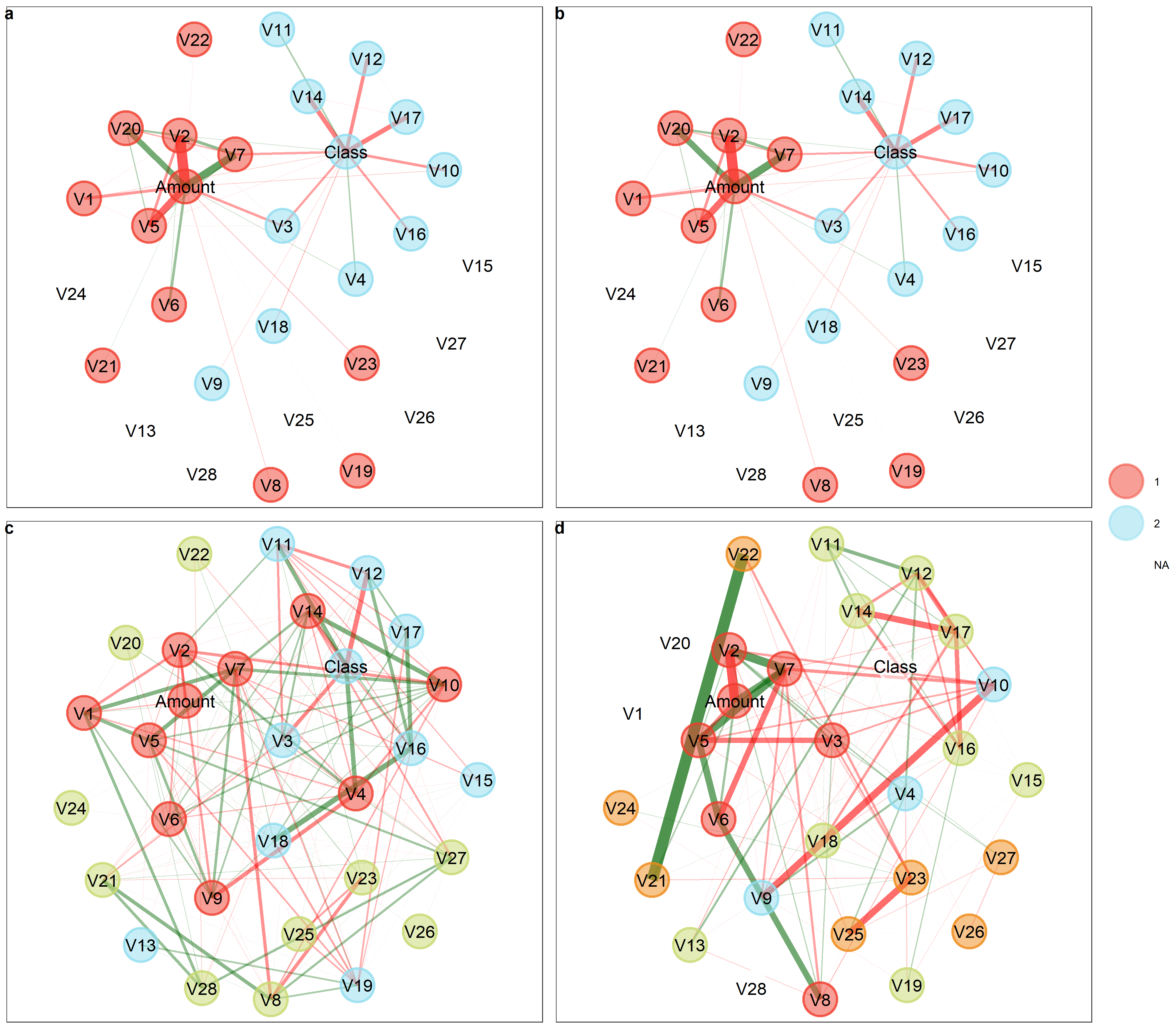}
    \caption{Representative network plots of the dependency between real and synthetic data for the Credit dataset: (a) train set, (b) resampling of the train set, (c) CTGAN, and (d) GPT-2 fine-tuned. The node colors represent the clusters found using the Louvain community detection algorithm. The edge colors reflect the directionality of the relationships (red for negative correlations, green for positive correlations). Displayed models were chosen according to the lowest absolute determinant of the dependency matrix.}
    \label{fig:credit_graph}
\end{figure}

\begin{table}
\centering
    \begin{tabular}{lcccccc}
    \multirow{2}{*}{Trial} & \multicolumn{2}{c}{Adult} & \multicolumn{2}{c}{Breast} & \multicolumn{2}{c}{Credit} \\
    & $N_{\text{dropped}} \ (\%) $ & $N_{\text{remain}} \ (\%)$ & $N_{\text{dropped}} \ (\%)$ & $N_{\text{remain}} \ (\%)$ & $N_{\text{dropped}} \ (\%)$ & $N_{\text{remain}} \ (\%)$ \\
    \hline
    0 & 6049 (20.8) & 23039 (79.2) & 6 (1.5) & 392 (98.5) & 13 (72.2) & 5 (27.8) \\
    1 & 7001 (22.1) & 24657 (77.9) & 2 (0.5) & 396 (99.5) & 5 (62.5) & 3 (37.5) \\
    2 & 7054 (22.3) & 24603 (77.7) & 7 (1.8) & 391 (98.2) & 6 (46.2) & 7 (53.8)\\
    3 & 6898 (21.8) & 24761 (78.2) & 4 (1.0) & 394 (99.0) & 9 (60.0) & 6 (40.0)\\
    4 & 1705 (21.3) & 6294 (78.7) & 1 (0.3) & 398 (99.7) & 10 (71.4) & 4 (28.6) \\
    5 & 1446 (19.9) & 5821 (80.1) & 3 (0.8) & 396 (99.2) & 2 (33.3) & 4 (66.7) \\
    6 & 6703 (21.2) & 24958 (78.8) & 8 (2.0) & 390 (98.0) & 5 (50.0) & 5 (50.0)\\
    7 & 6817 (21.5) & 24844 (78.5) & 2 (0.5) & 396 (99.5) & 6 (85.7) & 1 (14.3) \\
    8 & 6628 (21.6) & 23991 (78.4) & 7 (1.8) & 391 (98.2) & 1 (50.0) & 1 (50) \\
    9 & 6869 (21.7) & 24790 (78.3) & 8 (2.0) & 391 (98.0) & 8 (57.1) & 6 (42.9)\\
    10 & 6750 (21.3) & 24906 (78.7) & 5 (1.3) & 393 (98.7) & 6 (75.0) & 2 (25.0) \\
    11 & 6685 (21.1) & 24970 (78.9) & 3 (0.8) & 395 (99.2) & 6 (66.7) & 3 (33.3) \\
    12 & 7138 (22.5) & 24521 (77.5) & 4 (1.0) & 394 (99.0) & 2 (66.7) & 1 (33.3) \\
    13 & 2387 (22.1) & 8396 (77.9) & 9 (2.3) & 390 (97.7) & 8 (66.7) & 4 (33.3) \\
    14 & 6703 (21.9) & 23895 (78.1) & 5 (1.3) & 393 (98.7) & 12 (85.7) & 2 14.3) \\
    \hline
    \end{tabular}
    \caption{{\bf Number of failed example datapoints returned by few-shot prompted GPT-2.} A table of the number and percentage of malformed datapoints returned by few-shot prompted GPT-2 per trial. The number of datapoints that we prompted GPT-2 to return can be found at Supplemental Table \ref{table:zeroshot_examples_samples}. Note that the number of examples that GPT-2 returned can differ from the number we prompted it to return.}
    \label{table:cleaned_counts}
\end{table}
